# SEER: MAXIMUM LIKELIHOOD REGRESSION FOR LEARNING-SPEED CURVES

BY

CARL MYERS KADIE

B.S., University of Illinois, 1985
M.S., University of Illinois, 1989

THESIS

# Seer: Maximum Likelihood Regression for Learning-Speed Curves

Carl Myers Kadie
Department of Computer Science
University of Illinois at Urbana-Champaign, 1995
David C. Wilkins, Advisor

The research presented here focuses on modeling machine-learning performance. The thesis introduces Seer, a system that generates empirical observations of classification-learning performance and then uses those observations to create statistical models. The models can be used to predict the number of training examples needed to achieve a desired level and the maximum accuracy possible given an unlimited number of training examples. Seer advances the state of the art with 1) models that embody the best constraints for classification learning and most useful parameters, 2) algorithms that efficiently find maximum-likelihood models, and 3) a demonstration on real-world data from three domains of a practicable application of such modeling.

The first part of the thesis gives an overview of the requirements for a good maximum-likelihood model of classification-learning performance. Next, reasonable design choices for such models are explored. Selection among such models is a task of nonlinear programming, but by exploiting appropriate problem constraints, the task is reduced to a nonlinear regression task that can be solved with an efficient iterative algorithm. The latter part of the thesis describes almost 100 experiments in the domains of soybean disease, heart disease, and audiological problems. The tests show that Seer is excellent at characterizing learning-performance and that it seems to be as good as possible at predicting learning performance. Finally, recommendations for choosing a regression model for a particular situation are made and directions for further research are identified.

# Acknowledgments

Thanks to Nanci, my wife, for encouragement, support, patience, and love. Thanks to Benjamin, my son, for an extra boost of motivation. Thanks, also, to family and friends for their encouragement.

Thanks to my thesis advisor David C. Wilkins for his guidance. Thanks to James Edstrom, Ziad Najem, and Gunner Blix for reviewing drafts of this thesis. Thanks also to my thesis committee.

This research was conducted at the Beckman Institute for Advanced Science and Technology at the University of Illinois at Urbana-Champaign. Support was provided by the Fannie and John Hertz Foundation. This research was also supported in part by AFOSR grant F49260-92-J-0545, ONR grant N00014-88-K-0124, and ONR grant N00014-94-1-0432.

# Table of Contents

# 1. Introduction

Inductive learning methods now routinely create the initial knowledge base of classification expert systems from a library of classified examples [Gervarter, 1987]. Researchers have developed effective inductive learning methods for a variety of classification expert system representations. These include:

- the back propagation method for artificial neural networks [Hinton, 1989], the ID3 and PLS algorithms for decision trees [Quinlan, 1986; Rendell, 1986],
- the AQ15 and C4.5 methods for production rules [Michalski, 1986; Quinlan, 1992],
- parameter adjustment for Perceptrons [Minsky and Papert, 1988],
- the genetic algorithm for Holland classifiers [Booker, 1989],
- the Protos method for case-based systems [Porter, 1990], and
- algorithms for belief networks [Cooper, 1992].

For the expert system developer using these inductive approaches, the major effort in terms of time and cost is usually the creation of the library of classified cases. In domains such as equipment and medical diagnosis the data collection and encoding of a single case can take days of effort and considerable cost. Regression on learning performance as a tool in the creation of classified examples offers these benefits: regression models can tell whether the induction program can achieve a required level of classification accuracy in a particular domain. This is important because the classification level achievable by induction varies greatly from problem domain to problem domain. Shavlik *et al.* [1991] and Weiss [1991] reported variation ranging from 55% to 100%. If a developer requires a level of classification accuracy for a particular domain and this level is achievable by induction, regression models can also specify the approximate number of classified cases that the developer must collect to achieve the desired accuracy.

This thesis presents Seer, a system that generates observations of inductive-classification-learning performance and then uses those observations to create models of learning performance. Compared to earlier systems, Seer offers better and more useful models of learning performance, the ability to efficiently find the *maximum-likelihood* model, and validation with a practical inductive learning system (C4.5) on the real-world diagnosis domains of soybean disease, heart disease, and audiological proems. In a broader sense, regression is among the most important tools in experimental science. It allows fields as diverse as biometrics and econometrics make strong scientific conclusions based on the results of empirical observation. Seer shows how regression can be a practical analysis tool for the field of empirical machine learning.

The rest of this chapter offers a brief overview of the important ideas of the thesis. It defines inductive classification learning task and the task of regression on learning-performance data. It then lists the goals of this research. Finally, it previews the other chapters of this thesis.

## 1.1. Inductive Classification Learning

Inductive classification learning is the type of learning that Seer models and about which it makes predictions. An instance of (inductive classification) learning has 6 components:

- An *example space*, *E*, of examples, $e_1$, $e_2$, $e_3$, ... described in terms of attributes. For example, here is an *E* and an $e_1$ (from a made-up domain):

$$E = \{0.0 \ldots 10.0\} \times \{red,blue,green\} \times \{true,false\}$$

$$e_i = \langle 3.4,\text{blue},\text{true} \rangle$$

- A *class space*, *C*. For example:

$$C = \{\text{star}, \text{galaxy}\}$$

- A *target* function , $T\!: E \rightarrow C$ , drawn from a space of possible targets

```
If  2.3 < brightness < 4.0 and
    color in {red, blue}
      then if symmetric then dangerous
                             else normal
      else normal
```

- A set of classified *training examples* in *E*×C drawn independently according to some fixed by unknown probability distribution *D*

$$\{\langle\langle 3.4, \text{blue}, \text{true}\rangle, \text{dangerous}\rangle \ldots, \langle\langle 9.1, \text{red}, \text{true}\rangle, \text{normal}\rangle\}$$

- A *classification rule* (or hypothesis), $H: E \rightarrow C$ produced by a learning program (e.g. C4.5) from the training examples

```
If color in {blue}
      then if brightness < 5.0 then dangerous
                                   else normal
      else normal
```

- A set of labeled *testing examples* drawn according to *D*

The *accuracy* of a classification rule is the fraction of testing examples it classifies the same as the target.

## 1.2. Regression on Inductive Learning

The problem addressed by Seer is modeling and predicting learning performance based on learning-performance data. The input to Seer has two parts: a set of classified examples and an inductive learning program, such as C4.5. Figure 1.1 gives an example of classified examples from the domain of audiological problems.

```
t,mild,t,?, ... ,p3,mixed_cochlear_age_fixation
t,mild,t,?, ... ,p4,mixed_cochlear_age_otitis_media
f,mild,f,normal, ... ,p8,cochlear_unknown
...
t,normal,f,elevated, ... ,t26,cochlear_age_and_noise
```

Figure 1.1: Classified examples from the domain of audiological problems -- Such examples are part of the input to Seer. Each line is an example. The last value in the line is the example's classification. Chapter 5 gives more details of the Audiological and other real-world domains used in this thesis.

The output of Seer is a model of learning performance that can questions such as:

1. How many examples like these would the learner need to achieve 78.8% accuracy?

2. What accuracy would be possible if unlimited examples were available?

3. Factoring out the effects of noise, skewed classes, and multiple classes, what Vapnik-Chervonenkis dimension would produce worst-case learning performance most like the performance observed?

4. How would stratified sampling (changing the proportion of examples of each class) affect learning?

Seer's solution approach has two parts. First, it uses the classified examples to generate learning-performance data. Figure 1.2 shows a plot of learning-performance data. Each dot in the plot shows the result of one run of the learning program. For each run, Seer splits the classified examples into three disjoint sets: training examples, testing examples, and a possibly empty set of unused examples. The learning program is run on the training examples and the classification rule produced is tested on the testing examples. In the plot, the *x*-axis shows the number of training examples. The *y*-axis shows the accuracy of the classification rule on the testing examples.

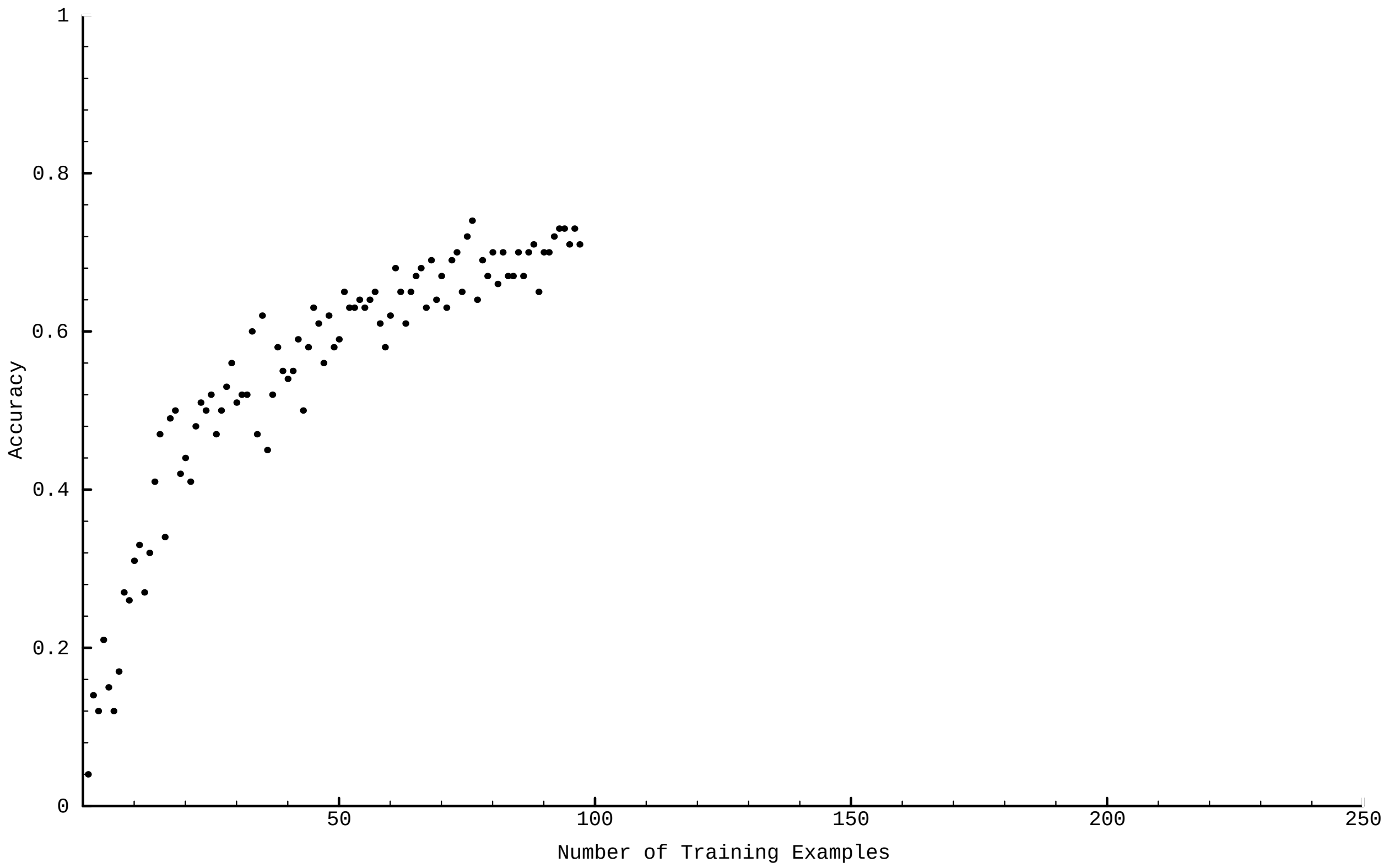

Figure 1.2: Learning-performance data -- Each dot shows the accuracy of a classification rule created with the number of training examples specified by the *x*-axis.

The second part of Seer's solution approach is to find a good fitting model for the learning-performance data. Figure 1.3 shows the model that Seer creates for the learning performance data of Figure 1.2.. The thick line shows the deterministic part of the model. A nondeterministic part of the model, not shown, predicts how far the "accuracy dots" will scatter from the thick line. The dotted line is the model's prediction of the largest possible accuracy.

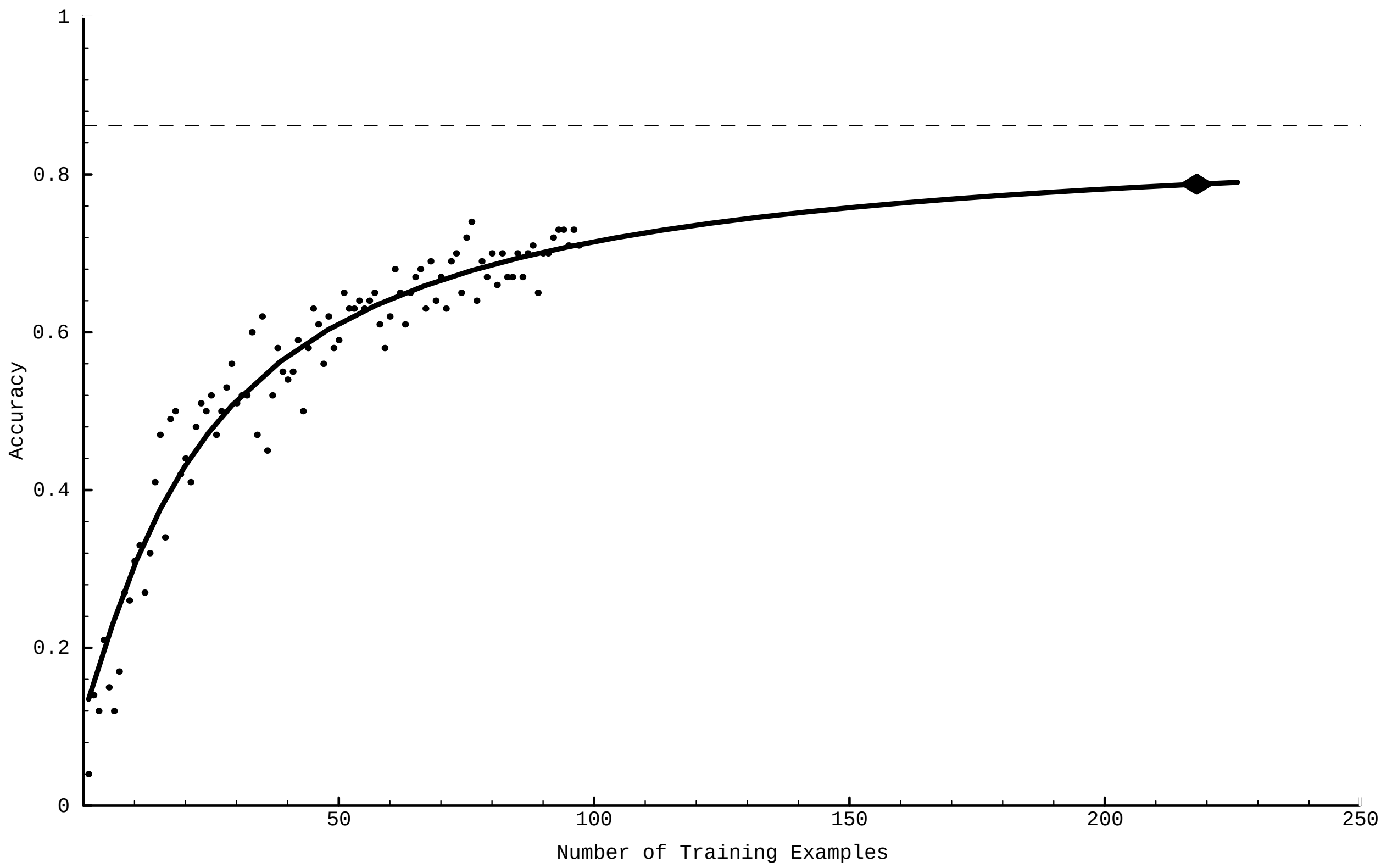

Figure 1.3: The thick curve shows the deterministic part of the learning-performance model created by Seer. The dotted line is the model's prediction of the largest achievable accuracy. The model predicts that the expected number of examples needed. The diamond shows that the actual accuracy achieved with 218 examples is very close to the 78.8% that the model predicted.

From the model, Seer can predict answers to the questions posed at the beginning of this section. (See Chapter 5 for representative examples of Seer's predictions.)

1. How many examples like these would the learner need to achieve 78.8% accuracy?

The model predicts that with 218 examples the learner would achieve 78.8%. The diamond in Figure 1.3 shows the actual accuracy achieved with 218 examples. It is very close to what is predicted.

2. What accuracy would be possible if unlimited examples were available?

The model predicts a maximum accuracy of 86.2%.

3. Factoring out the effects of noise, skewed classes, and multiple classes, what Vapnik-Chervonenkis dimension would produce worst-case learning performance most like the performance observed?

Seer can create a model that predicting that a learning task with a VC dimension $d$=0.559 would have a worst-case learning performance similar to the performance observed. (See Chapter 4 for details of how the VC dimension is generalized to real-number values and to values less than 1.)

4. How would stratified sampling (changing the proportion of examples of each class) affect learning?

The model's *start* parameter has value 0.149 and its *skew* parameter has value 1.191. Chapter 4 details how these parameters model the effect of class frequencies on learning performance.

## 1.3. Overview of Thesis

The thesis is organized as follows. Chapter 2 describes previous work in three areas. Its first section reviews theoretical work on learning performance. Its second section discusses work characterizing the form of learning curves. The third discusses previous work on the effect of skew and multiple classes on learning performance. Chapter 3 gives an overview of learning-performance models. It defines them and argues that maximum-likelihood is the most appropriate criterion by which to select among them. It shows why popular regression methods, such as ordinary linear regression, are inappropriate for learning data and it demonstrates how to generate learning-performance data from a relatively small set of classified examples. Chapter 4 treats the selection of a set of candidate models as a design problem and enumerates some possible design choices for both the deterministic and nondeterministic part of the models. The chapter also develops a heuristic model of the effect of noise, skewed classes and multiple classes on learning. Finally, it shows how to reduce the difficult problem of finding the best model from a set of candidates to a problem of nonlinear regression that can be solved with an efficient iterative algorithm. In Chapter 5, the models are put to the test in a series of almost 100 experiments on real data from three diagnosis domains: soybean disease, heart disease, and audiological problem. The chapter also provides an analysis of the experiments. The thesis concludes with Chapter 6, which provides a summary and suggestions for future work.

# 2. Related Work

Researchers have been creating models of learning performance since at least 1919 [Thorstone, 1919]. This chapter divides previous work into two approaches. Section 2.1 describes theoretical approaches, concentrating on computational learning theory. Section 2.2 covers empirical approaches and includes a table summarizing previous curve-fitting approaches and showing how Seer advances the state of the art. Finally, Section 2.3 describes specific previous work related to predicting the effect of multiple classes and skewed classes on learning.

## 2.1. Theoretical Approaches: Computational Learning Theory

Work in computational learning theory (CLT) goes back at least to Gold [1967], but the work most relevant to this thesis started with Valiant [1984]. In that paper, Valiant introduced what is now called probably-approximately-correct (PAC) learning. Roughly, a set of concepts is said to be PAC learnable if any concept in the set can be learned, with probability 1-δ, to within error less than ε, in time (and training-example set size) polynomial in 1/δ and 1/ε.

Blumer *et al*. [1989] refined Valiant's approach by introducing Vapnik-Chervonenkis (VC) dimension analysis. VC analysis can be used to measure the expressibility of some representation schemes and to show if the schemes are PAC learnable. In addition to showing what is learning in polynomial time (and with a polynomial number of examples), computational learning theory is also used to determine worst-case learning performance bounds. Shawe-Taylor *et al.* [1993] give this upper bound on the number of training examples needed to learn any concept in a concept set with VC dimension, *d* (>=2):

$$m = \mathrm{m}_{\mathrm{VC}}(\varepsilon, \delta, d) = \frac{\log(\frac{d}{(-1+d)\delta}) + 2d\log(\frac{6}{\varepsilon})}{(1-\sqrt{\varepsilon})\varepsilon} \tag{2.1}$$

For practical problems, such as estimating the number of additional classified learning examples needed to achieve a classification rule with some desired accuracy, computational learning theory is of limited utility. One limitation is that, to date, CLT only has results for classification rules with simple representations, such as 3-term conjunctive normal form functions. Developers cannot easily use computational learning theory on machine learning's standard representations such as the ID3 decision tree and the AQ15 rule representations, although research is proceeding in this direction. Another limitation is that most computational learning theory analyzes only worst-case problems and classified-example distributions; the resulting learning-performance curves are very difference from the average-case analysis needed to predict learning performance within the context of a real-world application. A third limitation is that the Vapnik-Chervonenkis dimension approach used by computational learning theory cannot quantify the effect of noise, and essentially all real-world domains to an inductive learning program will have some degree of noise. Finally, computational learning theory generally analyzes only consistent learning algorithms, algorithms that produce classification rules that are consistent with all training examples. They ignore the less-than-consistent algorithms used for real-world problems. A less-than-consistent algorithm employs a stronger inductive bias to guide the search for a generalization of the data, such as heuristically trying to choose the generalization with the fewest numbers of disjunctions.

With time some or all of these limitations might be overcome, and there are some hopeful signs of progress. For example, the works of Kearns and Schapire [1989] and Goldman and Sloan [1992] deal with worst-case learning in the presence of noise. The works of Pazzani and Sarrett [1990], Hirschberg and Pazzani [1991], Iba and Langley [1992], and Langley *et al*. [1992] describe methods that produce average-case analysis for some simple induction algorithms. The work of Ehrenfeucht and Haussler deals with worst-case learning of a less heuristic variant of the ID3 algorithm [Ehrenfeucht and Haussler, 1989]. To date, these techniques are not suitable for everyday use. For example, Iba and Langley [1992] perform an average-case analysis of a simple machine learning algorithm that creates one-level "decision trees". The resultant learning-performance model is a complex combinatorial expression that could take more computer time to evaluate that would be taken to actually run a learning expression [Langley, personal communications, 1994]. This eliminates any advantage that analytic methods might have over empirical approaches.

## 2.2. Empirical Approaches

In contrast to the theoretical approaches that start with some model of the learning task and then predict (or bounds) learning performance, empirical approaches start with observations of learning performance. Sometimes these observations are used to select a model from a set of models; sometimes they stand on their own.

### 2.2.1. Learning Performance Analysis Without Model Selection

The work of Gaines [1989] is an empirical exploration of how the *quality* of training examples created by an expert (called the expert's case library) affects the number of training examples needed create a satisfactory set of rules via induction.

Gaines used the INDUCT knowledge acquisition tool in the domain of diagnoses of contact lens. He found that INDUCT created satisfactory rules when INDUCT was supplied with 18 "critical cases." This number of needed cases, however, jumped to 90 cases when INDUCT was supplied with a representative sample of "merely correct cases." In addition, INDUCT could still create satisfactory rules even if these correct expert cases contained 25% errors; however, it then required 326 cases. Moreover, the number of cases required jumped to 1970 cases when the cases contained 10% error and 1 irrelevant attribute on the average. Gaines' research results demonstrate how the quality of the data dramatically affects the number of library cases required for induction.

Note that the method used by Gaines is empirical in nature. The determination of the effect of a particular amount of noise, irrelevant attributes and the like was calculated by running more and more cases under a particular set of conditions until a satisfactory set of rules was obtained.

Seer also gathers observations of learning performance, but in addition it has a set of candidate models—for example, a family of curves. It selects the model from the model set that best fits the data. If that data is representative and the model set is appropriate, the resulting model may be useful for predicting future learning performance over a wider range conditions that was observed. For example, Seer might be able to predict that approximately 2000 cases are required when the cases contain 10% errors, without having to construct and run 2000 cases with these characteristics.

### 2.2.2. Learning Performance Analysis With Model Selection

This section reviews 75 years of research analyzing learning-performance data by fitting models to the data and highlighting Seer's unique contributions.

Table 2.1 summarizes the section's conclusions.

Over the years researchers in many fields have fit many curves (sets of candidate learning-performance models) to learning-performance data from many kinds of learning. The most popular candidate curves have been on these orders:

$$\text{Exponential: } \log(\varepsilon) = m \tag{2.2}$$

$$\text{Hyperbolic: } \varepsilon = 1/m \tag{2.3}$$

$$\text{Power Law : } \varepsilon = 1/m^{x} \tag{2.4}$$

$$\text{Logit: } \text{logit}[\varepsilon] = m \tag{2.5}$$

where $m$ is the number of training examples and ε is some measure of error or reaction time., The most popular kinds of learning considered have been skill acquisition (also called speed-up learning) and classification learning. A rat learning to run a maze faster is an example of skill learning. The time to complete the maze would be ε.

Quantitative modeling of learning was first considered in the field of experimental psychology. In Thorstone's 1919 analysis of 40 curve shapes corresponding to data derived from humans learning to use a typewriter, he concluded that the hyperbolic curve fit best. . Mazur and Hastie [1978] reviewed of 75 years of debate in the psychological community about whether learning curves are exponential or hyperbolic. They then put these two curve families to the test on a word-recall task and concluded that the hyperbolic fits the data better.

| **Summary** | T19 | M72 | N81 | A89 | K91 | C92 | I92 | A93 | S93 | Seer |
|---|---|---|---|---|---|---|---|---|---|---|
| Applied to classification learning [1] | N | N | N | N | Y | Y | Y | Y | Y | Y |
| Applied to practical machine learners?[2] | N | N | N | N | Y | Y | N | Y | Y | Y |
| Put to practical use? | N | N | N | N | N | N | N | N | N | Y |
| Used for accuracy prediction? (not just testing fit) | N | N | N | N | N | N | Y | N | N | Y |
| Orders considered (e.g. log, $1/m^x$,1/m)d | 1/m and others | log, 1/m | log, $1/m^x$, 1/m | 1/m, $1/m^x$ | 1/m | 1/m, log(e) | [3] | 1/m | all | 1/m, $1/m^x$, logit |
| Uses VC bounds? | N | N | N | N | Y | Y | N | N | Y | Y |
| Uses tightest VC bounds? | N/A | N/A | N/A | N/A | N | N | N/A | N/A | Y | Y |
| Models noisy concepts? | Y | Y | Y | N/A | N | N | Y | N | N | Y |
| Applied empirically to learning performance data? | Y | Y | Y | Y | Y | Y | Y | N | N | Y |
| Applied to learning on "natural" examples | Y | Y | Y | Y | N | N | N | N/A | N | Y |
| Applied to binary-response data?[1] | Y | Y | N | Y | Y | Y | Y | N/A | N/A | Y |
| Finds maximum-likelihood models | N | N | N/A | N | N | N | N/A | N/A | N/A | Y |
| Uses a discrete distribution? | N | N | N/A | N | N | N | Y | N/A | N/A | Y |
| Binomial distribution? | N/A | N/A | N/A | N/A | N/A | N/A | N/A | N/A | N/A | Y |
| Beta-binomial distribution? | N/A | N/A | N/A | N/A | N/A | N/A | N/A | N/A | N/A | Y |

Notes:

*[1] Other kinds of learning are skill learning (where reaction time is measured) and pair-recall learning where the ability to recall a word indexed by another word is tested. Reaction times are continuous, not binary. Pair-recall learning, like classification learning, has a binary response.*

*[2] A practical machine learner is defined here as a learner stronger than linear perceptron.*

*[3] The expression derived is a complex combinatorial expression that can take longer to evaluate on a computer than running the learning experiment it is designed to predict.*

| **KEY** | |
|---|---|
| *I92* | *Iba and Langley 1992* |
| *A89* | *Anderson 1989* |
| *A93* | *Amari 1993* |
| *C92* | *Cohen 1992* |
| *K89* | *Kadie 1989.* |
| *K91* | *Kadie 1991* |
| *M72* | *Mazur and Hastie 1972* |
| *N81* | *Newell & Rosenbloom 1981* |
| *T19* | *L. L. Thurstone 1919* |
| *S93* | *Shaw-Taylor 1993* |
| *Seer* | *This thesis* |

Table 2.1: Summary of previous work on the shape of learning curves

Newell & Rosenbloom [1981] bridged the gap between psychology and artificial intelligence. Concentrating on skill learning, they reviewed work with skill-learning curves back to 1926. They also tested the applicability of the Power Law to a dozen skill-acquisition domains, concluding:

> "There exists a ubiquitous quantitative law of practice: It appears to follow a power law`; that is, plotting the logarithm of the time to perform a task against the logarithm of the trial number always yields a straight line, more or less. We shall refer to this law variously as the log-log linear learning law or the power law of practice."

Anderson [1989] showed that the Power Law of Learning could be a consequence of "rational" memory retrieval. He also referred to other models of memory and skill learning that are consistent with the Power Law of Learning. Logan [1992] gave a useful overview of Power Law research on skill acquisition since 1981.

Kadie [1991] analyzed constructive-inductive classification learning by fitting a model/curve based on the VC dimension bound to learning-performance data. The least squares fit resulted in an $R^2$ of greater than 0.99

Amari [1993] reviewed "1/*m* results" from the fields of general stochastic descent dynamics, computational learning theory, statistical mechanics, information theory, and Bayesian statistics. Cohen [1992] reviewed "1/*m*" computational learning theory work. Cohen fit linear perceptron and a more complex artificial neural net to synthetic data from artificial targets and found some synthetic problems in which the exponential models fit better and some where the hyperbolic models fit better.

This thesis does not examine speed-up learning. Its general goal is to provide the field of empirical machine learning with a practical and useful regression tool for *classification learning*. It extends the work of Kadie [1991] and the others' work in three ways.

**Learning-performance models: Best constraints and most useful parameters --** We will see in Chapter 5 that all models of the same order (e.g., hyperbolic) fit the (classification) learning-performance data about equally well. What differentiates models is not their general shape but the details of their constraints and parameters. For example, unlike Seer, many learning models do not capture the constraint that classification accuracy is bounded by 0.0 and 1.0. Such models can predict expected accuracies less than 0 for low *m* values. Also, many models treat variance as continuously distributed. Such models can erroneously predict that it is possible to achieve 101% classification accuracy. Chapter 4 will show that Seer correctly models variance as a discrete distribution and that it can distinguish between variance caused by variation in the training examples from variance cause by variation in the testing examples. Like most of the models, Seer has a parameter, *max*, that models noise. Unlike the other models, it also has parameters, *start* and *skew*, that directly measure (and can be used to predict) the effect of multiple classes and skewed classes. Also, unlike other models, its measure of learning difficulty, *d*, can be calibrated to correspond to the Vapnik-Chervonenkis dimension from computational learning theory.

**Fitting criteria and algorithm: Finds maximum-likelihood models efficiently --** As the quotation from Newell & Rosenbloom [1981] above suggests, many systems do curve fitting by first transforming the data and then applying linear regression. This introduces two problems. First, the procedure can fail. For example, if a learner's imperfect classification rule just happens to correctly classify 5 of 5 examples, the observed error is 0. The fact that the logarithm of 0 is undefined results in the failure of the procedure. The second problem is that a least-squares fit does not find the maximum-likelihood model for classification learning. Even if nonlinear regression is used rather than linear regression, least-squares fit only finds the maximum-likelihood model if the data is continuous and the variance is constant. Neither assumption holds for (classification) learning-performance data. (See Section 3.1.) Seer works directly on the data (no transformations), finds the maximum-likelihood model directly, and makes the reasonable assumption that variance, rather than being constant, is determined by the binomial or similar discrete distribution. This would seem to make Seer's fitting task one of difficult nonlinear programming, but as Section 4.4 details, Seer is able to reduce the problem to nonlinear regression and which it can solve with a quickly converging iterative algorithm.

**Application and experimental validation: Demonstrates a practical application on real-world data --** It is the author's belief that Seer shows the first practical application for learning-performance modeling of machine learning data, namely, the estimation of the number of training examples needed to develop a classification expert system via induction. Related to this, it is one of the few efforts used to make *predictions*, not just characterizations. Although several other efforts have applied learning-performance modeling to noisy learning situations, Seer is the first (within the author's knowledge) to use noisy data produced by a practical machine learning system such as C4.5. Experimentally, Seer provides (to the best of the author's knowledge), the first learning-performance modeling of a machine learning algorithm on "natural" learning examples (See Chapter 5). The other previous machine learning work has either been entirely theoretical (and not applied to any empirical data) or has been applied only to synthetic learning examples.

## 2.3. Effect of Skew and Multiple Classes on Learning Performance

One of the contributions of this thesis is an analysis of how skewed classes and multiple classes affect learning performance and the learning curve. The analysis shows how class frequency information constrains the form of the curve. Previous work on the effect of class frequencies developed from a statistical design method called *stratification* [Brewer, 1982].

In some statistical problems, examples can be chosen by class. By way of illustration, a statistician may decide to poll exactly 25 Libertarians and exactly 25 Republicans even if Libertarians are less frequent than Republicans in the population being studied. Such a design would allow the statistician to make better conclusions about Libertarians and about the differences between the two groups while keeping cost constant relative to random sampling.

In inductive machine learning, work with skewed classes has concentrated on adapting algorithms designed for even class frequencies to work well with skewed class frequencies. Buntine [1989] created a decision tree learner that could be given class frequency information. Catlett [1991] performed a series of experiments on the effect of reducing the number of examples in the most frequent class (so that the learning algorithm would run faster). Although he concluded this is a useful tool for some situations, he believed that throwing away training examples (even from the most frequent class) is not a general solution to the problem of scaling to large data sets.

Quinlan [1991] documented some of the difficulty of learning with two skewed classes and shows how machine learners can minimize some of the difficulty by selectively adopting a “when in doubt, guess the most frequent class” policy.

Seer provides an explicit and predictive empirical model of the effects that previous work has observed.

## 2.4. Summary

Theoretical approaches of learning-performance modeling can be used, if enough is known about the learning task, to predict bounds on learning performance. However, for practical problems with noise and that use the best machine learning algorithms, theory cannot yet predict average-case performance. Empirical approaches try to gather additional information, usually about a more narrow learning task, by observing actual learning performance. This information can then be generalized into quantitative models. Seer does this by fitting a curve (a learning-performance model) to observed learning-performance data. It advances the state of the art with: 1) learning-performance models that embody the best constraints (for classification learning) and most useful parameters 2) fitting algorithms that efficiently find maximum-likelihood models, and 3) a demonstration, on real-world data, of a practical application.

# 3. Overview of Learning-Performance Models

Our ultimate goal is to create a statistical model from observations of learning performance that can predict future learning performance. This chapter addresses two prerequisite questions. First, given enough learning-performance data and some candidate models of learning performance, how do we evaluate the fit of those models on that data? Second, how can we generate enough learning-performance data from a relatively small number of classified examples?

For both questions, assume learning-performance data is a set of tuples. Specifically, let $L_1, L_2, \cdots, L_w$ represent *w* runs of the learning program. For each run of the learning program a tuple $<m_i, y_i, k_i>$ is recorded, where $m_i$ is the number of training cases given to the inductive learning program, $y_i$ is the number of testing cases the classification rule classifies correctly, and $k_i$ is the number of cases used to test the classification rule created by the learning program.. Figure 3.1 shows an example plot of $m_i$ vs. $y_i/k_i$.

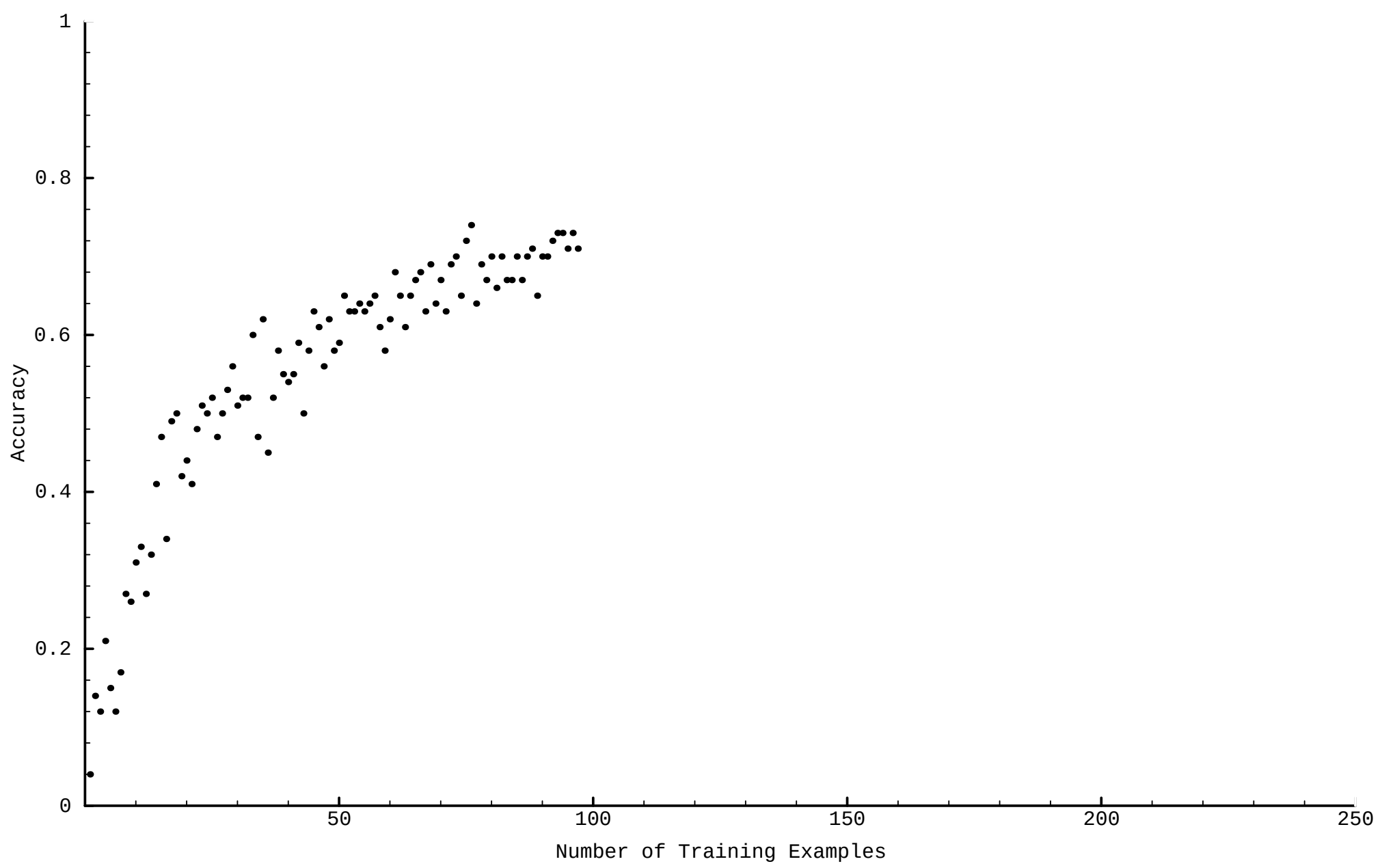

Figure 3.1: Learning-performance data. The horizontal axis is $m_i$, the number of training examples. The vertical axis is $y_i/k_i$, the fraction of testing examples the learner's classification rule correctly classifies.

## 3.1. Good-Fitting Models of Learning-Performance

This section will answer the question “What is a good-fitting model of learning performance?”.

### 3.1.1. Models of Learning Performance

A model is a function from *m*, the number of training examples, to a probability density. Given a value *m*, the number of training examples, and *k*, the number of testing examples, a model predicts the probability of each possible value of *y*, the number of correctly classified examples. Regression models are often defined in two parts: a deterministic part and a nondeterministic part [Aldrich and Nelson, 1984; McClave and Dietrich, 1988]. For example, in the model in Figure 3.2a, the deterministic part of the model has the form of a line $z = \mathrm{a} + \mathrm{b}m$, and the nondeterministic part of the model is a normal distribution of fixed variance centered on *z*. This is the kind of model found by ordinary linear regression. Figure 3.2b illustrates another model. Again, there is a deterministic part, in this case taking the form of a curve, and a nondeterministic part, in this case taking the form of a discrete probability density. Chapter 4 will discuss specific candidates for the curve and the density. For now, our focus is on what a model does.

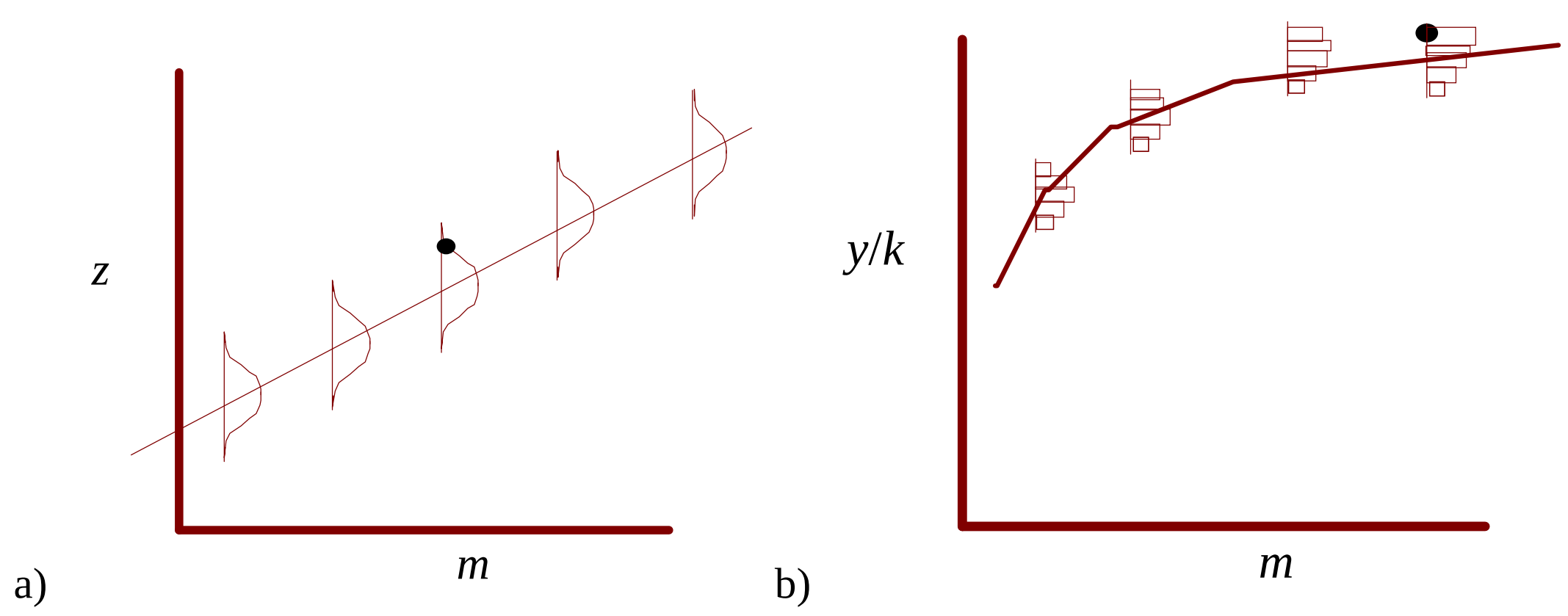

Figure 3.2: Two statistical models

### 3.1.2. Regression Methods

Regression is well established as a field of statistics and includes many specific techniques, such as *ordinary linear regression via least-squares minimization.* This section shows why continuous-response methods such as ordinary linear regression are not adequate for learning-performance data.

In situations with one real-valued dependent variable and one real-valued independent variable (that is, *continuous-response data*), one can think of ordinary linear regression as a method of finding the straight line that best characterizes a set of observations. "Best" here means the line that minimizes the sum of the squares of vertical distances between the data points and the line. Two properties of learning-performance data make it inappropriate for ordinary linear regression. First, learning curves (plots of *m* vs. *y*/*k*) are not typically linear. They range from 0% to 100% expected accuracy, typically rising quickly and then slowly converging to or toward 100%. This violates ordinary-linear regression's linearity assumption and results in predicted accuracies that are greater than 100%, such as 110%. Second, ordinary linear regression (and many other nonlinear continuous response methods) assume that errors -- that is, differences between the data and the model -- are distributed according to some constant variance. In other words, They assume that a situation where the model predicts 30% accuracy and the data shows 50% is just as likely as a situation where the model predicts 99% accuracy and the data shows 79% accuracy. In practice, the variance of the error shrinks significantly for learning-performance data as the accuracy gets closer to 100%.

These problems can sometimes be overcome by transforming the data so that it has more appropriate properties. For example, averaging observed accuracy and then transforming it with the $p' = \log(1-p)$ function [or the function $p' = 1/(1-p)$ ] will tend to linearize its relation with the number of training examples, *m*. Such a transformation will also tend to make the variance of the errors more constant.

A problem remains, however, for all regression methods that assume learning-performance data is continuous-response, for example, by assuming that errors are normally distributed. Learning-performance data is *binary-response data*, not continuous-response data; on each classification test case, the classifier will either have an accuracy of 0 or an accuracy of 1. If p(*m*) is the accuracy predicted by the model for a learning classification rule based on m training examples, the error on a single testing example will, with probability p(*m*), be 1-p(*m*) and will, with probability, 1-p(*m*), be 0-p(*m*). Thus, the deviance between prediction and actual result will at best have a mean of zero and a variance equal to p(*m*)[1-p(*m*)]. In other words, it is distributed discretely according to the binomial (or some similar distribution), not continuously. Although the binomial distribution can sometimes be approximated with the normal distribution, that approximation is not accurate if the sample size is small or if p(*m*) is near 1. Because small samples and accuracies near 1 are of special interest to us when predicting learning performance, we should not use continuous-response regression on learning-performance data.

An alternative to continuous-response regression is binary-response regression. For some types of continuous-response regression, such as ordinary linear regression, finding a least-squares fit produces the model with *maximum likelihood.* This is not true for binary-response regression so that it is usual to use other, usually more direct, but less efficient, methods for finding maximum-likelihood models [Cramer, 1986; Hosmer and Lemeshow, 1989]. Chapter 4 demonstrates Seer's use of nonlinear optimization to this end. The next section discuses maximum likelihood further.

### 3.1.3. Maximizing Likelihood

If our general goal is to find the most probable model from a set $M_i \in \{M_1, M_2, \ldots\}$, given some observed data *D*.:

$$\text{Find } M_i \in \{M_1, M_2, \ldots\} \text{ that maximizes } \Pr(M_i | D) \tag{3.1}$$

Then by Bayes' rule, this is the same task as:

$$\text{Find } M_i \in \{M_1, M_2, \ldots\} \text{ that maximizes } \frac{\Pr(M_i)\Pr(D|M_i)}{\Pr(D)} \tag{3.2}$$

If the number of $M_i$'s is finite, all $M_i$'s equiprobable (and that D is constant), this is the same task as:

$$\text{Find } M_i \in \{M_1, M_2, \ldots\} \text{ that maximizes } \Pr(D| M_i) \tag{3.3}$$

This assumption is typical but usually implicit in the field of statistics [Cramer, 1986, p. 7]. It is the reasonable given that no other knowledge is available. The result is called the *maximum-likelihood model*.

Consider a simple example of computing the likelihood of a model given some learning-performance data. Suppose the learning performance data is

$$\{<100, 1, 1>, <100, 0, 0>, <150, 0, 1>, <150, 1, 1>, <200, 1, 1>, <200, 1, 1>\}. \quad (3.4)$$

Suppose furthermore one of our candidate models predicts an expected accuracy of 0.70 for classification rules based on 100 training examples, an expected accuracy of 0.80 for classification rules based on 150 training examples, and an expected accuracy of 0.90 for classification rules based on 200 training examples. Table 3.1 shows the likelihood calculation. Note that maximizing likelihood is equivalent to minimizing the log likelihood; this fact is used to simplify some calculations..

| Learning Performance Datum | Model’s Predicted Expected Accuracy | Likelihood | Log (natural) Likelihood |
|---|---|---|---|
| <100, 1, 1> | 0.70 | 0.70 | -0.36 |
| <100, 0, 1> | 0.70 | 0.30 | -1.20 |
| <150, 0, 1> | 0.80 | 0.20 | -1.61 |
| <150, 1, 1> | 0.80 | 0.80 | -0.22 |
| <200, 1, 1> | 0.90 | 0.90 | -0.11 |
| <200, 1, 1> | 0.90 | 0.90 | -0.11 |
| | Total: | 0.027 | -3.60. |

Table 3.1. An example of computing the likelihood of a model

## 3.2. Generalized Cross-Validation

In many applications, a small, fixed set of examples (a case library) must supply both training examples for the inductive learning program and testing examples for measuring the performance of the classification rule the program produces. This section tells how to collect learning performance data from a small number of classified examples with a heuristic called *generalized cross-validation*.

The standard statistical technique of cross-validation (see Figure 3.3) would try to solve the problem by dividing the $|N|$ classified examples into $v$ disjoint subsets, each of approximate size $|N|/v$. Then it would repeatedly hold one subset out for testing and give the other examples in the other $v$-1 subsets to the learner for training. The result would be learning-performance data of the form:

$$\{<m_1,y_1,k_1>, <m_2,y_2,k_2>, ..., <m_v,y_v,k_v>\} \quad (3.5)$$

where the $m_i$'s are the number of training examples given to the learner (approximately $(v-1)|N|/v$). The $k_i$'s are the number of testing examples (approximately $|N|/v$). The sum of the $k_i$'s will be exactly $|N|$ because each case is used as a testing example exactly once. The $y_i$'s are the number of examples a classification rule classified correctly. They will range in value between 0 and $k_i$.

Input: A set $N$ of classified examples

Procedure:

1. Shuffle order of examples
2. Partition into subsets (say, $v$ subsets):

    $S_1$, $S_2$, ... $S_v$

3. For each subset, train on the others, test on it.

    Run 1: Train on $N$- $S_1$; Test on $S_1$

    ...

    Run $i$: Train on the set $N$- $S_i$ ; Test on $S_i$

    ...

Output: $\{<m_1,y_1, k_1>,<m_2,y_2,k_2>, ..., <m_i,y_i,k_i>,...\}$

Figure 3.3: The cross-validation procedure

As an example, if $|N|$ is 101 and $v$ is 20, then each subset will have 5 or 6 cases in it and resultant learning-performance data will look something like:

$$\{<95, y_1,6>, <96, y_2,5>, ... <96, y_{20},5>\} \quad (3.6)$$

Cross-validation is good in that it uses each case exactly once for testing. This means that the $y_i$'s are independent with respect to the testing examples (but not the training examples). Cross-validation is inappropriate for learning-performance analysis, however, because it only provides data for one or two distinct values of *m*. In the example, it only provided data for *m*=95 and *m*=96. To do learning-performance analysis, we need data for many distinct values of *m*.

To meet this need, Seer uses *generalized cross-validation,* a new heuristic. The idea is to perform cross-validation for each distinct value of *m* of interest (see Figure 3.4).

Input: A set *N* of classified examples, set of *m*-values of interest

Procedure:

For each *m:*

a. Shuffle the order of the examples

b. Partition the examples into *v* subsets:

c. For each subset, $S_i$

Train on *m* examples randomly selected without replacement from the set *M*-S*i*

Test on $S_i$

Output: $\{<m_1,y_{1,1}, k_{1,1}>,<m_1,y_{1,2},k_{1,2}>, ..., <m_i,y_{i,1},k_{i,1}>,...\}$

Figure 3.4: The generalized cross-validation procedure

For example, suppose |*N*| is 101 and that one *m*-value of interest is 45. We could divide the data into three subsets *a, b,* and *c* of size 33, 34, and 34, respectively. Next we could create three pairs of training data and testing data:

$$\{<N-c, c>, <N-b, b>, <N-a, a>\} \quad (3.7)$$

Each pair consists of 1) the union of all but one subset and 2) the remaining subset. The first element of the pair is the training set. In the example, each training set has 67 or 68 examples in it. That is more than we want, so we randomly select (without replacement) 45 examples from each training subset to actually be used for training. The process is repeated for all *m*-values of interest.

In the example, the data was divided in the three initial subsets. In general, it will be divided into $v$ subsets. The only constraint is that the number of initial subsets $v$ must be such that $\left\lfloor \frac{v-1}{v} \right\rfloor |N| > m$. In other words, it must be large enough that there will be enough training examples. In the experiments of Chapter 5, the default value of $v$ is 10. For larger $m$, it is the minimum value that meets the constraint.

With generalized cross-validation, for each distinct value of $m$, each of the $|N|$ cases is used for testing once and only once. Because we assume the $|N|$ cases are independent, learning performance data with the same $m$ value is independent (with respect to the testing examples). Learning performance data with different values of $m$, however, are not independent with respect to the testing examples because the same examples used for testing when, say, $m$=49 are reused for testing when $m$=50. Considered as a whole, the learning performance data is, thus, not mutually independent. This will prevent us from making some statistical conclusions that depend heavily on the independence of the data. For example, it will make it impossible to create good confidence intervals on predictions because such confidence intervals depend on a good estimate of variance and that requires independent data. This is unfortunate, but it is inevitable when working with a small amount of data.

## 3.3. Conclusion

This chapter gave an overview of the regression problems on which Seer works. It defined the form of learning-performance data of interest to Seer as a set of tuples, $\{<m_1,y_1,k_1>, <m_1,y_1,k_1>, ...\}$, where $m_i$ is the number of training examples given to a learning program, $k_i$ is the number of testing examples given to the learner's classification rule, and $y_i$ is the number of testing examples the classification rule classifies correctly. The chapter also defined models of learning performance. A model is a function that given the number of training examples, $m$, and the number of testing examples, $k$, a model tells the probability of correctly classifying any $y_i$ number of examples. The chapter also showed how to apply the notion (from the field of statistics) of maximum likelihood and that continuous-reposes regression methods will not finding maximum-likelihood models of the learning-performance data. Finally, the chapter showed that when the total number of classified examples is small, a new procedure called generalized cross-validation can be used to gather learning-performance data (at a cost of some statistical independence).

The next chapter gives the specifics of Seer's regression. It details the learning-performance models that Seer considers and shows how Seer uses nonlinear optimization to find maximum-likelihood models of learning performance.

# 4. Candidate Models of Learning Performance: Design and Selection Method

The previous chapter defined learning-performance models and showed how the maximum-likelihood criterion defines the best model in a possibly infinite set of candidate models. But which sets of candidate models should we consider? And how can we, algorithmically, find the best model from a set? This chapter answers these two questions. The chapter treats the first question (which sets of models to consider) as a three-part design problem: 1) designing the deterministic part of the model for noise-free learning with two equiprobable classes, 2) generalizing the deterministic model to handle multiple classes, skewed classes, and noise, and 3) designing the nondeterministic part of the model. Figure 4.1 gives an overview of the design choices considered. The chapter answers the second question (how to algorithmically find the best model from a set) in Section 4.4 which details the optimization techniques used. This chapter draws no conclusions about which designs are best. That question will be addressed in Chapter 5 by putting the designs to the test on real learning-performance data.

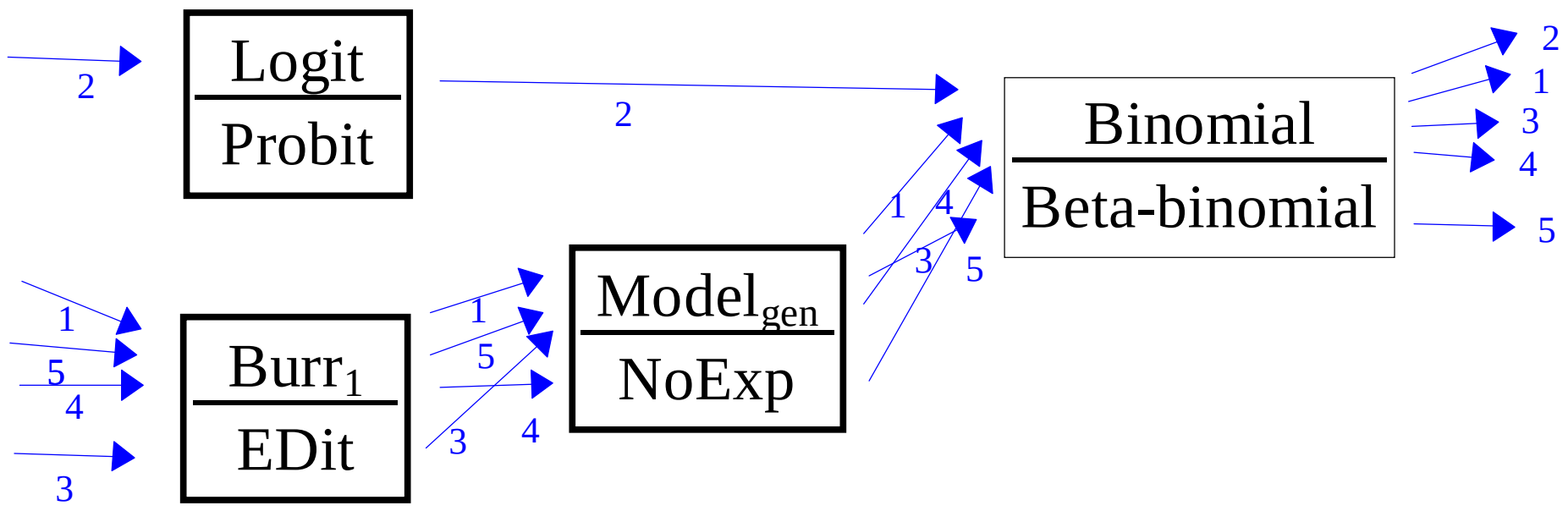

Figure 4.1: Design of sets of candidate models. We will start with sets of deterministic models based on logit, probit, Burr$_1$, and ED. Then we will add Model$_{gen}$ or NoExp (NoExp defined in the next chapter) to improve fit when classification classes are skewed, multiple, or noisy. Finally, we will add a nondeterministic component to the (sets of) models. The numbered paths show the five designs put to the test on real data in Chapter 5.

## 4.1. Candidate Deterministic Models

Recall from section 3.1.1 that the deterministic part of a learning-performance model is the part that, given the number of training examples provided to the learner, predicts the expected accuracy of the learner's hypothesis. (The nondeterministic part of the model predicts the error in the accuracy prediction.) This section finds some reasonable and some unreasonable deterministic (sets of) models. The most interesting of these, the effective-dimension learning-performance model, is inspired by models of learning developed by computational learning theory.

All the models considered will be *binary-response* models, that is; they predict number of occurrences of some binary event. In the context of learning-performance data, they predict the expected number of correctly classified examples, $y$, when the learner's classification rule is tested on $k$ examples. Put another way, they predict the expected accuracy, $y/k$, on a set of testing examples, where accuracy can range from 0 to 1.

This section will start out assuming that the learning task of interest is two-class classification (for example, distinguishing healthy people from sick people). Also, it will assume that the two classes are equally probable and that classification is noise-free. In the example, this would mean that it assumes that it will be asked to classify about the same number of healthy people as sick people and that any two people with the identical symptoms will have the same classification (either healthy or sick). Section 4.2 relaxes this assumption. It shows how to generalize a learning-performance model of learning two, equiprobable, noise-free classes into a learning-performance model of learning multiple skewed classes with noise.

### 4.1.1. Popular Binary-Response Models

A common way to approach the problem of choosing a deterministic model is to try to find a *link function*. Its inverse, a regression model, should fit the data well with few parameters (simplicity). Four candidate link functions and their corresponding regression models are summarized in Table 4.1.

| Link Function | | Corresponding Regression Model | |
|---|---|---|---|
| Name | Value | Name | Value |
| logit | z= log(p/(1-p)) | logistic | $p = \frac{e^z}{1+e^z}$ |
| probit | $z = \Phi^{-1}(p)$ | normal | $p = \Phi(z)$ |
| EDit | $z = \frac{2\log(\frac{6}{1-p})}{(1-\sqrt{1-p})(1-p)}$ | EDmodel | $p = EDit^{-1}(z)$ |
| $Burr_1$ | $z = 1\Big/\left(\frac{1}{p}-1\right)$ | BurrModel | $p = z/(z+1)$ |

Table 4.1 SEER's link functions and regression models -- Φ is the cumulative distribution function for the normal distribution, *p* is a predicted probability and *m* is the number of training examples. The variable *z is (m-1)/d*+1, where *m is* the number of training examples and *d* is a parameter.

The two most commonly used link functions for binary-response regression are *logit* and *probit.* They are based on statistical distributions [McCullagh and Nelder, 1989]. The logit link function is the inverse of the cumulative distribution function (CDF) for the logistic probability distribution. Logit's inverse is called the *logistic regression model*. It is defined as:

$$p = \frac{e^z}{1+e^z} \tag{4.1}$$

Figure 4.2a shows its plot. Because the logistic regression model is a CDF, it maps values from the range $-\infty$ to $+\infty$ to the interval 0 to 1. The probit link function's inverse is called the *normal regression model* because it is the CDF of the standard normal distribution. Figure 4.2b plots the normal regression model.

### 4.1.2. Models Inspired by Computational Learning Theory

EDit and $Burr_1$ are link functions especially designed to fit learning-performance curves. The EDit link is inspired by results in computational learning theory [Shawe-Taylor *et al.*, 1993]. The next section explains the relationship. The $Burr_1$ link, a simplification of the EDit link, is of a class of functions (the inverse polynomials) that have been used for least-squares regression [Nelder, 1966], but not (to the author's knowledge) for binary-response regression [Nelson, 1994].

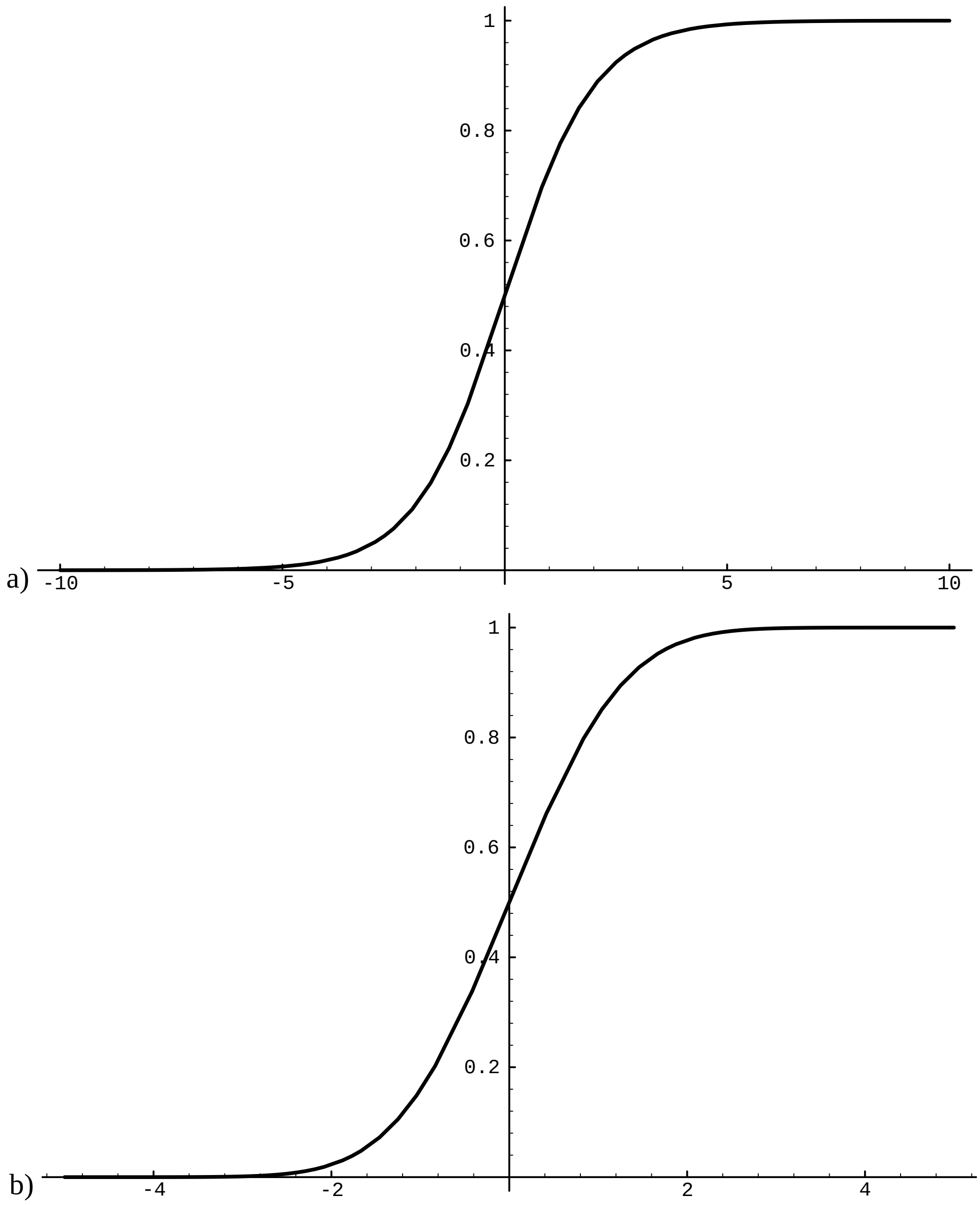

Figure 4.2: a) Logistic model: $p = \dfrac{e^x}{1+e^x}$ b) Probit (normal) model: $p = \Phi(x)$

### 4.1.2.1.The Effective-Dimension Model

Informally, the effective dimension is just a backward version of the Vapnik-Chervonenkis (VC) dimension [Blumer *et al.*, 1986]. It is a measure of complexity based on the expressiveness of a hypothesis (classification-rule) language.

Taking as input the complexity of the hypothesis space of the learning problem, a given VC dimension, *d*, defines a (worst-case) learning curve. In contrast, the effective dimension of a learning problem is defined by a learning curve. Given a set of learning-performance data, the effective dimension is the *d* that defines a curve that best fits the data. Thus, unlike the way that computational learning theory uses the VC dimension, effective dimension analysis can take advantage of existing empirical performance data of an induction algorithm to make quantitative predictions of the behavior of the learning algorithm if it was given additional cases or was subjected to different environmental conditions, such as different levels of noise, irrelevant attributes, number of hidden units, etc.

Formally, Shawe-Taylor *et al* [1993] gives this upper bound on the number of cases needed to learn any hypothesis in a hypothesis space with VC dimension, d (>=2):

$$\mathrm{m} = \mathrm{m}_{\mathrm{VC}}(\varepsilon,\delta,d) = \frac{\log\left(\frac{d}{(-1+d)\delta}\right) + 2d\log\left(\frac{6}{\varepsilon}\right)}{(1-\sqrt{\varepsilon})\varepsilon} \tag{4.2}$$

In these equations ε is the maximum allowed error (and 1 minus the minimum required accuracy). $\delta$ is the minimum probability of failure. Log is natural logarithm. An intriguing aspect of this formulation is that the VC-dimension approach offers statistical guarantees about induction which we think of as offering no guarantee.

The Shawe-Taylor *et al.* relation, as used in computational learning theory, is unsuitable as a regression model for average-case learning. It is only defined on integer *d* greater than 2, but even *d*=2 gives a slower raising learning curve than is typically seen with average-case learning. The first step to defining the EDit link function of SEER is to change the VC bound so that it is defined on all nonnegative real-valued *d*:

$$\text{for constant } \delta,\ \mathrm{x}_{\mathrm{ED}}(\varepsilon) = \lim_{d\to\infty}(\mathrm{m}_{\mathrm{VC}}(\varepsilon,\delta,d)/d) = \frac{2\log(\frac{6}{\varepsilon})}{(1-\sqrt{\varepsilon})\varepsilon} \quad (4.3)$$

Also, the function is expressed in terms of accuracy, *p*, rather than error, ε:

$$\mathrm{EDit}_0(p) = \mathrm{x}_{\mathrm{ED}}(1-p) = \frac{2\log(\frac{6}{1-p})}{(1-\sqrt{1-p})(1-p)} \quad (4.4)$$

$\mathrm{EDit}_0$ corresponds to a fixed learning curve. To allow it to be fit to data, parameters must be added. Putting in the effective dimension parameter, *d*, allows the curve to expand and compress horizontally:

$$\mathrm{EDit}_1(p,d) = d\,\mathrm{x}_{\mathrm{ED}}(1-p) = \frac{2d\log(\frac{6}{1-p})}{(1-\sqrt{1-p})(1-p)} \quad (4.5)$$

Figure 4.3 plots the original VC curve and EDit at *d*=2, the value of *d* at which they are most different. Even here they are close.

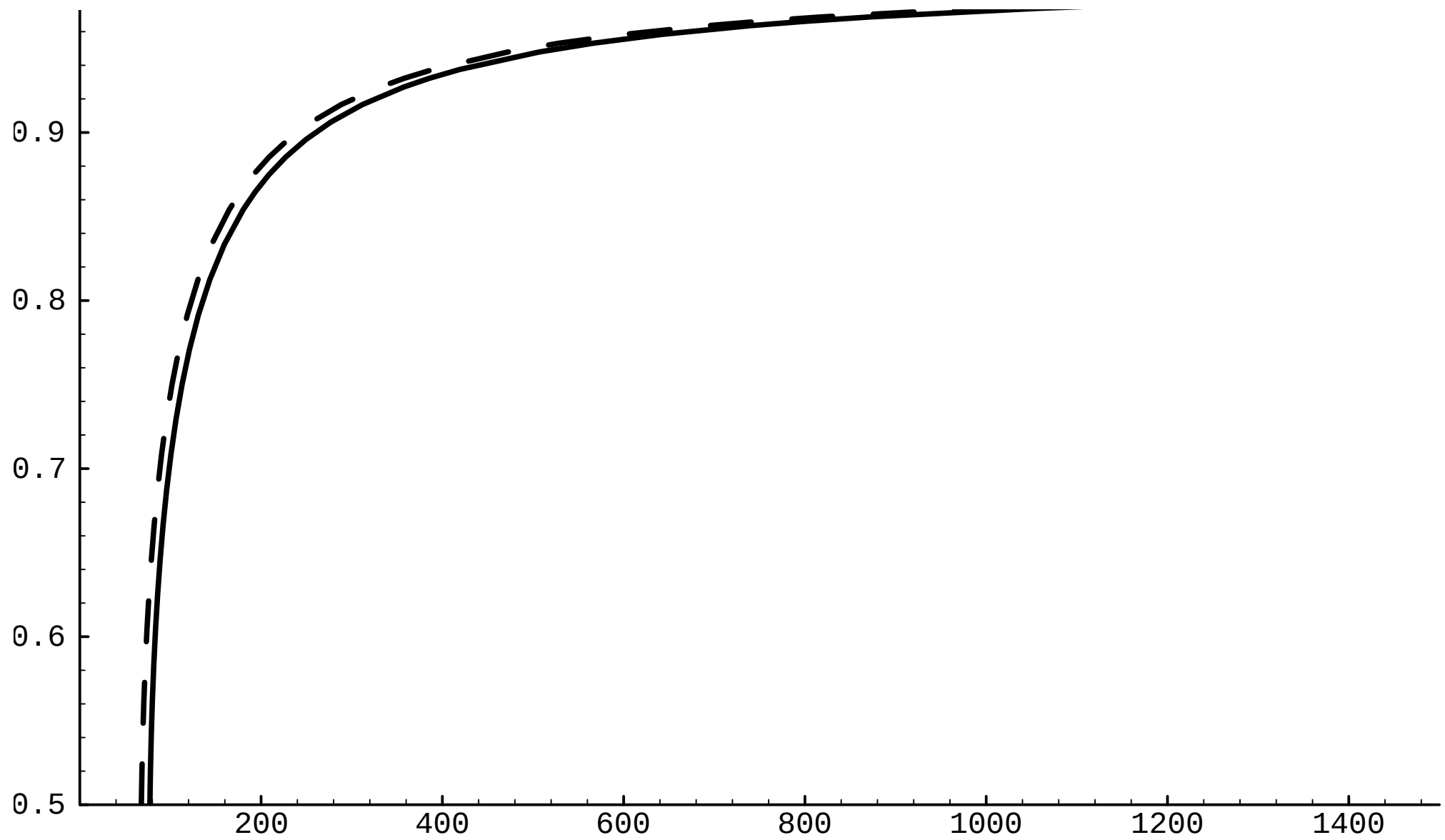

Figure 4.3: The solid line is the learning accuracy curve for the $\mathrm{m}_{\mathrm{VC}}(1-p,0.5,d)$ for *d*=2. The dashed line is for $\mathrm{EDit}_1(p,d)$ for *d*=2.

We will see in the next section that a learning curve for learning problems with two equiprobable classes should map one training example to an accuracy of 0.5. We can do this with this curve by shifting it to the left:

$$m = \mathrm{EDit}_2(p,d) = d(\mathrm{x}_{\mathrm{ED}}[1-p] - 32.936) = d\left( \frac{2\log(\frac{6}{1-p})}{(1-\sqrt{1-p})(1-p)} - 32.936 \right) \tag{4.6}$$

The effective-dimension model is defined in terms of the inverse of $\mathrm{EDit}_0$:

$$p = \mathrm{EDmodel}(m,d) = \mathrm{EDit}_0^{-1}(\frac{m+32.936}{d}) \tag{4.7}$$

Because $\mathrm{EDit}_0$ is of the form $z\log(z)$ it has no closed-form inverse. $\mathrm{EDit}_0$ is, however, differentiable, so its inverse can be computed numerically using Newton's method. For even greater speed, a table can store selected values of $\mathrm{EDit}_0^{-1}$, and other values can be determined via linear interpolation.

### 4.1.2.2. The $\mathrm{Burr}_1$ Model

The $\mathrm{Burr}_1$ model is similar in shape to, but simpler than, the EDmodel:

$$p = \mathrm{BurrModel}(z) = z/(z+1) \tag{4.8}$$

where $z = (m-1)/d+1$. Nelder [1966] describes a class of regression models he calls *inverse polynomial models* and develops procedures for using these models with continuous-response data. The reciprocal model is a specialization of this class of models. Aldrich and Nelson [1984] mention the function, which they call $\mathrm{Burr}_1$ as a candidate for regression, but in personal communications Nelson [1994] says that, as far as he knows, no one has used it for regression on binary-response data before. See Chapter 2 for a discussion of the use of similar functions with continuous-response data.

### 4.1.3. A Comparison of the Deterministic Models

In this section the four deterministic models are compared. The comparison shows that the two models based on computational learning theory (EDmodel, based on EDit, and BurrModel, based on $Burr_1$) have different shapes than the better-known logistic and probit (normal) binary-response models. We will see in the next chapter that, as one might suspected, the learning-inspired models fit and predict learning-performance data better.

Figure 4.4 plots BurrModel, EDmodel, and the logistic and probit (normal) regression models fitted so that all four intersect at $p$ = 0.70 and $p$ = 0.75.

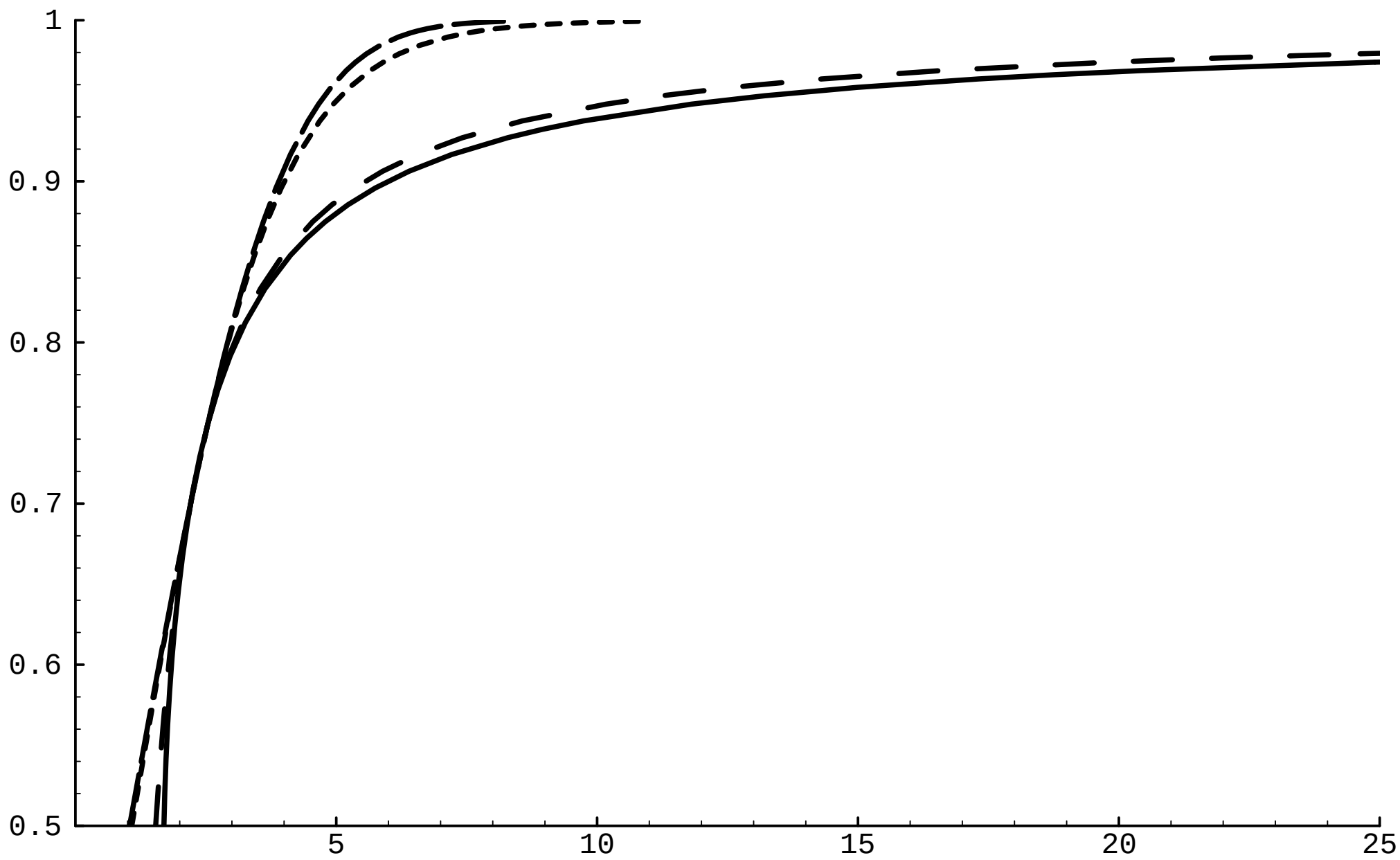

Figure 4.4: Four regression models fitted to intersect at accuracies 0.70 and 0.75. The top two curves are the logistic and probit (normal) regression models. The lower dashed line is BurrModel. The lower solid line is EDmodel.

Figure 4.5 shows the slope in the curves of Figure 4.4 as a function of accuracy. It shows that EDmodel and BurrModel are similar to each and quite different from the logistic model and the normal model. Figure 4.6 shows the ratio of EDit to $Burr_1$. It highlights the difference between EDit and $Burr_1$. As accuracy goes to 1, EDit becomes infinitely larger than $Burr_1$.

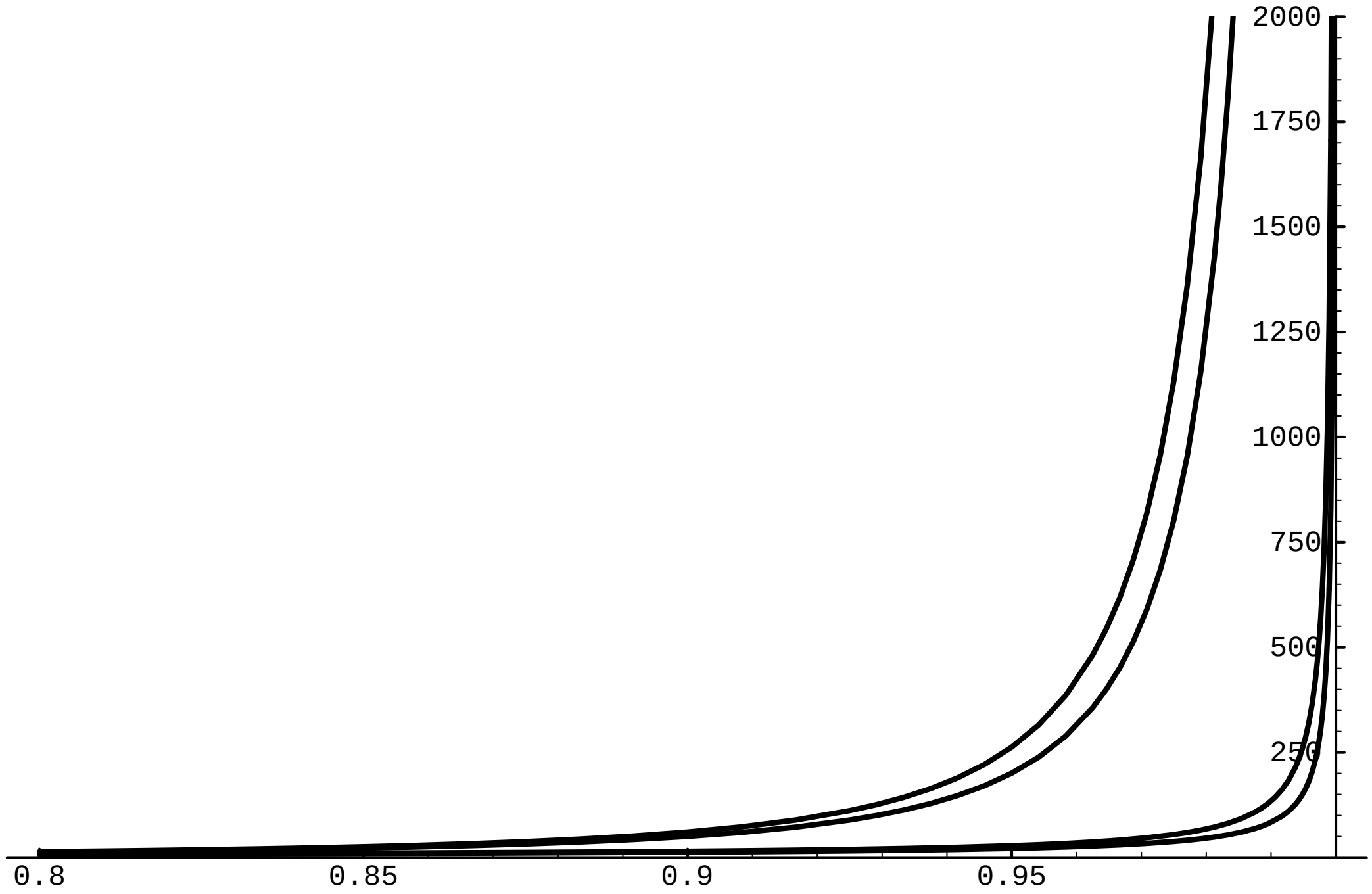

Figure 4.5: The slopes of the four curves as a function of accuracy. It shows that the logit and probit models are much different than the new models designed for learning curves.

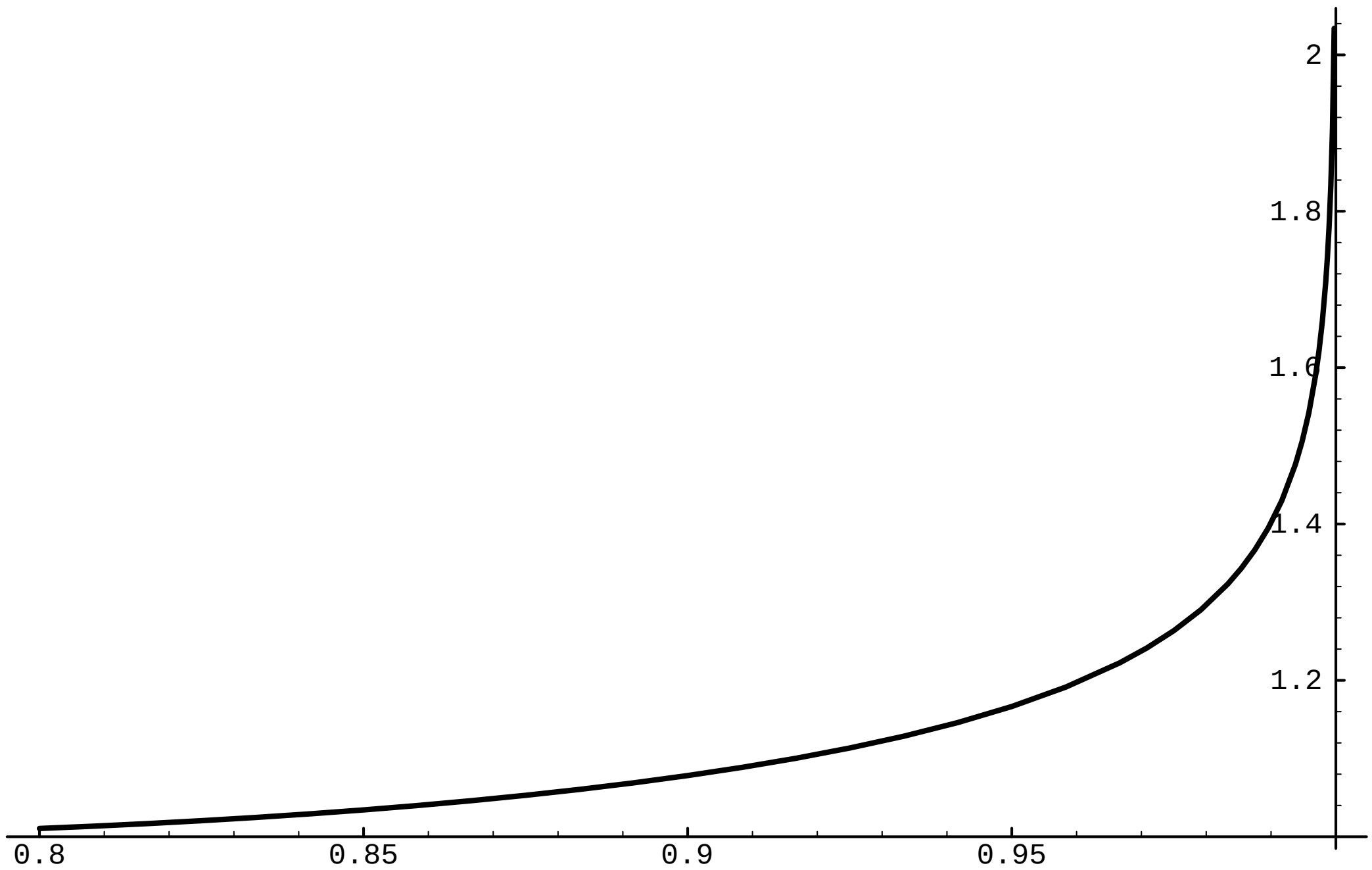

Figure 4.6: The ratio for EDit and $Burr_1$ as a function of accuracy. At high accuracies, the functions are not identical

## 4.2. Modeling the Effect of Multiple Classes, Skewed Classes, and Noise

This section develops a learning-performance model for learning in the presence of multiple classes, skewed classes, and noise. Like any good model, it makes predictions. Three of the model's most interesting predictions are:

1. The effects of class skew and class multiplicity can be predicted from the class frequencies. Actual learning-performance data need not be considered.

2. Once the effects of skew, multiple classes, and noise are factored out, learning performance can be well characterized by a single parameter.

3. When classes are skewed, learning error does not decrease linearly with the reciprocal of the number of examples; rather, it decreases in a predictable exponential way with that reciprocal.

The learning-performance model presented in this section is not meant to be a statement of computational learning theory. Rather, it is meant to be a heuristic that will be tested empirically.

Intuitively, we would expect **multiple classes** to make learning more difficult. For example, learning to distinguish 19 diseases *should* be harder than learning to distinguish two. Similarly, **noise** should also make learning harder. For example, if 10% of the training and testing examples given to a learner have the wrong class label, the ultimate achievable accuracy as measured on those noisy testing examples is reduced to at least 90%. On the other hand, **skewed** data can make learning easier. For example, suppose the learning problem is skewed so that 95% of the training and testing data is of one class. The learner could be 95% accurate just by always guessing the most frequent class. Even with just one training, the learner would have a 95% chance of seeing the most frequent class, and thus by hypothesizing that everything is of the class it sees, obtain an expected accuracy of at least 90.25% (=95%×95%).

The model ultimately developed in this section, $\text{model}_{gen}[z,start,skew,max]$, is this parameterized function from examples, *m*, to accuracy, *p* (as illustrated in Figure 4.7):

$$p = \text{model}_{\text{gen}}\left[(m-1)/d+1, start, skew, max\right] \\ = start + \left(max - start\right)\frac{\text{model}_5\left[(m-1)/d+1\right]^{skew(\text{Log}_{1/2}[start/2]-1)} - 0.5^{skew(\text{Log}_{1/2}[start/2]-1)}}{1-0.5^{skew(\text{Log}_{1/2}[start/2]-1)}} \tag{4.9}$$

where:

- model5[z] is a learning-performance model for the base case: 2 classes, 50%/50% class frequencies, no noise. One possible candidate is *z*/(*z*+1), the BurrModel of Section 4.1.2.2.
- *d* measures the learning rate.
- *start* is the expect accuracy when the number of training examples, *m*, is 1.
- *skew* is a measure of skew. A value of 1 is the skew when all classes are equiprobable.
- *max* is the highest possible expected accuracy. It is a measure of noise.

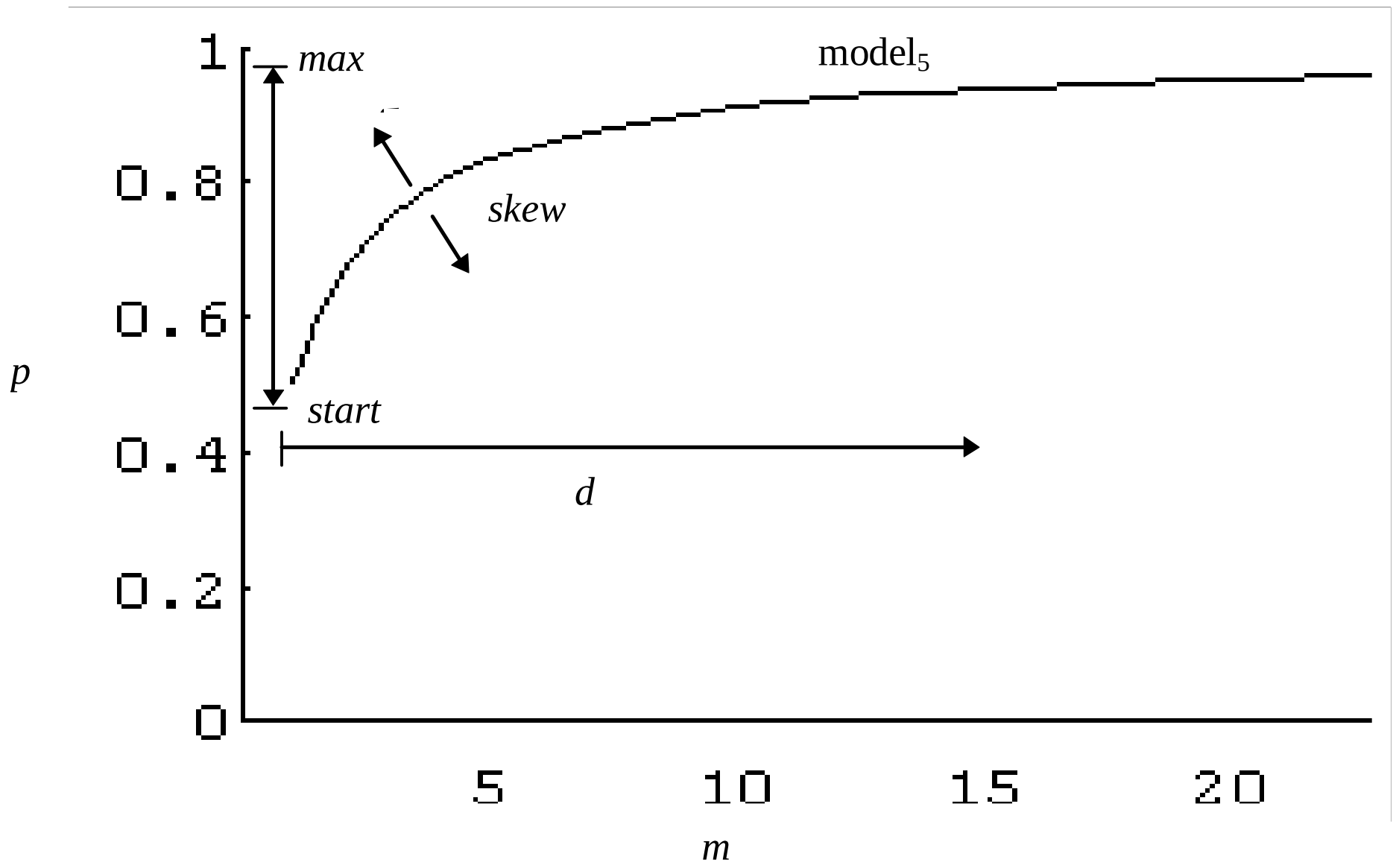

Figure 4.7: Function $\text{model}_{\text{gen}}[z,start,skew,max]$ maps *m*, the number of examples, into *p*, the expected accuracy. It is defined in terms of a function $\text{model}_5[m]$. Its *d* parameter expands or contracts the curve along the *x*-axis. Its *start* and *max* parameters define the low and high y-axis values. Its *skew* parameter makes the curve more concave or less concave.

The $\text{model}_{\text{gen}}[z,start,skew,max]$ learning-performance model depends on a $\text{model}_5[z]$ learning-performance model must be assumed. Section 4.1 discusses reasonable candidates for $\text{model}_5[z]$. Values for *start* and *skew* can be determined from the class frequencies observed in the training examples (Section 4.2.5). Values for *d* and *max* can be determined with maximum likelihood estimation. That is the topic of Section 4.2.

### 4.2.1. Learning from Only One Example

As a base case, consider the learning performance of an inductive learning when it is given only a single training example. Its best strategy if it has no other knowledge is to guess what it sees. That is, it should hypothesize that all examples are of the same class as the only training example it has seen.

- Observation 4.1: If there are two classes each with probability 0.5, the expected accuracy of the guess-what-you-see classification rule on independent testing data is 0.5.

- Observation 4.2: If there is only one class with class frequency 100% -- in other words, if all the testing and training examples have the same class label -- then the expected accuracy is 1.

- Observation 4.3: As the number of classes goes to infinity each with an infinitesimal probability of appearing, the expected accuracy goes to 0.

- Observation 4.4: If there are two classes, class "A" with probability 0.9 and class "B" with probability 0.1 -- the expected accuracy is:

| | | |
|---|---|---|
| The example is A: | probability 0.9, expected accuracy 0.9, | 0.81 |
| The example is B: | probability 0.1, expected accuracy 0.1, | 0.01 |
| | TOTAL EXPECTED ACCURACY = | 0.82 |

- Observation 4.5: Generalizing, given 1 example, if the probability of any class *c* is f[*c*], then the expected accuracy on that class is f[*c*]. Let *fvector* be the vector of class probabilities, then overall expected accuracy is:

$$p = \sum_{c \in classes} \mathrm{f}[c]^2 = fvector^{\mathrm{T}} \cdot fvector \tag{4.10}$$

So, if *fvector* is $<0.9, 0.1>$, $p = <0.9, 0.1>^{\mathrm{T}} \cdot <0.9, 0.1> = 0.9^2 + 0.1^2 = 0.82$

### 4.2.2. Learning from *m* Examples

Modeling the effect of *m* training examples requires some additional assumptions.

- Assumption 4.1: For learning tasks with two classes each with frequency 0.5, the expected accuracy *p* can be well estimated by some function $model_5[z]$, where *z* is $(m-1)/d+1$, *m* is the number of training examples and *d* is a parameter related to learning rate. From Observation 4.1, the function should have the property that:

$$model_5[1]=0.5 \tag{4.11}$$

This assumption and possible candidates for $model_5[z]$ were the subject of Section 4.1. One candidate for the function is BurrModel:

$$model_5[z] = z/(z+1) \tag{4.12}$$

It is plotted in Figure 4.8.

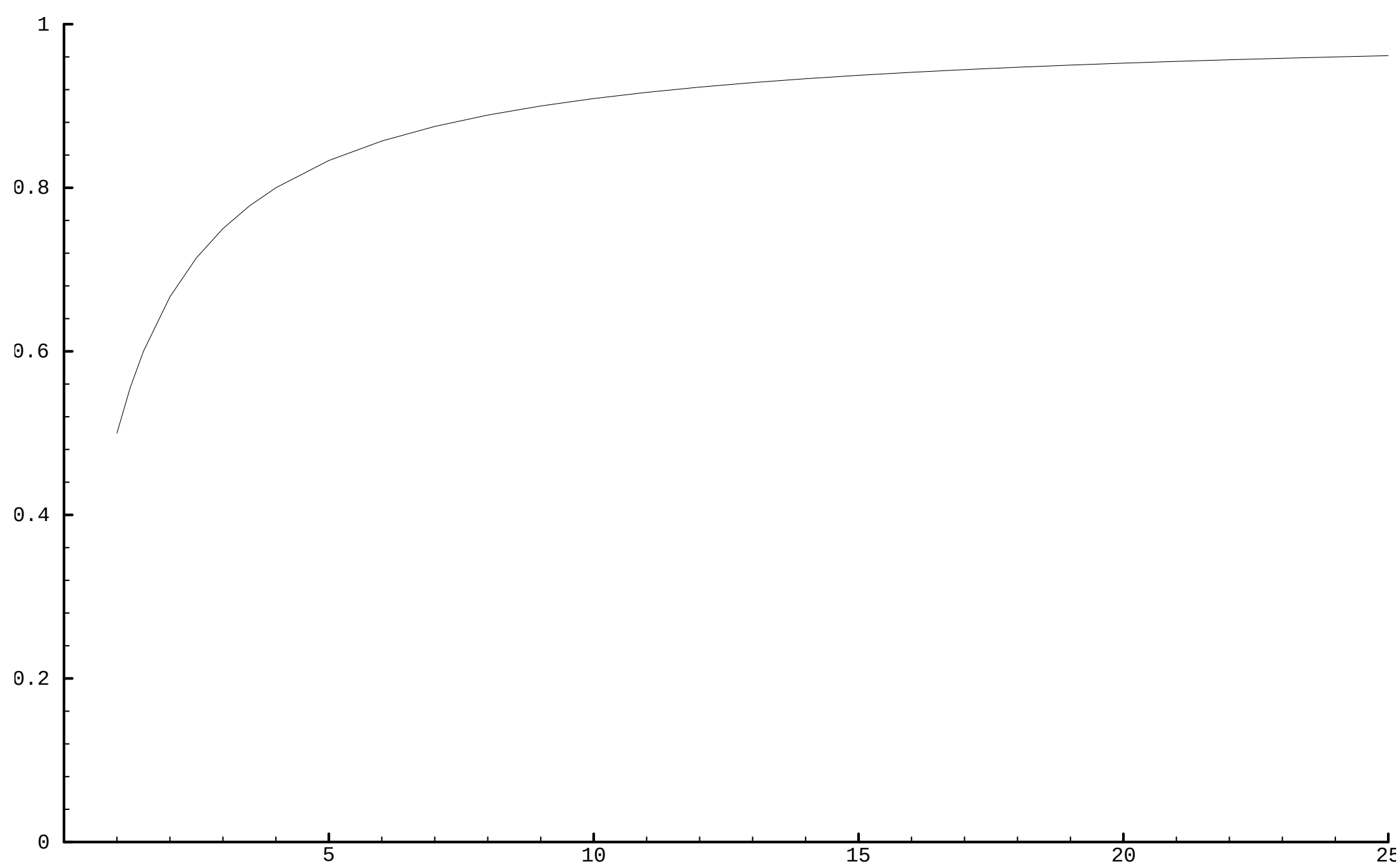

Figure 4.8. The *z/(z+1)* function, a candidate for model5[z]. The function maps *z* -- *z* is (*m*-1)/*d*+1, where *m* is the number of training examples and *d* is a constant -- to *p* , accuracy. model5[z] is assumed to model the learning on data with two classes, each with a 0.5 frequency.

- Observation 4.6: If there is only one class with probability 1, the expected accuracy for all $m$ is 1 (as before).

- Observation 4.7: As the number of classes all with an infinitesimal probability of appearing goes to infinity, the expected accuracy goes to 0 (for all finite $m$).

- Definition 4.1: Let $\text{model}_{\text{cla}}[z,f[c]]$, where $z$ is $(m-1)/d+1$, be the expected accuracy on any class $c$ with class probability f[$c$] of a learner given $m$ training examples. This definition assumes that expected accuracy on a class is only a function of class frequency and number of examples and not of the class itself.

- Assumption 4.2: If $f_1 < f_2$, $\text{model}_{\text{cla}}[z,f_1] < \text{model}_{\text{cla}}[z,f_2]$ (for all finite $z$). In other words, for a given number of training examples, the expected accuracy on a class will be greater the larger that class's frequency.

In general, given $m$ examples, the expected accuracy on all classes is:

$$p = \text{model}_{\text{all}}\left[z,\, fvector\right] = \sum_{f \in fvector} f \cdot \text{model}_{\text{cla}}\left[z,\, f\right] \tag{4.13}$$

the weighted average of the accuracy on each class. Note that when there are two classes and each has probability 0.5, by (4.13) and Definition 4.1:

$$\begin{aligned} p &= \text{model}_{\text{all}}\left[z,\langle 0.5,0.5\rangle\right] \\ &= 0.5\,\text{model}_{\text{cla}}\left[z,0.5\right] + 0.5\,\text{model}_{\text{cla}}\left[z,0.5\right] \\ &= 0.5\,\text{model}_5\left[z,\right] + 0.5\,\text{model}_5\left[z\right] = \text{model}_5\left[z\right] \end{aligned} \tag{4.14}$$

In other words, $\text{model}_5[z]$ is both the total expected accuracy on all classes when there are 2 classes of 50%/50% probability and the expected accuracy on a single class with 50% probability.

### 4.2.3. Heuristic for $\text{model}_{cla}[z,f]$ in Terms of $\text{model}_5[z]$

The next goal is to find a simple heuristic expression of $\text{model}_5[z]$ in terms of $\text{model}_5[z]$. The constraints are:

$$\text{model}_{cla}[z,0.5] = \text{model}_5[z] \qquad \text{Assumption 4.1}$$

$$\text{model}_{cla}[z,0.0] = 0.0 \text{ (for finite } z) \qquad \text{Observation 4.7}$$

$$\text{model}_{cla}[z,1.0] = 1.0 \qquad \text{Observation 4.6}$$

$$\text{model}_{cla}[1,f]= f \qquad \text{Observation 4.5}$$

$$\text{If } f_1 < f_2,\ \text{model}_{cla}[z,f_1] < \text{model}_{cla}[z,f_2] \text{ (for finite } z) \qquad \text{Assumption 4.2}$$

The intuition motivating the heuristic is that the $\text{model}_5[z]$ curve is our standard learning curve and that when a class probability is large, this standard curve will be pulled up toward higher accuracy and that when a class probability is low the class probability will be pulled low toward zero.

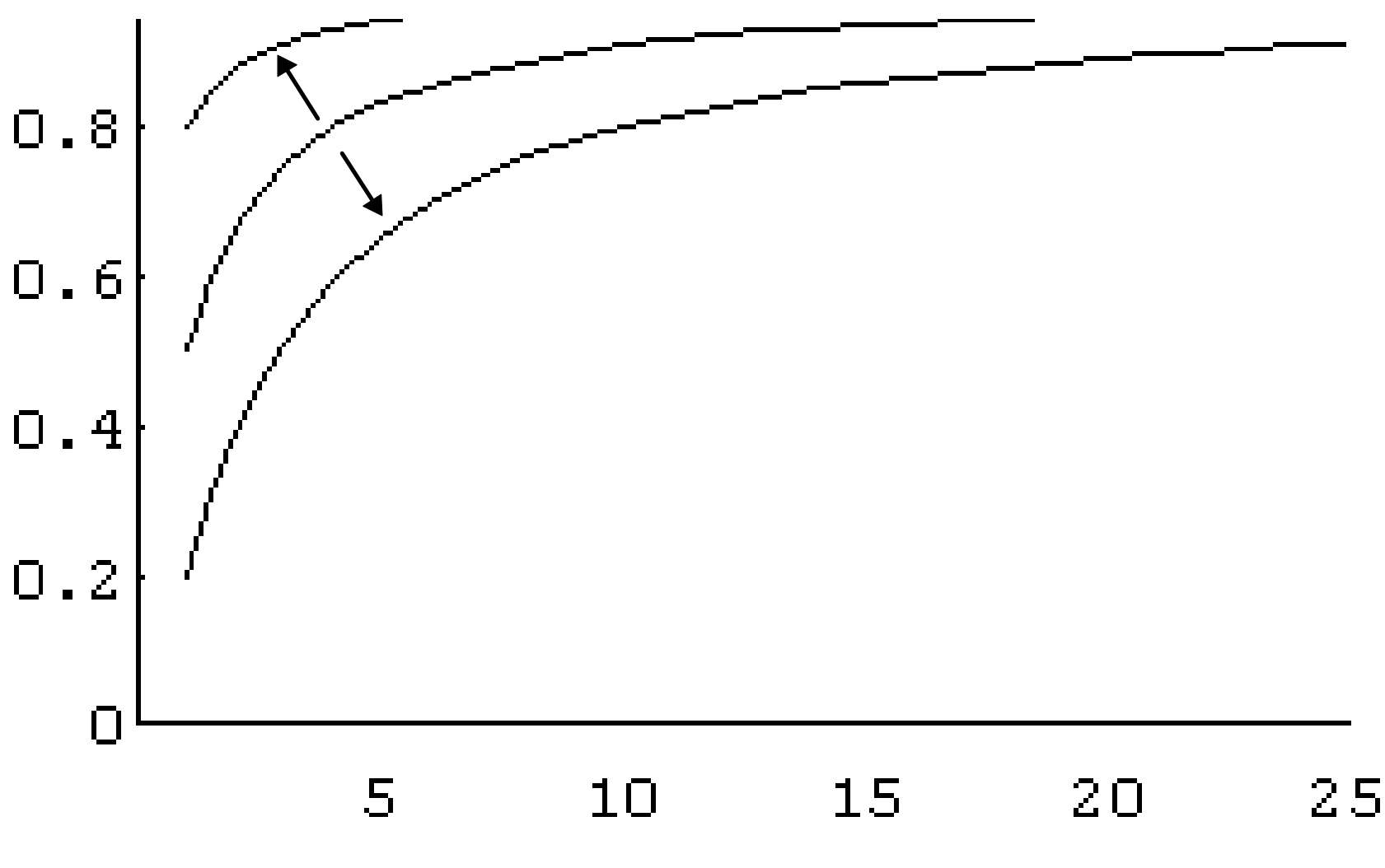

Figure 4.9. How a standard learning-performance model which has value 0.5 when $z$=1, might be pulled up to produced a curve with greater value when $z$=1 or pulled down to produced a lesser value when $z$=1.

For any $z$ and $f$,

$$\text{if } 0.0 < f < 0.5,\ 0.0 < \text{model}_{\text{cla}}[z,f] < \text{model}_5[z] \qquad (4.15)$$

and

$$\text{if } 0.5 < f < 1.0,\ \text{model}_5[z] < \text{model}_{\text{cla}}[z,f] < 1.0 \qquad (4.16)$$

Here is a simple function with the right properties:

$$p = \text{model}_{\text{cla}}\left[z, f\right] = \text{model}_5\left[z\right]^{\text{Log}_{1/2}[f/2]-1} \qquad (4.17)$$

For example, if $\text{model}_5[z]$ is $z/(z+1)$, here are the curves for $f$= 0.0, 0.1, …, 1.0.

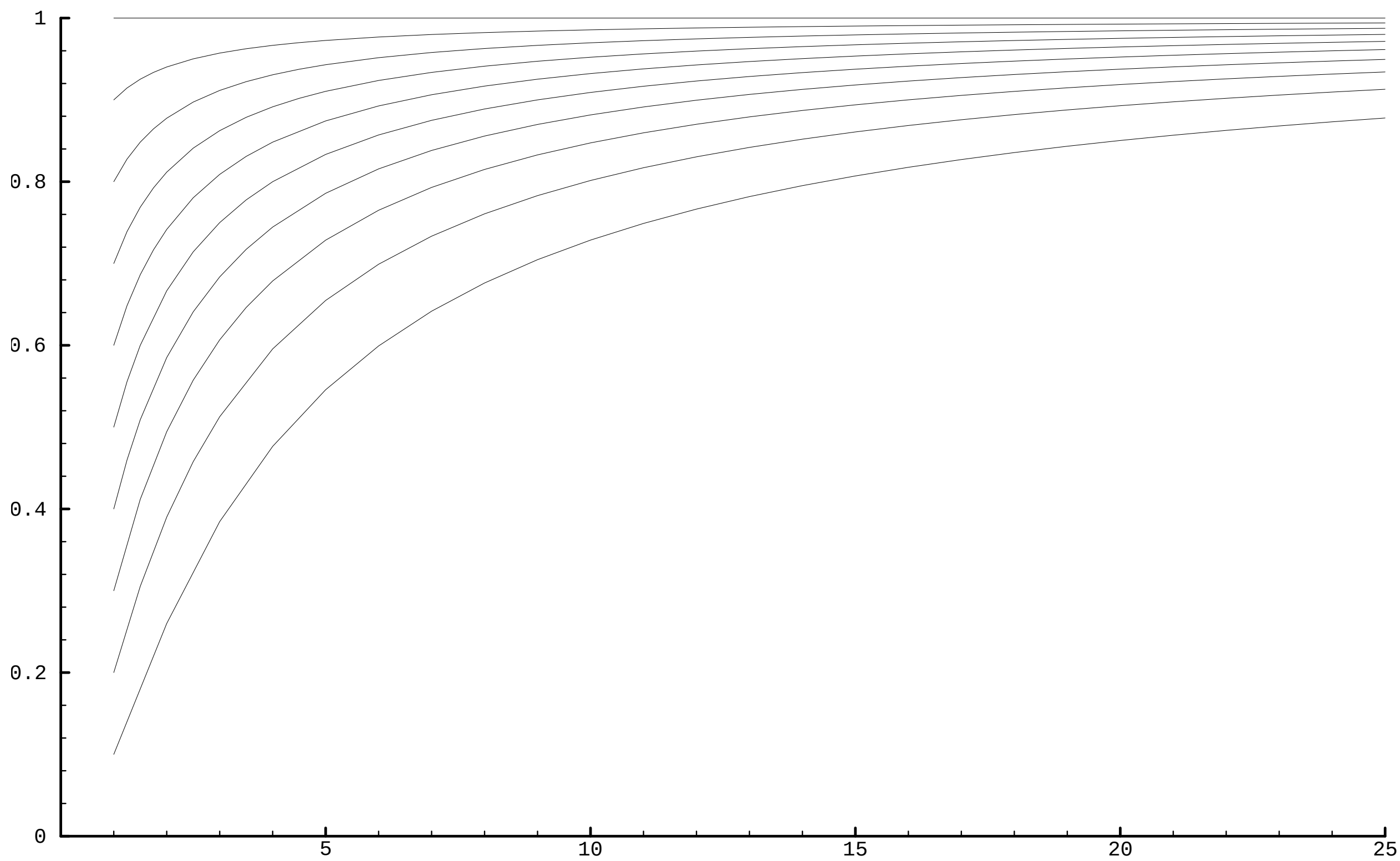

Figure 4.10: Learning-performance models for classes with class frequencies $f$= 0.0, 0.1, …, 1.0. Each model is a learning curve function, $\text{model}_{\text{cla}}[z,f]$, that maps $z$ (a linear function of the number of training examples) to $p$, accuracy in identifying that class. Note that when the number of training examples is 1 (and $z$=1), the accuracy is the class frequency. This plot assumes that the “standard” learning-performance model, $\text{model}_5[z]$, is $z/(z+1)$ .

Applying the heuristic to $\text{model}_{\text{all}}[z,\mathit{fvector}]$ gives

$$p = \text{model}_{\text{all}}\left[z, \mathit{fvector}\right] = \sum_{f \in \mathit{fvector}} f \cdot \text{model}_5[z]^{\text{Log}_{1/2}[f/2]-1} \tag{4.18}$$

### 4.2.4. Learning with Multiple, Equiprobable Classes

If we have *classnum* classes each of equiprobability, the heuristic predicts this overall accuracy:

$$\begin{aligned} p &= \text{model}_{\text{all}}\left[z, \left\langle \frac{1}{\mathit{classum}}, \cdots, \frac{1}{\mathit{classum}} \right\rangle \right] \\ &= \frac{1}{\mathit{classnum}} \text{model}_5[z]^{\text{Log}_{1/2}[1/(2\mathit{classnum})]-1} + \cdots + \frac{1}{\mathit{classnum}} \text{model}_5[z]^{\text{Log}_{1/2}[1/(2\mathit{classnum})]-1} \\ &= \text{model}_5[z]^{\text{Log}_{1/2}[1/(2\mathit{classnum})]-1} \\ &= \text{model}_{\text{cla}}[z, 1/\mathit{classnum}] \end{aligned} \tag{4.19}$$

### 4.2.5. Learning with Skewed Classes

If we have *classnum* classes with a variety of class probabilities, the heuristic predicts an overall accuracy of:

$$\begin{aligned} p &= \text{model}_{\text{all}}\left[z, \left\langle f_1, f_2, \cdots, f_{\mathit{classnum}} \right\rangle \right] \\ &= f_1 \,\text{model}_5[z]^{\text{Log}_{1/2}[f_1/2]-1} + f_2 \,\text{model}_5[z]^{\text{Log}_{1/2}[f_2/2]-1} \cdots + f_{\mathit{classnum}} \,\text{model}_5[z]^{\text{Log}_{1/2}[f_{\mathit{classnum}}/2]-1} \end{aligned} \tag{4.20}$$

This is the weighted sum of $\text{model}_{\text{cla}}[z,f]$ functions. The sum can't be expressed exactly as a $\text{model}_{\text{cla}}[z,f]$ curve, but it can be closely approximated with a generalization of the $\text{model}_{\text{cla}}[z,f]$ function. Such an approximation makes sense since the $\text{model}_{\text{cla}}[z,f]$ function is itself only a heuristic.

To see the intuition behind the approximation, start with the $\text{model}_{\text{cla}}[z,f]$ function:

$$p = \text{model}_{\text{cla}}[z, f] = \text{model}_5[z]^{\text{Log}_{1/2}[f/2]-1} \tag{4.21}$$

Recall that when $z$=1, , the $\text{model}_{\text{cla}}[z,f]= f$. Thus, the exponent, $\text{Log}_{½}[f/2]-1$, and value when $z$=1 coupled. The approximation is a generalization that uncouples these two values:

$$p = \text{model}_{\text{skew}}\left[z, start, skew\right] = start + (1 - start)\frac{\text{model}_5\left[(m-1)/d + 1\right]^{\text{Log}_{1/2}[start/2]-1} - 0.5^{\text{Log}_{1/2}[start/2]-1}}{1 - 0.5^{\text{Log}_{1/2}[start/2]-1}} \quad (4.22)$$

where *start* is the expected accuracy when $z$=1 and *skew* is the exponent.

The final step in the approximation is setting the parameters *start* and *skew* so that $\text{model}_{\text{skew}}[z,start,skew] \approx \text{model}_{\text{all}}[z,fvector]$. The strategy will be to set *start* and *skew* so that the two functions intersect for three values of $z$.

First, we would like them to intersect as $z$ goes to infinity. This is automatic because both functions do go to 1 as $z$ goes to infinity.

Second, we would like them in intersect when $z$=1. The value of $\text{model}_{\text{all}}[1,fvector]$ is $fvector^{\text{T}} \cdot fvector$. The value of $\text{model}_{\text{skew}}[1,start,skew]$ is *start*. So we can make the two functions intersect be setting *start* to $fvector^{\text{T}} \cdot fvector$.

Third, we would like them to intersect at a third point. A convenient point is that $z$ such that $\text{model}_5[z]$=0.75. At that value of $z$, $\text{model}_{\text{skew}}[z,start,skew]$ will equal $\text{model}_{\text{all}}[z,<f_1,f_2,\ldots,f_{classnum}>]$ if and only if:

$$start + (1 - start)\frac{0.75^{skew(\text{Log}_{1/2}[start/2]-1)} - 0.5^{skew(\text{Log}_{1/2}[start/2]-1)}}{1 - 0.5^{skew(\text{Log}_{1/2}[start/2]-1)}} = f_1 0.75^{\text{Log}_{1/2}[f_1/2]-1} + f_2 0.75^{\text{Log}_{1/2}[f_2/2]-1} \cdots + f_{classnum} 0.75^{\text{Log}_{1/2}[f_{classnum}/2]-1} \quad (4.23)$$

The only unknown is *skew*; its value can be determined numerically with, for example, the secant method or with the Newton-Raphson Method [Press *et al*., 1992].

Here is how $\text{model}_{\text{skew}}[z,start,skew]$ fits in with our previous results for learning in which all the classes are equiprobable. When number of classes is two and each has 50%/50% probability, our model is $\text{model}_5[z]$. This exactly equals $\text{model}_{\text{skew}}[z,0.5,1]$. More generally, when number of classes is *classnum* and each has the same probability, our model was $\text{model}_5[z]^{\text{Log}_{½}[f/2]-1}$. This exactly equals $\text{model}_{skew}[z,1/classnum,1]$

The plot shows $\text{model}_{\text{skew}}[z,0.5,skew]$ for *skew*=0.001, 1, 2, 4, and 8. The topmost curve is *skew*=0.001.

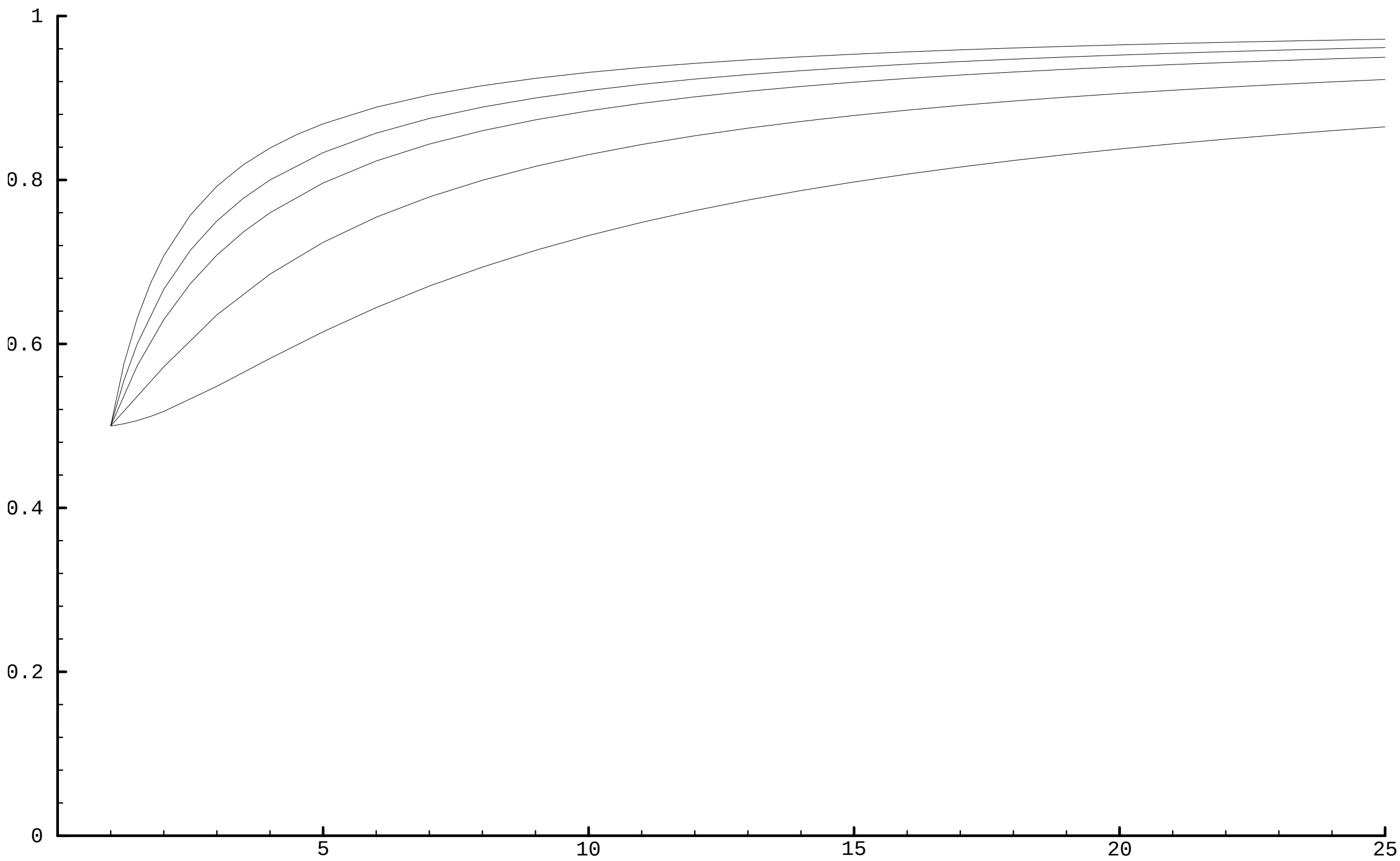

Figure 4.11: Learning-performance models with a *start* value of .5 and *skew* values of 0.001, 2, 4, and 8. Each model is a learning curve function, $\text{model}_{\text{skew}}[z,start,skew]$, that maps *z* (a linear function of the number of training examples) to *p*, accuracy in identifying that class. This plot assumes that the “standard” learning-performance model, $\text{model}_5[z]$, is $z/(z+1)$.

Here is an example from soybean-disease diagnoses in which there are many classes and the class frequencies vary widely . Suppose the class frequencies are:

$$\langle 0.0130293, 0.00325733, 0.019544, 0.019544, 0.130293, 0.130293, 0.0325733, 0.0651466, 0.0325733, 0.0325733, 0.0325733, 0.130293, 0.0325733, 0.0325733, 0.0651466, 0.130293, 0.0325733, 0.0325733, 0.0325733 \rangle \qquad (4.24)$$

The $model_{all}[z,fvector]$ heuristic for these class frequencies is:

$$0.521172\ model_5[z]^{2.94017} + 0.130293\ model_5[z]^{3.94017} + 0.29316\ model_5[z]^{4.94017} + 0.039088\ model_5[z]^{5.67713} + 0.0130293\ model_5[z]^{6.2621} + 0.00325733\ model_5[z]^{8.26209} \quad (4.25)$$

If $model_5[z]$ is assumed to be $z/(z+1)$, $model_{all}[z,fvector]$ looks like

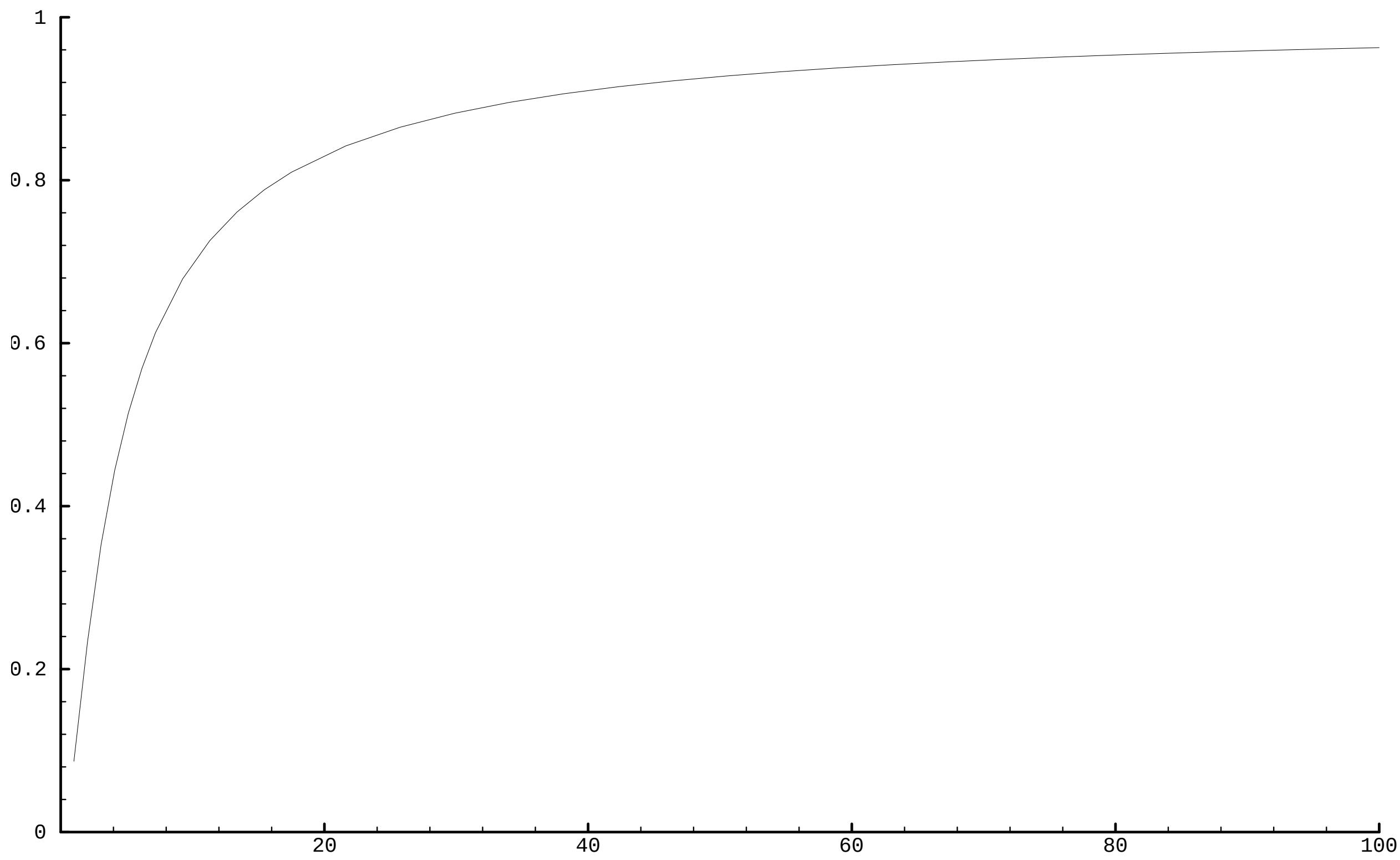

Figure 4.12: A learning-performance model for the soybean disease data. The model is a function, $model_{all}[z,fvector]$, that maps $z$ (a linear function of the number of training examples) to $p$ , accuracy in identifying that class. The quantity *fvector* is the vector of class frequencies for the soybean data. This plot assumes that the "standard" learning-performance model, $model_5[z]$, is $z/(z+1)$.

To determine the $model_{skew}[z,start,skew]$ approximation to $model_{all}[z,fvector]$, we need to determine *start* and *skew*. The value of *start* is just $fvector^T \cdot fvector = 0.0868867$.

To determine *skew*, we want to find a value for it such that

$$model_{skew}[z,start,skew]=model_{all}[z,fvector] \text{ when } model_5[z]=0.75. \quad (4.26)$$

Substituting in values, we want to find a value for *skew* that will solve

$$0.0868867+\frac{0.913113\left(0.75^{3.52472\,skew}-0.5^{3.52472\,skew}\right)}{1-0.5^{3.52472\,skew}}=0.346491 \tag{4.27}$$

The value is found numerically to be 1.07556. So, $\text{model}_{\text{skew}}[z,start,skew]$ is

$$0.0157871 + 0.984213\ \text{model}_5[z]^{3.79106} \tag{4.28}$$

which is much simpler than $\text{model}_{\text{all}}[z,fvector]$.

Again assuming that $\text{model}_5[z]$ is $z/(z+1)$, here is a plot of both the $\text{model}_{\text{all}}[z,fvector]$ heuristic and its $\text{model}_{\text{skew}}[z,start,skew]$ approximation:

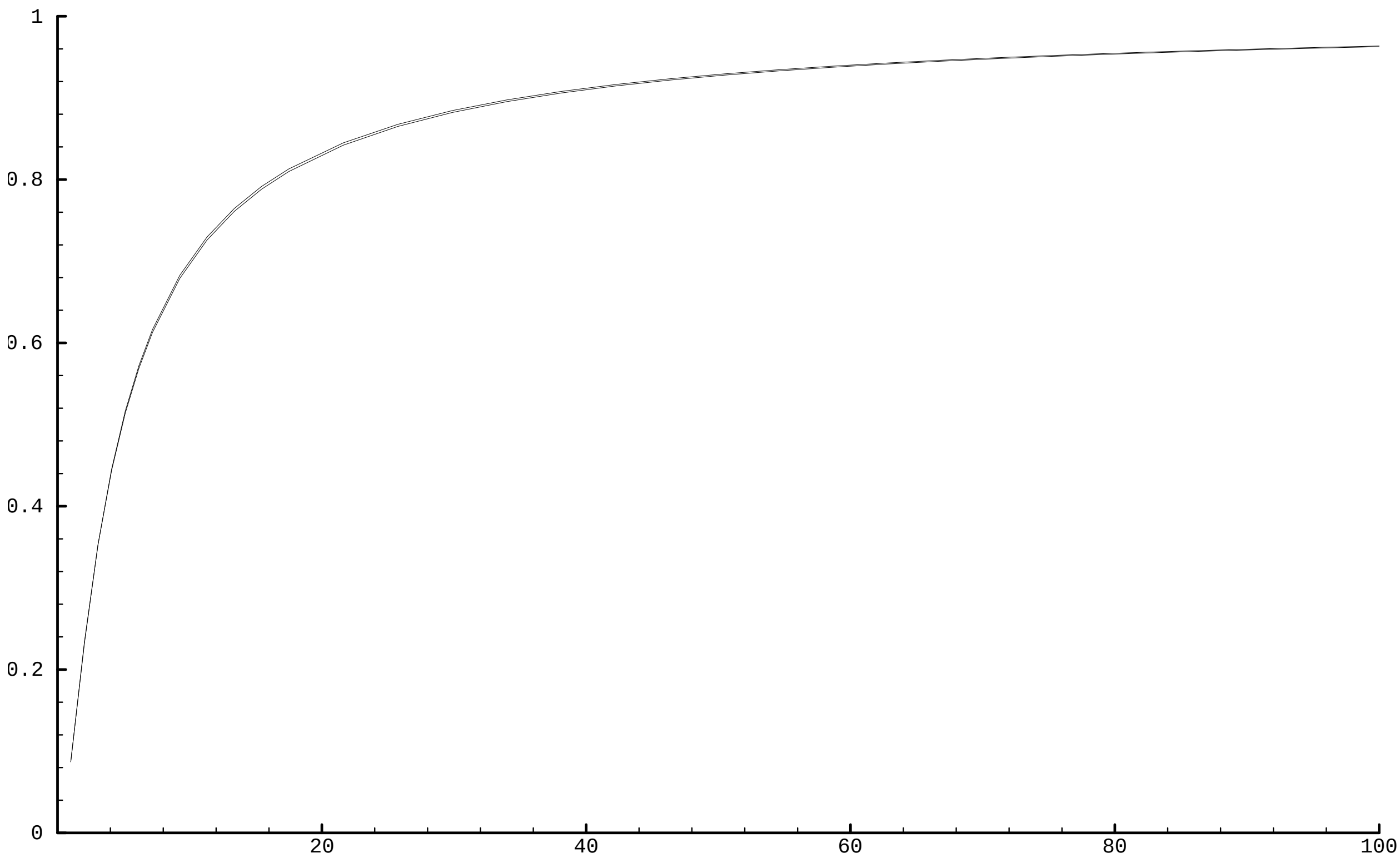

Figure 4.13: Two almost indistinguishable learning-performance models for the soybean disease data. One is the same $\text{model}_{\text{all}}[z,fvector]$ function plotted in Figure 4.12. The other is $\text{model}_{\text{skew}}[z,start,skew]$, a simpler function with much the same shape. This plot assumes that the "standard" learning-performance model, $\text{model}_5[z]$, is $z/(z+1)$.

### 4.2.6. Learning with Noise

There are many possible ways to define noise. For the purposes of this thesis, it is defined as the difference between 100% accuracy and *max*, the expected accuracy as *z* (and *m*) go to infinity.

This suggests that noise can be added to a learning curve model through scaling a learning curve model by the quantity *max*. This doesn't quite work because it would scale down *start*. The *start* value is determined from the observed class frequencies. If there is noise, the noise will already be reflected in these class frequencies and in *start*. Thus, a model of noise should not affect *start*.

We will do the next simplest thing, scale down the learning curve model, but do so with *start* as the origin.

$$\begin{aligned} p &= \text{model}_{\text{gen}}[z, start, skew, max] \\ &= start + (max - start)\frac{\text{model}_{\text{skew}}[z, start, skew] - start}{1 - start} \\ &= start + (max - start)\frac{\text{model}_5[(m-1)/d+1]^{skew(\text{Log}_{1/2}[start/2]-1)} - 0.5^{skew(\text{Log}_{1/2}[start/2]-1)}}{1 - 0.5^{skew(\text{Log}_{1/2}[start/2]-1)}} \end{aligned} \tag{4.29}$$

The result is an expression very similar to the expression for $\text{model}_{\text{skew}}[z,start,skew]$. The plot shows $\text{model}_{\text{gen}}[z,0.5,1,0.8]$ with $\text{model}_5[z]$ defined as $z/(z+1)$. Notice that $\text{model}_{\text{gen}}[z,start,skew,1] = \text{model}_{\text{skew}}[z,start,skew]$.

### 4.2.7. Working with $\text{model}_{\text{gen}}[z,start,skew,max]$

In outline, the steps to fitting a $\text{model}_{\text{gen}}[z,start,skew,max]$ curve to data are:

1. From the training data, measure *fvector*, the class frequencies.

2. From *fvector*, compute *start* and *skew*.

3. Into $model_{gen}[z, start, skew, max]$ substitute the values of *start* and *skew*, also substitute in $(m-1)/d+1$ for $z$ and whatever function is being used for $model_5[z]$. For example, using the $z/(z+1)$ for $model_5[z]$ and the soybean frequencies we saw earlier produces:

$$p = 0.0868867 + 1.07787(\max - 0.0868867)\left(\frac{d+m-1}{2d+m-1}\right)^{3.79106} \quad (4.30)$$

4. The learning-performance data gives observed values for $m$, the number of training examples, and $p$, the observed accuracy. Find the values of $d$ and *max* that produce the curve that best fits the data. The details on how to do this are the subject of Section 4.4.

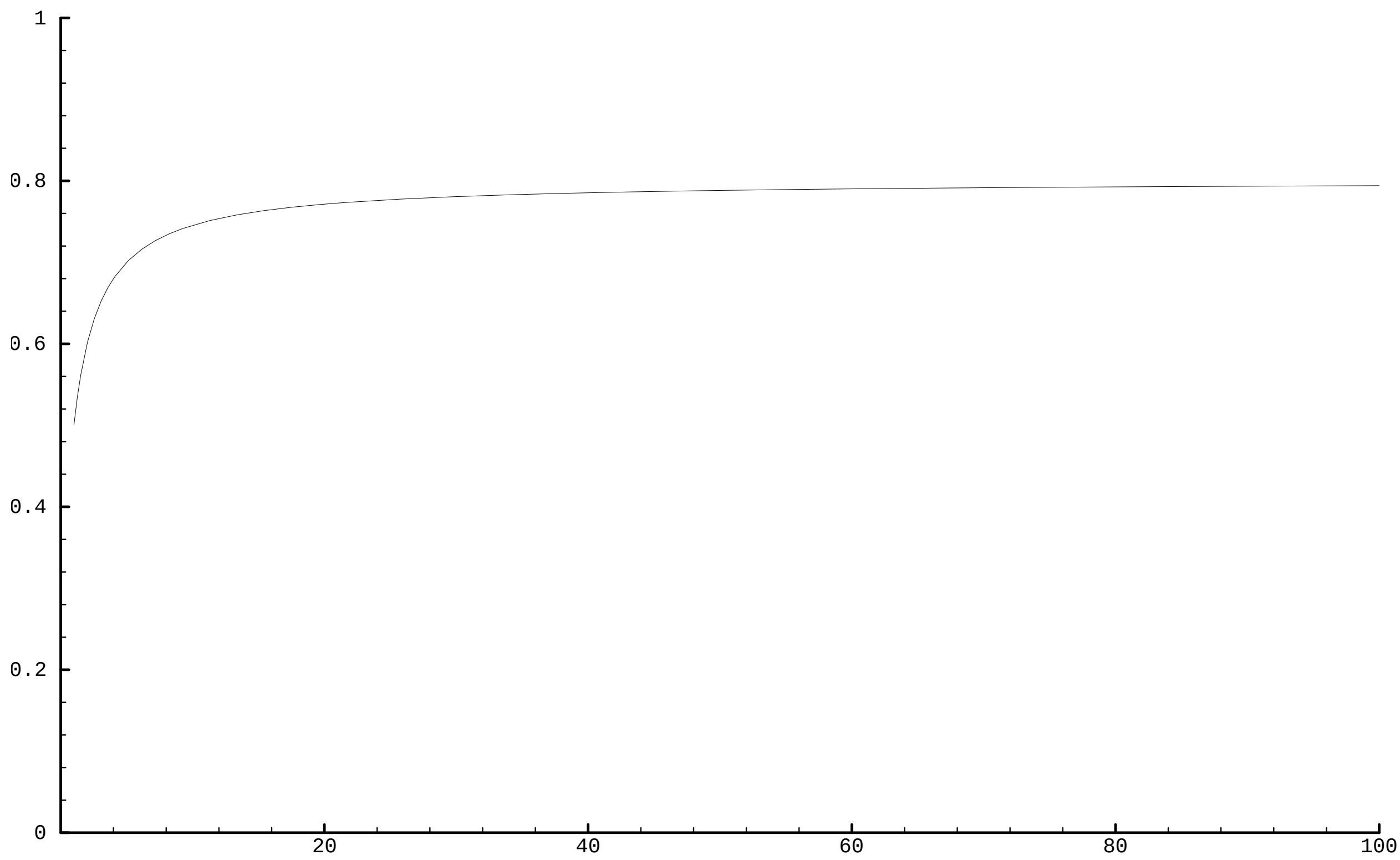

Figure 4.14: A learning-performance model that models the effect of 20% noise. The learning-performance model is $model_{gen}[z, 0.5, 1, 0.8]$. The 0.8 parameter is 100%-20%, the maximum attainable accuracy. This plot assumes that the "standard" learning-performance model, $model_5[z]$, is $z/(z+1)$.

### 4.2.8. Simplified Model

But is the complexity of Equation (4.29) really necessary? To find out, we create $model_{noexp,}$, a simpler model based on Equation (4.29) but without the exponents.

$$\begin{aligned} p &= \text{model}_{\text{noexp}}\left[(m-1)/d+1, start, max\right] \\ &= start + \left(max - start\right)\frac{\text{model}_5\left[\left(m-1\right)/d+1\right]-0.5}{1-0.5} \end{aligned} \qquad (4.31)$$

In these case of two equiprobable classes, Equation (4.29) reduces to Equation (4.31), but in the next chapter we will true it on other situations to see how important the exponents are.

## 4.3. Nondeterministic Design Choices

Recall from Section 3.1.1 that regression models are often constructed in two parts, a deterministic part and a nondeterministic part. In ordinary linear regression, for example, the deterministic part of the model is a line and the nondeterministic part is a normal distribution around the points in the line. Section 3.1.4 showed why the popular normal distribution was inappropriate for learning-performance data. This section develops two appropriate candidates for the nondeterministic part of learning-performance models.

The learning-performance data with which Seer works includes $y$, the number of testing examples correctly classified. This is necessarily an integer between 0 and $k$, the number of testing examples. If the testing examples are selected independently, the number of correct classifications should be binomially distributed because it is the result of independent identically distributed Bernoulli trials. For example, if the expected accuracy of the classification rule being tested is 0.705 and it is tested on 100 examples, the binomial distribution will specify the probability that the classification rule will get 70 examples correct, the probability it will get 71 correct, 69 correct, and so on for the range from 0 to 100.

The binomial distribution is:

$$p^{y}(1-p)^{(k-y)}\binom{k}{y} \tag{4.32}$$

where $y$ is the number of correctly classified testing examples, $k$ is the total number of testing examples, $p$ is the probability that a given testing example will be classified correctly, and $\binom{k}{y}$ is the number of distinct subsets of size $y$ of a set of size $k$.

The binomial distribution accounts for the randomness in testing examples, but not for other sources of randomness. For example, a learning program might produce a classification rule based on 49 training examples that is more accurate (as measured on an extremely large number of testing examples) than one based on 50 training examples. This could happen with a decision tree learner, for example, if the learner tended to make a split with 50 training examples that it didn't make when given only 49 examples. In other words, even with an infinite number of training examples, real-life learning curves may have an inherent roughness caused by the randomness in the training examples.

A reasonable way to model this roughness is with the beta distribution. Intuitively, it can be thought of as a normal distribution that is bounded by 0 and 1. The beta distribution is usually parameterized with two values, $\alpha$ and $\beta$. For our purposes, however, it is more convenient to parameterize it with $p$, its mean, and *pvar*, a parameter related to variance. With this parameterization the beta distribution has this density:

$$\frac{(1-p_1)^{(1-p-pvar)/pvar}\, p_1^{\,p/pvar-1}}{\text{Beta}[\frac{p}{pvar},\frac{1-p}{pvar}]}, \text{ where } p_1 \text{ ranges from 0 to 1.} \tag{4.33}$$

To see how this could be put together with the binomial distribution, consider a random learning-performance-like data generator. The input to the deterministic model would be $m$, the number of training examples. The output will be $p$, the expected accuracy before considering the randomness in the training and testing examples. This $p$ and a *pvar* value are the input into a beta random generator. The output is $p_1$ the expected accuracy after taking into account the randomness in the training examples. This $p_1$ along with $k$, the number of testing examples, is the input into a random binomial random generator. The output $y$ is the number of correctly classified training examples according to the random generator.

Combining a beta distribution with a binomial distribution, as done here, creates a *beta-binomial distribution*:

$$\frac{\text{Beta}[y + \frac{p}{pvar}, (k - y) + \frac{p}{pvar}]\binom{k}{y}}{\text{Beta}[\frac{p}{pvar}, \frac{1 - p}{pvar}]} \tag{4.34}$$

where $y$ is the number of testing examples, $k$ is the number of testing examples classified correctly, $p$ is the output of the deterministic part of the model, and *pvar* is related to the amount of roughness caused by randomness in the training examples. When *pvar* is zero, this distribution is becomes the binomial distribution.

The distribution is also known as the Polya, Polya-Eggenberger, negative hypergeometric, and generalized hypergeometric type IIA [Johnson *et al*., 1992; Griffiths, 1973]. It has been used in the past to model the scores of students in a class on a test, but not, so far as the author knows, as the nondeterministic part of a regression model [Wilcox, 1981].

When used as part of a regression model, $y$ and $k$ will be given as part of the learning-performance data. The value $p$ will be provided by the deterministic part of the model. The *pvar* parameter, however, will need to be estimated. The estimation method is analogous to estimating the variance of the data for linear regression except that instead of minimizing least square error, the method must find the value of *pvar* that maximizes likelihood. The next section has computational details on this.

## 4.4. Fitting Models to Data Efficiently

Chapter 3 defined the criteria with which to select the best learning-performance model for a set of data (namely maximizing likelihood). The previous sections of this chapter detailed the sets of learning-performance models to be considered. This section specifies the numerical techniques Seer uses to efficiently find the model with the maximum likelihood. Efficiency is important because the algorithms used are iterative. The difference in speed between a good implementation and a poor one could easily be 1000 to 1, and even more if the poor implementation doesn't converge.

For any given problem, Seer is given 1) learning performance data, 2) a nondeterministic model (either the binomial distribution or the beta-binomial distribution) and a deterministic model (for example, $model_{gen}$ with the BurrModel as $model_5$). Seer's task is to find the parameters of the deterministic model (e.g. *d* and *max*) and the parameters of the nondeterministic model (e.g. none or *pvar*) that produce a model with the maximum likelihood with respect to the data. Such parameters are called the maximum likelihood estimators.

As posed, the problem is an instance of nonlinear programming. The nonlinear-programming constraints relate to the parameters. For example, *pvar* and *d* must be greater than 0 and *max* must range between *start* and 1.0. The problem can be simplified by eliminating these constraints. This can be done by making the original parameters functions of parameters that can take any real value [Dixon, 1972, p. 89]. For example, *d* can be replaced with $daug^2$ and *pvar* can be replaced with $pvaraug^2$. Then maximum likelihood estimators can be found for *daug* and *pvaraug*. The square root of these estimators will be the estimators for *d* and *pvar*. The *max* parameter can be replaced with

$$\frac{1+start}{2}+\frac{1-start}{2}\sin(maxarg) \tag{4.35}$$

which will guarantee that *max* will be in the range from *start* to 1.

Removing the constraints turns the problem from one of nonlinear programming into one of nonlinear optimization. Seer uses the very fast Levenberg-Marquardt method of nonlinear optimization [Marquardt, 1963; Press *et al*., 1992]. This method smoothly varies between inverse-Hessian methods (like Newton-Raphson) and steepest descent methods. Its use normally requires both the first partial derivatives (the gradient) and second partial derivatives (the Hessian) of the function to optimize. Seer does require the first derivatives, but it avoids the need for the Hessian estimating the Hessian Q as

$$-\sum_i q_i(\theta)q_i(\theta)^T \tag{4.36}$$

where $q_i$ is the vector of first partial derivatives evaluated on the *i*th learning performance tuple [Cramer, 1986, p. 27].

When the learning-performance data is independent, covariance can be approximated. The asymptotic covariance matrix of the estimated parameters can be approximated with $-Q^{-1.}$ [Cramer, 1986]. When the generalized cross-validation method of Section 3.2 is used, the data is not independent, so covariance cannot be estimated.

## 4.5. Summary

This chapter detailed the design and selection method of the learning-performance models that Seer uses. We considered the design problem in three parts: First, create deterministic models for noise-free learning of two equally probable classes. We saw that the two models inspired by computational learning theory, EDmodel and BurrModel, have a much different shape than the two most popular binary-response regression models. The second part of the model design was modeling the effect of multiple classes, skewed classes, and noise. By considering progressively more complex cases, we traced the development of the new $model_{gen}$ heuristic. It models the effects of interest with remarkably few parameters. The third part of model design involved creating a nondeterministic model. We saw that the Binomial distribution should be sufficient if variation in testing data was our only concern. However, if we are also concerned about variation in the training data, then the Beta-Binomial distribution might be better. Finally, after all the design choices were enumerated, Section 4.4 specified how to efficiently find the best model in a possibly infinite set of models. The basic method is nonlinear optimization with an iterative algorithm.

Given these design choices and this selection method, which sets of models work best on real learning-performance data? That is the topic of the next chapter.

# 5. Experimental Procedure and Results

The previous chapter dealt with designing various reasonable components of learning-performance models. In this chapter we assemble these components into five learning-performance model sets. Using Seer, each set is tested on three real-world learning domains . Each model set is tested on how well it can characterize learning-performance data (3 trials) and on how well it can predict learning performance data (30 trials). Figure 5.1 summarizes the experimental variables (domain, task, and learning-performance model set) investigated. The experiments show that many of the model sets characterize well and predict as well as the data permits. They suggest that the choice of a model set is a tradeoff between computational ease and the ability to measure interesting aspects of the learning problem.

| Domain | Description | # Examples | # Attr | # Classes |
|---|---|---|---|---|
| Soybean | Soybean Disease Diagnosis | 307 | 35 | 19 |
| Heart | Heart Disease Diagnosis | 303 | 13 | 2 |
| Audio | Audiological Problems | 226 | 69 | 24 |

×

| Task | # Trials |
|---|---|
| Characterize | 3 |
| Predict | 30 |

×

| Model Set | Det. #2 | Det. #1 | Nondet. |
|---|---|---|---|
| Modelgen | $Burr_1$ | $model_{gen}$ | binomial |
| Logistic | Logistic model | - | binomial |
| ED | EDit | $model_{gen}$ | binomial |
| NoExp | $Burr_1$ | $model_{noexp}$ | binomial |
| Pvar | $Burr_1$ | $model_{gen}$ | beta-binomial |

Figure 5.1: Experimental variables (domain, task, and model set) explored.

## 5.1. Experimental Procedure

This section details the learning examples and inductive learning system Seer used in the experiments. It details the *domain* and *task* experimental variables. The *model set* experimental variable is the subject of chapter 4. Also, the section discusses how results were measured and statistical significance determined.

### 5.1.1. Experimental Variable: Domain

The experiments were run on classified learning examples from three domains, the same as were used in Shavlik *et al*. [1991] The sets used were:

**Soybean disease diagnosis ("Soy")**

These classified examples come originally from R.S. Michalski and R.L. Chilausky [1980]. The collection consists of 307 examples with 35 attributes labeled with one of 19 classes. The classified examples are available on the Internet in `<ftp://ics.uci.edu/pub/machine-learning-databases/soybean/>`.

**Heart disease diagnosis ("Heart")**

Robert Detrano, M.D., Ph.D., created these classified examples at the V.A. Medical Center, Long Beach and Cleveland Clinic Foundation. They were originally used in Sandhu *et al.*, [1989]. The examples consist of 303 examples with 13 attributes and 2 classes. The classified examples are available on the Internet in `<ftp://ics.uci.edu/pub/machine-learning-databases/heart-disease/>.`

**Audiological problems ("Audio")**

Professor Jergen developed these classified examples at the Baylor College of Medicine. They were originally used in Bareiss & Porter [1987]. The examples consist of 226 examples with 69 attributes labeled with one of 24 classes. The examples are available on the Internet in `<ftp://ics.uci.edu/pub/machine-learning-databases/audiology/>.`

### 5.1.2. Experimental Variable: Task

Seer was tested on two tasks: characterization and prediction. For the characterization task, all available classified learning examples were used to generate learning-performance data (using the generalized-cross-validation procedure described in Section 3.2). From each model set, Seer found the maximum-likelihood model. Success at characterization was measured (as detailed in Section 5.1.4) by how well a model fit the data in terms of loglikelihood. In order to see the variation caused by generalized cross-validation, three trials were run for each combination of domain and model set.

For the prediction task, 100 classified learning examples were randomly selected (without replacement) from all the available examples. Using generalized cross-validation on the 100 examples, Seer produced learning-performance data. For each model set, Seer found the model with maximum likelihood with respect to the learning-performance data. Success at prediction was measured by how well the model, which was based on only 100 classified examples, predicted learning-performance when the learner was given the entire data set. Figure 5.2 illustrates the process. To see the variation caused by the selection of the 100 examples and by generalized cross-validation, 30 trials were run for each combination of domain and model set. Section 5.1.4 details how Seer compared the models to observed learning performance.

### 5.1.3. Inductive Learning Program

All the experiments reported here used the C4.5 machine-learning program [Quinlan, 1992]. C4.5 does its initial learning with decision trees and then tries to improve its generalization and noise handling by turning the trees to rules. It was run with its default settings.

### 5.1.4. Measuring Results

Results were measured primarily with loglikelihood:

$$L_1 = \log \Pr\left(D \mid M_1\right) \tag{5.1}$$

where Pr is probability, $D$ is observed learning-performance data, and $M_1$ is a learning-performance model. Because probably (or likelihood) ranges between 0 and 1, loglikelihood is a negative number. One disadvantage of the loglikelihood measure is that it depends on the amount of learning-performance data. Doubling the amount of data tends to double the value of the loglikelihood. This is a common problem in the field of statistics. The usual solution is to develop a normalized measure of it. By analogy to the $R^2$ measure of linear regression, these measures are usually named $R^2_X$ where "X" is some identifier. The rest of these section will develop an $R^2_X$ measure for use on learning-performance model.

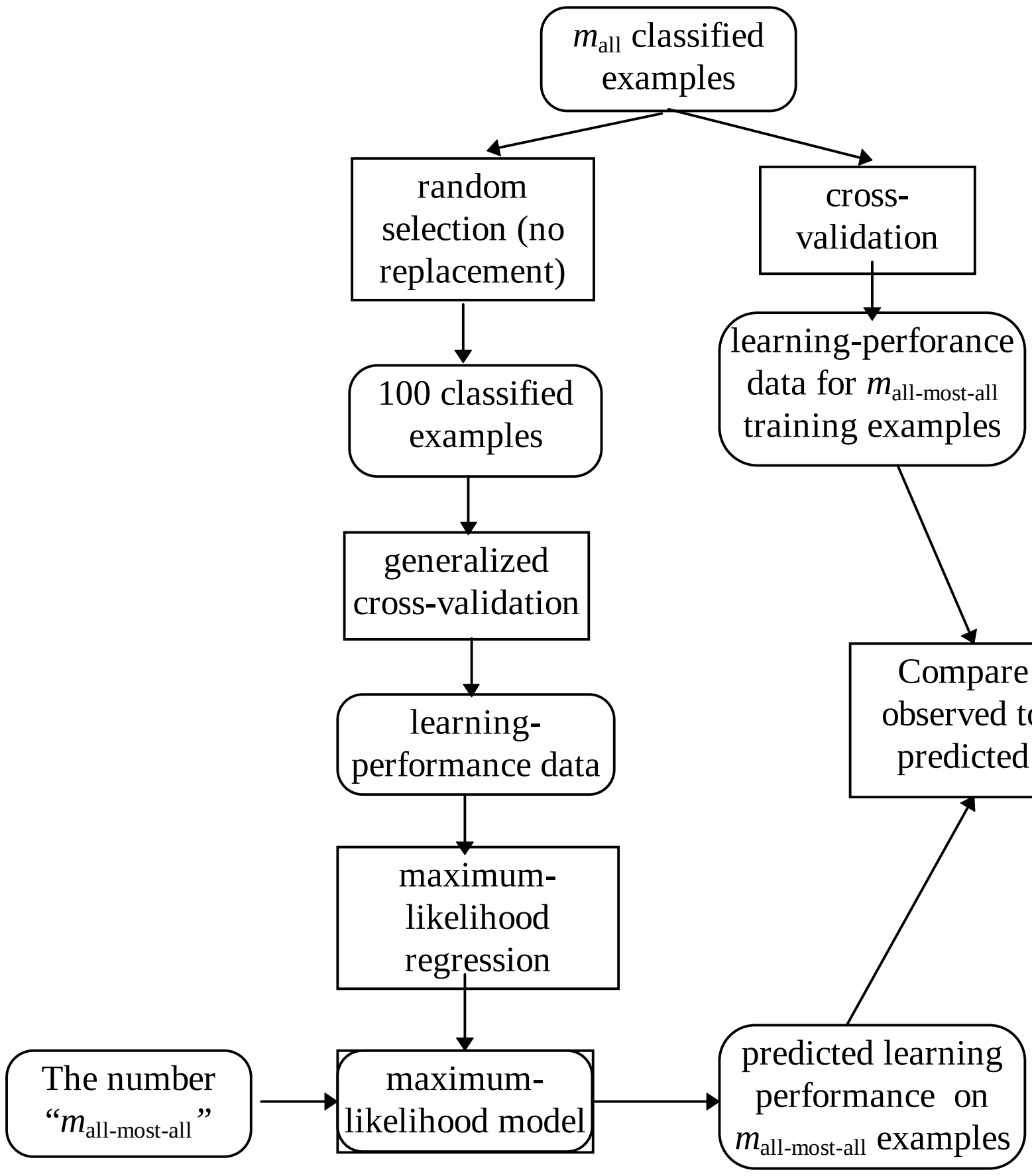

Figure 5.2: Prediction task -- From 100 classified examples, Seer predicts the learning performance if all most all available [l] classified examples were available, where $m_{all-most-all}$ =300, 290, 218 for the Soy, Heart, and Audio domains, respectively.

### 5.1.4.1. Input to the $R^2_X$ Measure

The measures needs to take as input:

1. A learning-performance model or, if repeated experiments are conducted, a set of learning-performance models. This model is produced by a system such as Seer from a small set of classified examples.

2. A likelihood criterion, for example, the binomial distribution

3. Learning-performance data based on a larger set of classified examples -- Each datum is of the form <*m*, *y*, *k*>, where *m* is the number of training examples, *y* is the number of testing examples the learner's hypothesis classified correctly, and *k* is the number of testing examples.

For example, suppose we wish to compare "A" and "B", two ways of producing learning-performance models. They could, for example, could select the maximum-likelihood model from two sets of candidate models. Suppose that we give each method three sets of learning-performance data, each based on 100 classified examples, and call the resulting models "A1," "A2," "A3," "B1," "B2," and "B3.". Next, we ask each of the six models to predict the learning performance when *m* is 300. Suppose they predict: $p_{A1} = 0.89$, $p_{A2}$=.091, $p_{A3}$=.095, $p_{B1} = 0.81$, $p_{B2}$=0.85, $p_{B3}$=0.89. To create the final piece of input information, we put the predictions to the test. For example, suppose we have a total of 400 examples and do 4-way cross-validation with these results:

$$\{<300, 92, 100>, <300, 100, 100>, <300, 96, 100>, <300, 89, 100>\} \quad (5.2)$$

where the first element of each tuple is the number of training examples, the second element is the number of correctly classified testing examples, and the third element is the total number of testing examples.

### 5.1.4.2. Output of the Measure

Intuitively, the output should be a better measure for better predictions. Hosmer and Lemeshow [1989, p. 148] recommends this measure that returns a value between 0 and 1 for logistic regression:

$$R^2{}_L = \frac{L_0 - L_1}{L_0 - L_S} \quad (5.3)$$

$L_1$ is the loglikelihood of the model of interest. $L_0$ is the loglikelihood of the baseline model, the model that predicts the same accuracy regardless of *m*. $L_S$ is the loglikelihood of the perfect, *saturated* model. The saturated model is the model with as many parameters are there are distinct values of *m;* for each *m*, it returns exactly the frequency observed. It is the optimum model with respect to likelihood (assuming error is measured with a distribution such a the binomial or the normal).

For our purposes, $R^2_L$ is unsatisfactory. The problem is that it assumes that $L_0$, the log likelihood of the baseline model has a worse fit than $L_1$. Because our model is being tested on learning data it has not seen before, the assumption does not necessarily hold.

A new measure that avoids the problems and is, thus, more suitable is:

$$R^2_{Seer} = L_s / L_1 \tag{5.4}$$

Returning to the example, if the likelihood criterion is the binomial distribution, loglikelihood is:

$$\mathrm{ll}_{\mathrm{binom}}(p, y, k) = y \log(p) + (k - y)\log(1 - p) + \log(\mathrm{binomial}(k, y)) \tag{5.5}$$

where *p* is predicted accuracy, *y* is the number of testing examples classified correct, and *k* is the total number of testing examples. Hosmer's measure can ignore the combinatorial term binomial(*k*,*y*) because the term cancels out. The term does not cancel out of the $R^2_{Seer}$ measure and so must be considered.

The loglikelihood of model "A1" with respect to observations is:

$$l1_{A1} = ll_{binom}(0.89,92,100) + ll_{binom}(0.89,100,100) + ll_{binom}(0.89,96,100) + ll_{binom}(0.89, 89,100) = -20.9849 \tag{5.6}$$

The value of the saturated model when *m*=300 is minimized (for the binomial likelihood criterion) when the saturated model predicts the mean observed frequency [Cramer, 1986, p. 159], which in this case is 0.9425.

The loglikelihood of this saturated model is:

$$ll_{binom}(0.9425,92,100) + ll_{binom}(0.9425,100,100) + ll_{binom}(0.9425,96,100) + ll_{binom}(0.9425, 89,100) = -14.2974 \tag{5.7}$$

So $R^2_{Seer}$ is -14.2974/-20.9849 = 0.681316.

The loglikelihood and $R^2_{Seer}$ values for all of the example models are given in Table 5.1.

| Model | Loglike. | $R^2_{Seer}$ |
|---|---|---|
| A1 | -20.9849 | 0.681316 |
| A2 | -17.2222 | 0.830170 |
| A3 | -14.5238 | 0.984412 |
| B1 | -43.9230 | 0.325510 |
| B2 | -31.1878 | 0.458429 |
| B3 | -20.9849 | 0.681316 |

Table 5.1: $R^2_{Seer}$ values for six models

### 5.1.4.3. Composite $R^2_{Seer}$ Results

The $R^2_{Seer}$ measure can be generalized to produce a single measure for a group of related models, using the fact that the loglikelihood of a set of independent models is the sum of the loglikelihood of each.

$$R^2_{Seer} = \frac{\sum \text{loglikelihood of saturated models}}{\sum \text{loglikelihood of models}} \tag{5.8}$$

Applied to our example, this yields:

$$R^2_{Seer}(\text{A's}) = \frac{-14.2974 + -14.2974 + -14.2974}{-20.9849 + -17.2222 + -14.5238} = 0.813417 \tag{5.9}$$

and

$$R^2_{Seer}(\text{B's}) = \frac{-14.2974 + -14.2974 + -14.2974}{-43.923 + -31.1878 + -20.9849} = 0.446349 \tag{5.10}$$

### 5.1.5. Comparing Results from Two Model Sets

Two methods of generating model sets can be compared by looking at composite $R^2_{Seer}$ values, but it is hard to evaluate the statistical significance of the values. Seer solves this problem by using paired-difference analysis, a standard statistical method [McClave & Dietrich, 1988]. With paired-difference analysis, each model-generation method is compared, head-to-head, in a series of trials In each trial, every individual model-generation method is applied to the same set of learning-performance data. As Table 5.2 illustrates -- using the example from Section 5.1.4 -- the loglikelihoods of the two models are found and the difference is calculated. Finally, these difference values are evaluated with the two-tailed Student *t*-test to determine if the mean of the differences is significantly different than zero.

| | A | B | |
|---|---|---|---|
| Trial | Loglike. | Loglike. | Diff. |
| 1 | -20.9849 | -43.9230 | 22.9381 |
| 2 | -17.2222 | -31.1878 | 13.9656 |
| 3 | -14.5238 | -20.9849 | 6.4611 |

Table 5.2: The first step of paired-difference analysis: For each trial's learning-performance data set, find the loglikelihood of the model produced by each of the two model-generation methods. Then, calculate the difference. The set of differences is evaluated with the two-tailed Student *t*-test to see if it is different than zero.

## 5.2. Experimental Results

Each series of experiments tested a set of learning-performance models The goals of this chapter were to:

1) Find which (if any) of the model sets' designs were useful for characterizing and predicting learning performance, and

2) Help develop criteria for choosing which of the useful models to use in a particular situation.

### 5.2.1. Modelgen Experiments

The Modelgen experiments tested the set of learning-performance models described in Table 5.3. The set's deterministic component is made up of the $model_{gen}$ and $Burr_{1.}$ Its nondeterministic component is the binomial distribution. The three goals of the Modelgen experiments were 1) to measure how well $model_{gen}$ can characterize machine learning performance data 2) to measure how well $model_{gen}$ can predict machine learning performance 3) to establish a benchmark for the other experiments.

| Com-ponent | Sub-model | Value |
|---|---|---|
| Det. #1 | $\text{model}_{\text{gen}}$ | $p = \text{model}_{\text{gen}}\left[(m-1)/d+1, start, skew, max\right]$ $= start + (max - start)\dfrac{\text{model}_5\left[(m-1)/d+1\right]^{skew(\text{Log}_{1/2}[start/2]-1)} - 0.5^{skew(\text{Log}_{1/2}[start/2]-1)}}{1-0.5^{skew(\text{Log}_{1/2}[start/2]-1)}}$ |
| Det. #2 | $\text{Burr}_1$ | $\text{model}_5[z] = z/(z+1)$ |
| Nondet. | binomial | $p^y (1-p)^{(k-y)} \binom{k}{y}$ |

Table 5.3: Learning-performance model set tested in the Modelgen experiments

#### 5.2.1.1. Characterizing Data with Modelgen

Figure 5.3 shows the first of three characterization trials for each of the three domain. The characterization task can be thought of as 1) generating accuracy points and 2) fitting a curve to those points. Table 5.4 quantifies the goodness-of-fit for all 3 trials in all 3 domains. The $R^2_{\text{Seer}}$ values are near 1.0 indicating that $\text{model}_{\text{gen}}$ can fit learning-performance data well. (To get a perfect $R^2_{\text{Seer}}$ value of 1.0, a learning-performance model would need to go through every accuracy point, which is generally undesirable).

| Domain | Loglike. | $R^2_{\text{Seer}}$ |
|---|---|---|
| Soy (3) | -7654.25 | 0.974 |
| Heart (3) | -7568.45 | 0.984 |
| Audio (3) | -5582.62 | 0.955 |

Table 5.4: Goodness-of-fit for Modelgen on the characterization task -- Fit is measured with loglikelihood and $R^2_{\text{Seer}}$. A perfect fit has a $R^2_{\text{Seer}}$ value of 1.0. All measures are composites of three trials.

#### 5.2.1.2. Predicting in the Audiological Domain

The next several subsections look at prediction with $\text{model}_{\text{gen.}}$ We focus first on just the Audio domain. In each of the 30 trials, Seer was given 100 classified examples. From those, it tried to answer two questions: 1) what accuracy could be expected with 218 training examples, and 2) what accuracy could be expected given an infinite number of training examples (asymptotic accuracy).

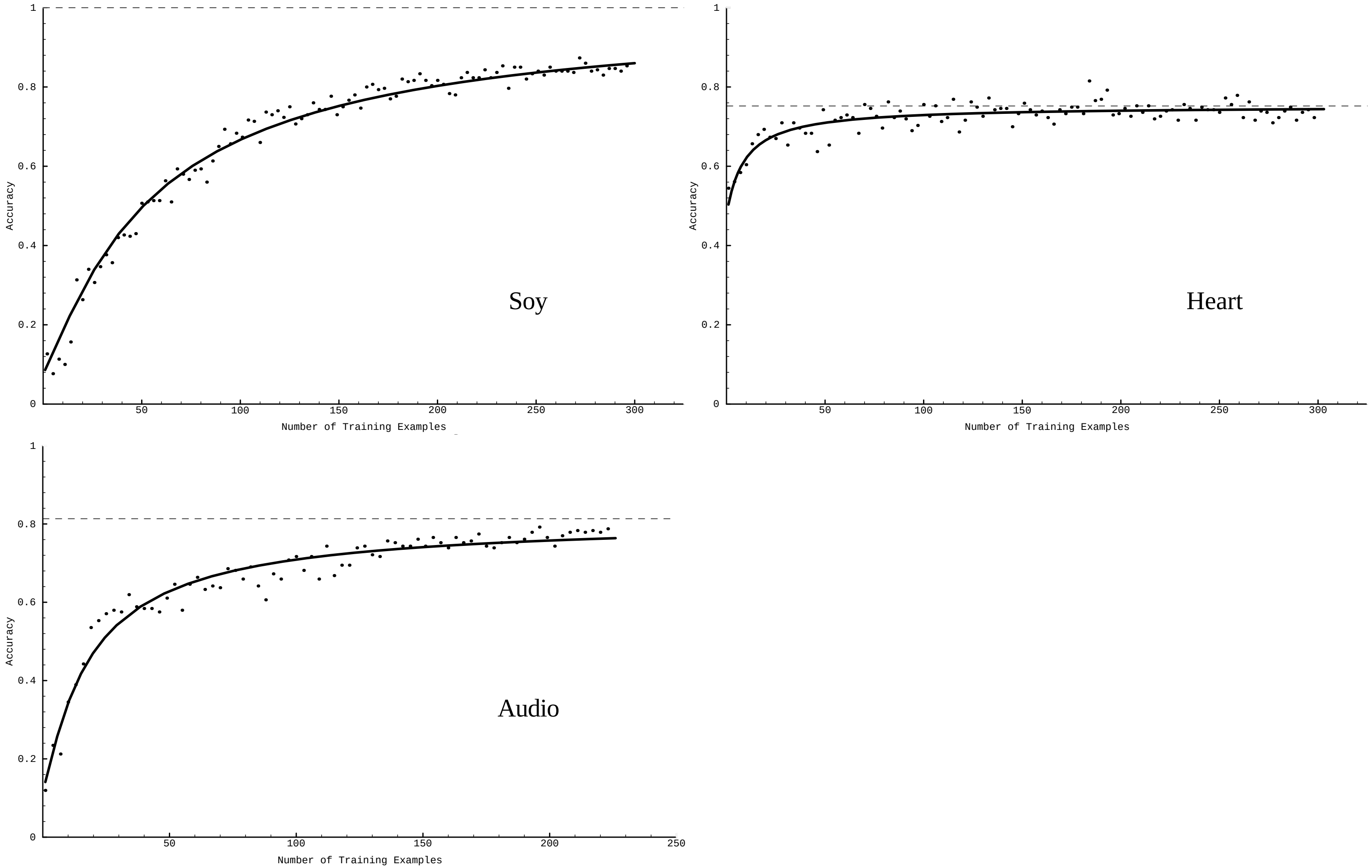

Figure 5.3: Characterization trial #1 for the three domains using Modelgen -- The points show the learning-performance data generated with generalized cross-validation. The solid curve is the deterministic part of the maximum likelihood Modelgen model. The dotted line is the estimated asymptotic accuracy, the accuracy given an infinite number of training examples.

Seer's first step toward answering these questions is to generate learning-performance data. The plots of Figure 5.4 represents the learning-performance data (for 3 of the trials) as accuracy points. (We will see in the next section that these 3 trials were the best, median, and worst of the 30 trials, as measured by prediction accuracy.)

The curves in Figure 5.5 show Seer's predictions for these 3 trials. The dotted lines show its estimate of the asymptotic accuracy. In the plots, the diamond shows "true" accuracy given 218 training examples. (Empirically, determining the real true accuracy would require an infinite number of testing examples; This "true" value is determined with cross-validation on all 226 available classified examples.)

Table 5.5 shows the accuracies Seer predicts would be observed if the learner was given 218 training examples or an infinite number of examples. Thirty trials were run; the best, median, and worst trial are shown in the table. The "true" accuracy, if an infinite number of training examples were provided, is estimated by the model fit on 226 examples in Section 5.2.1.1.

| | $p_{218}$ | $p_{\infty}$ |
|---|---|---|
| Best | 0.788 | 0.862 |
| Median | 0.717 | 0.757 |
| Worst | 0.606 | 0.619 |
| "True" | 0.788 | 0.823 |

Table 5.5: Modelgen's predicted accuracies (based on 100 classified examples) if the learner is given 218 or an infinite number of training examples -- The predictions from the best, median, and worst trials are shown. The "true" values are estimated from the full set of 226 classified examples.

Table 5.6 first quantifies how well the models fit the learning-performance data (shown in Figure 5.5) generated from the 100 classified examples. The models do well and have a composite $R^2_{Seer}$ of 0.981. Second, the table quantifies how well the models predicted $p_{218}$. Here they do less well, with a composite $R^2_{Seer}$ of 0.925. This suggests that the problem may not be with prediction given the 100 classified examples, but rather that the 100 classified examples are sometimes not representative of the full set of 218 classified examples We will further investigate this problem in the next section.

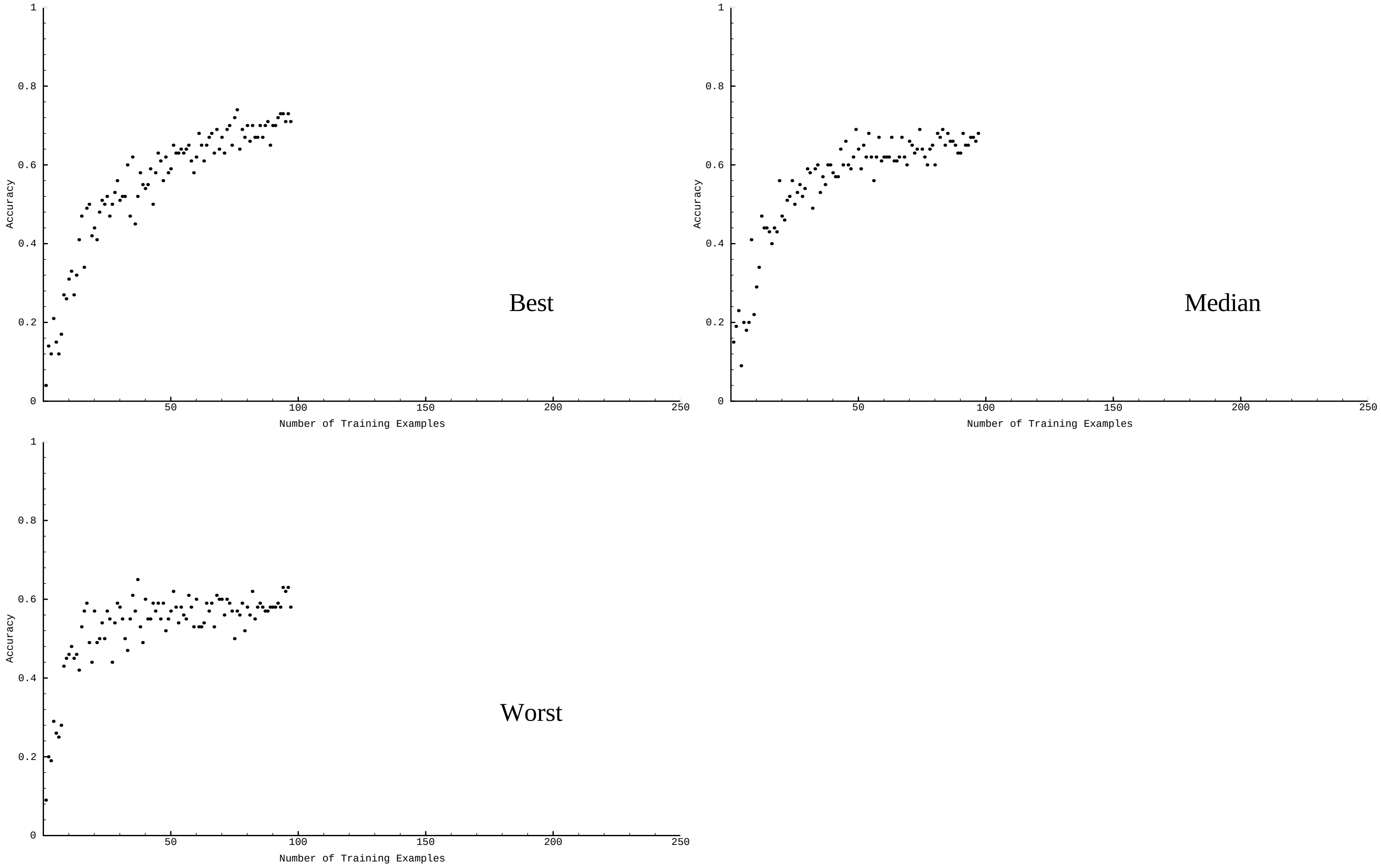

Figure 5.4: Audio-domain learning-performance data based on 100 classified examples for the best, median, and worst Modelgen trials -- Seer must find the Modelgen model that best fits each set of data. This maximum-likelihood model is used to predict the expected learning performance if more training examples were available.

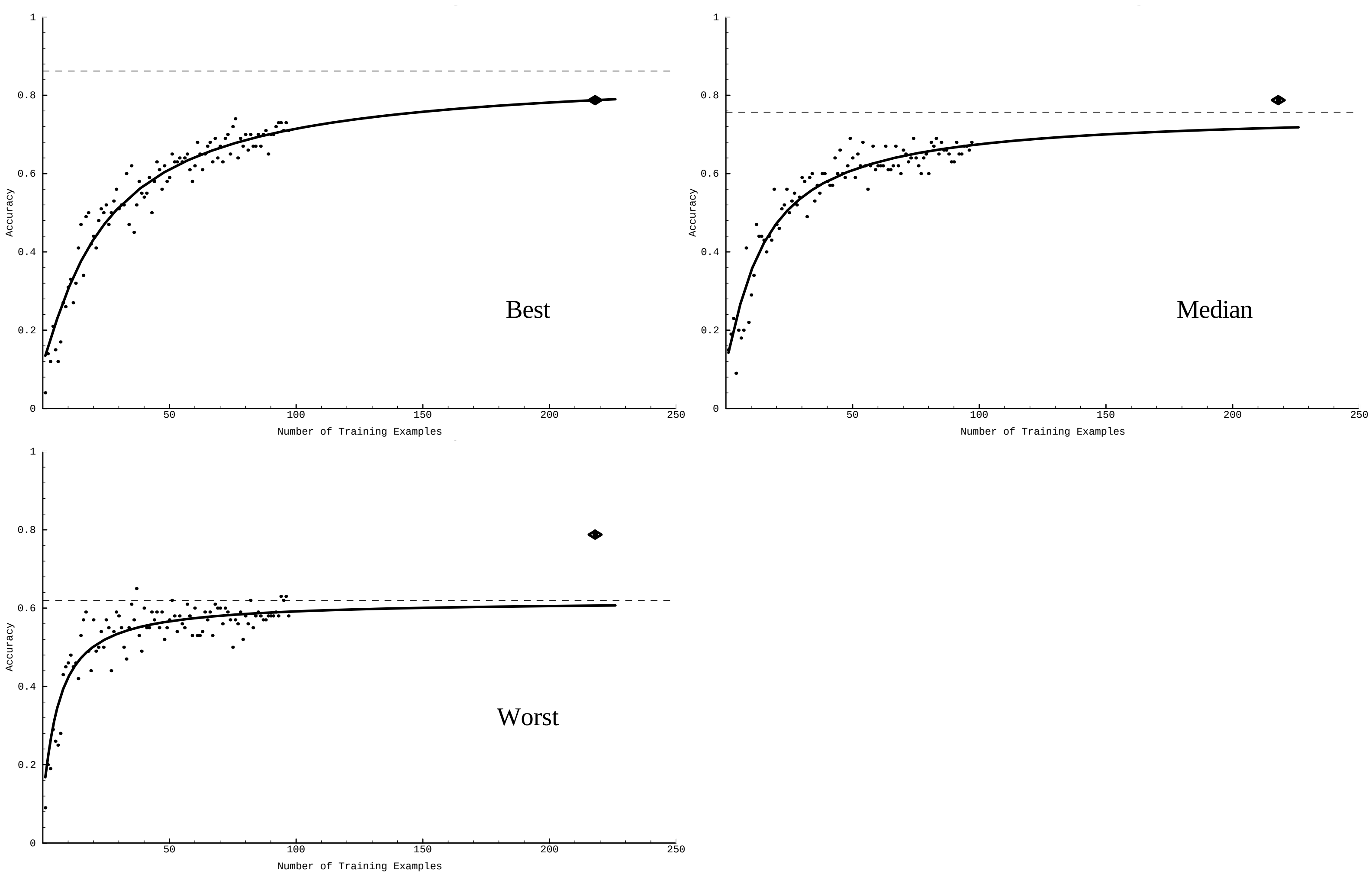

Figure 5.5: Audio-domain maximum-likelihood models for the best, median, and worst trials for Modelgen -- The solid curves are the deterministic models. The dotted lines are the estimated asymptotic accuracies. The diamond shows the “true” accuracy given 218 training examples.

| | Characterize the 100 | | Predict $p_{218}$ | |
|---|---|---|---|---|
| | Loglike. | $R^2_{Seer}$ | Loglike. | $R^2_{Seer}$ |
| Best | -1838.29 | 0.979 | -43.81 | 1.000 |
| Median | -1847.15 | 0.980 | -46.76 | 0.937 |
| Worst | -1902.73 | 0.981 | -60.74 | 0.721 |
| All (30) | -55923.45 | 0.981 | -1420.50 | 0.925 |

Table 5.6: Modelgen goodness-of-fit on learning-performance data from 1) the given 100 classified examples and 2) the full data set -- The better the prediction, the closer the "Predict $p_{218}$" $R^2_{Seer}$ value will be to 1.000.

### 5.2.1.3. Predicting in All Three Domains

This subsection looks at prediction with $model_{gen}$ on all three domains. The curves in Figure 5.6 show Seer's predictions for all 30 trials. The plots provide a graphical way to see the variation caused, mostly, by starting with different sets of 100 classified examples. In the plots, the diamond shows "true" accuracy given nearly all the available classified examples as training examples. In all three domains, some models predict too high, others too low. Particularly in the Soy and Heart domains, the models seem unbiased; that is, they are as likely to go too high as to go too low.

Table 5.7 quantifies the characterization and prediction in all three domains. For the Soy domain, the composite $R^2_{Seer}$ for prediction, 0.982, is high, higher even than the $R^2_{Seer}$ for characterization. For the Heart domain, the $R^2_{Seer}$ for characterization is higher, 0.978, and the $R^2_{Seer}$ for prediction is a little bit lower, 0.963. For the Audio domain, as we saw in the previous section, the $R^2_{Seer}$ for characterization is high, but the $R^2_{Seer}$ for prediction is relatively low.

| | Characterize the 100 | | Predict $p_{\text{all most all}}$ | |
|---|---|---|---|---|
| Domain | Loglike. | $R^2_{Seer}$ | Loglike. | $R^2_{Seer}$ |
| Soy (30) | -54507. | 0.969 | -1664.62 | 0.982 |
| Heart (30) | -54166. | 0.978 | -1446.20 | 0.963 |
| Audio (30) | -55923. | 0.981 | -1420.50 | 0.925 |

Table 5.7: Modelgen's goodness-of-fit on learning-performance data from 1) the given 100 classified examples and 2) the full data set -- The better the prediction, the closer the "Predict $p_{218}$" $R^2_{Seer}$ value will be to 1.000. All values are composites from the 30 trials.

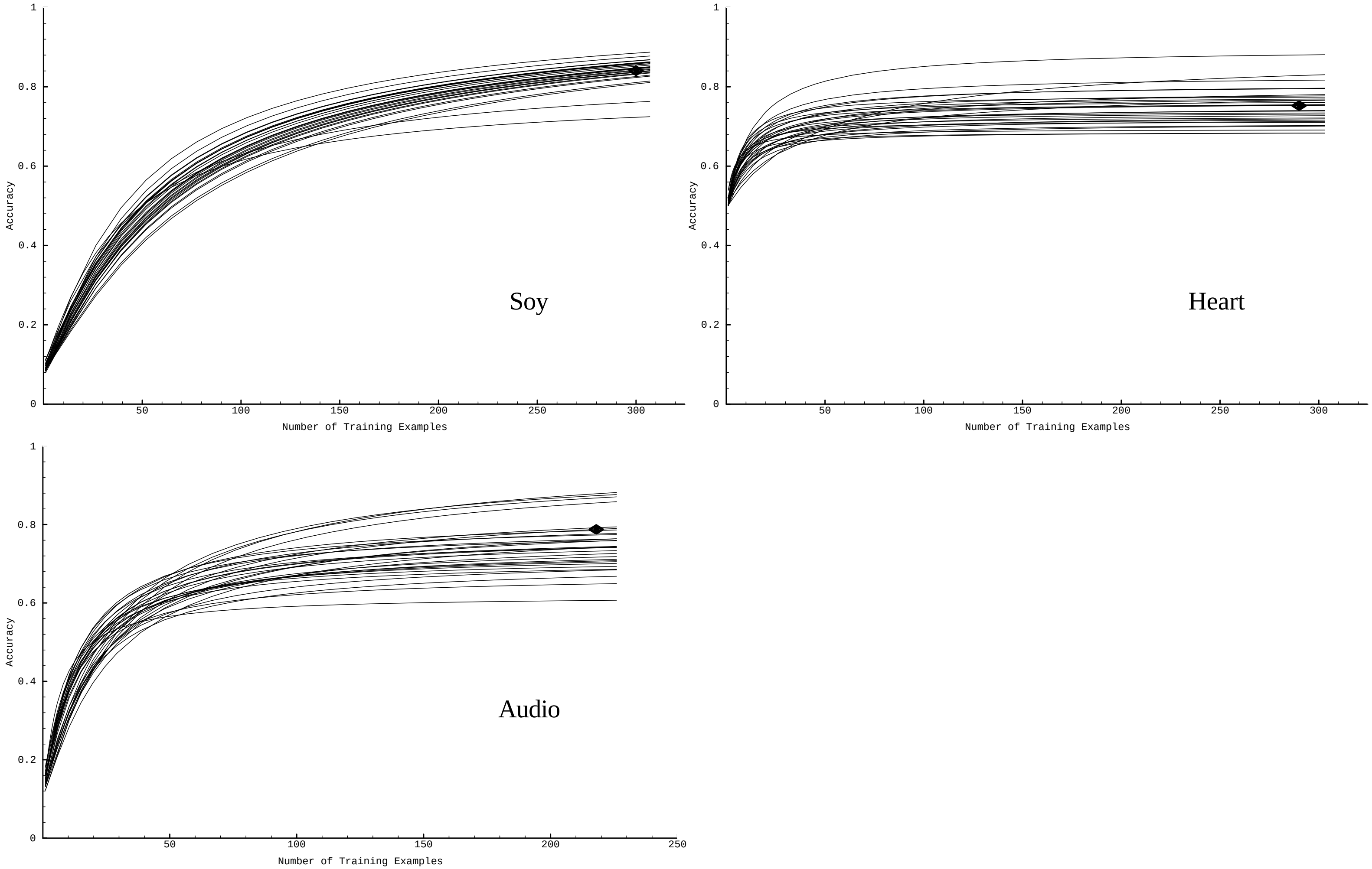

Figure 5.6: Modelgen's predictive models for 30 trials in 3 domains -- The diamond shows the "true" accuracy.

Why don't all the predictions hit the target exactly? We can get insight into this question by looking in more detail at the worst prediction in each domain. Figure 5.7 shows those predictions and the learning-performance data on which they are based. As the plots show and Table 5.8 confirms, even the worst-predicting models characterize the given learning-performance data well. The problem is that the learning-performance data from 100 classified examples is unrepresentative of the learning-performance of the full set of classified examples. This is most obvious in the Heart domain. Here, working with only 100 classified examples, Seer observed accuracies greater than 0.80, but working with the full set of 303 examples, the observed accuracy $p_{290}$ was 0.752. This strongly suggests that the main cause of bad predictions is that a random subset of classified examples may be unrepresentative of a larger set. Put more optimistically, Seer -- using $model_{gen}$ -- seems to be doing as well as possible with the classified learning examples it has available.

| | Characterize the 100 | | Predict $p_{\text{all most all}}$ | |
|---|---|---|---|---|
| Domain | Loglike. | $R^2_{Seer}$ | Loglike. | $R^2_{Seer}$ |
| Soy (worst) | -1855. | 0.979 | -66.28 | 0.822 |
| Heart (worst) | -1710. | 0.968 | -65.10 | 0.713 |
| Audio (worst) | -1903. | 0.981 | -60.74 | 0.721 |

Table 5.8: Modelgen's goodness-of-fit on learning-performance data from the worst trials

### 5.2.1.4. Estimating Parameters in All Three Domains

Another way to analyze the experiments is to plot the parameters estimated by each model. Because 30 trials were run for each domain, standard deviations can be estimated after the fact. The Modelgen model set has 4 parameters. Seer estimates two of the parameters, *start* and *skew,* from the class frequency of the classified examples. Seer estimates the other two parameters, *d* and *max,* from the learning-performance data. Figure 5.8 is a scatter plot for the Heart domain of the *d* and *max* parameters. In the plot, the letters "A", "B", and "C" are the parameter estimates in the 3 characterization trials. The numbers, 1 ... 30, are the parameters estimates for the 30 prediction trials. Note that the prediction points (which are based on only 100 classified examples) cluster well around the characterization points (which are based on the full sets of classified examples).

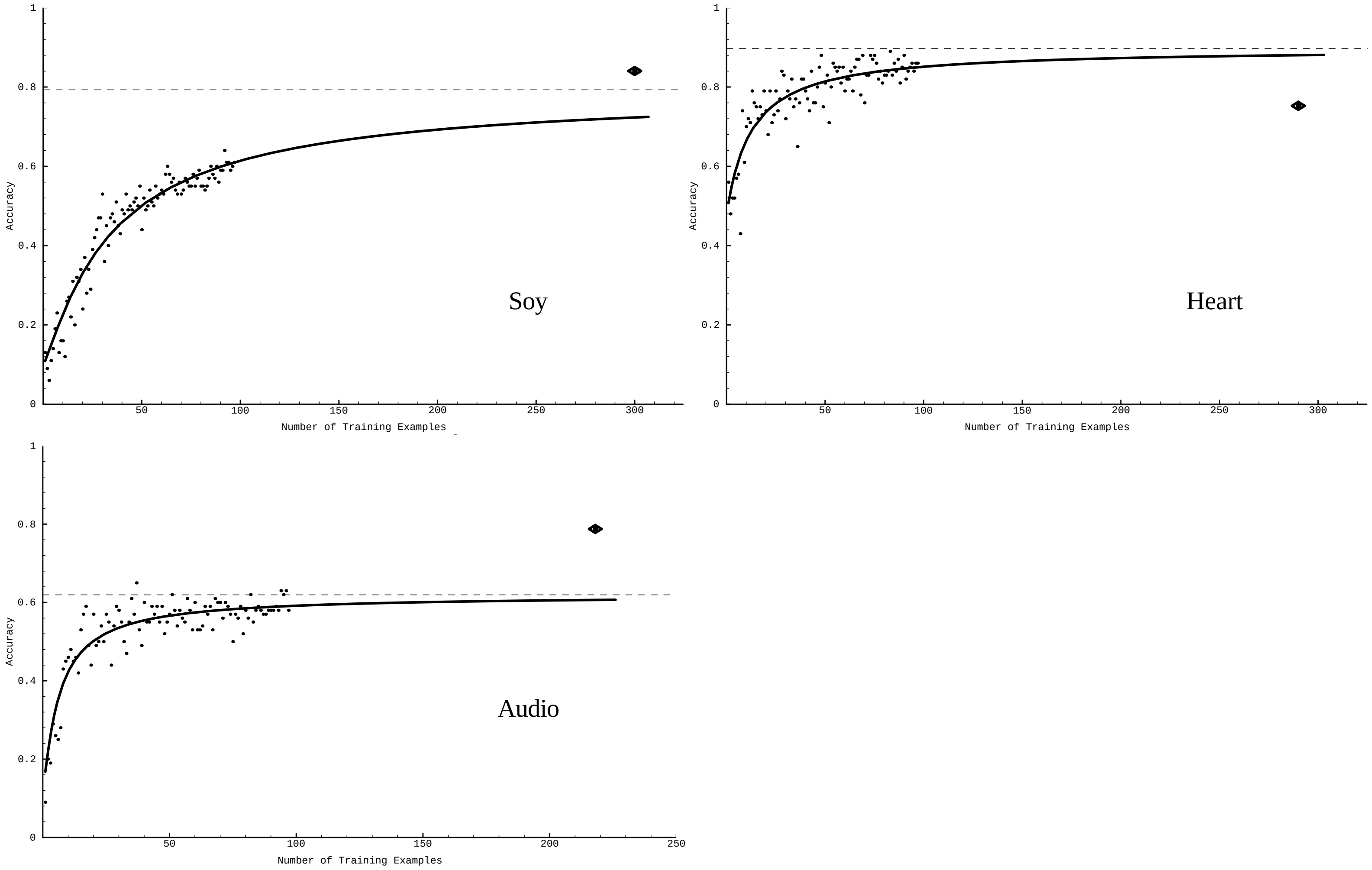

Figure 5.7: Modelgen maximum-likelihood models for the worst trials in the three domains -- The models, represented by the solid curves, fit the available learning-performance data well. The learning-performance data, however, is misleading, so the predictions of the accuracy to be expected on the set of training examples are inaccurate.

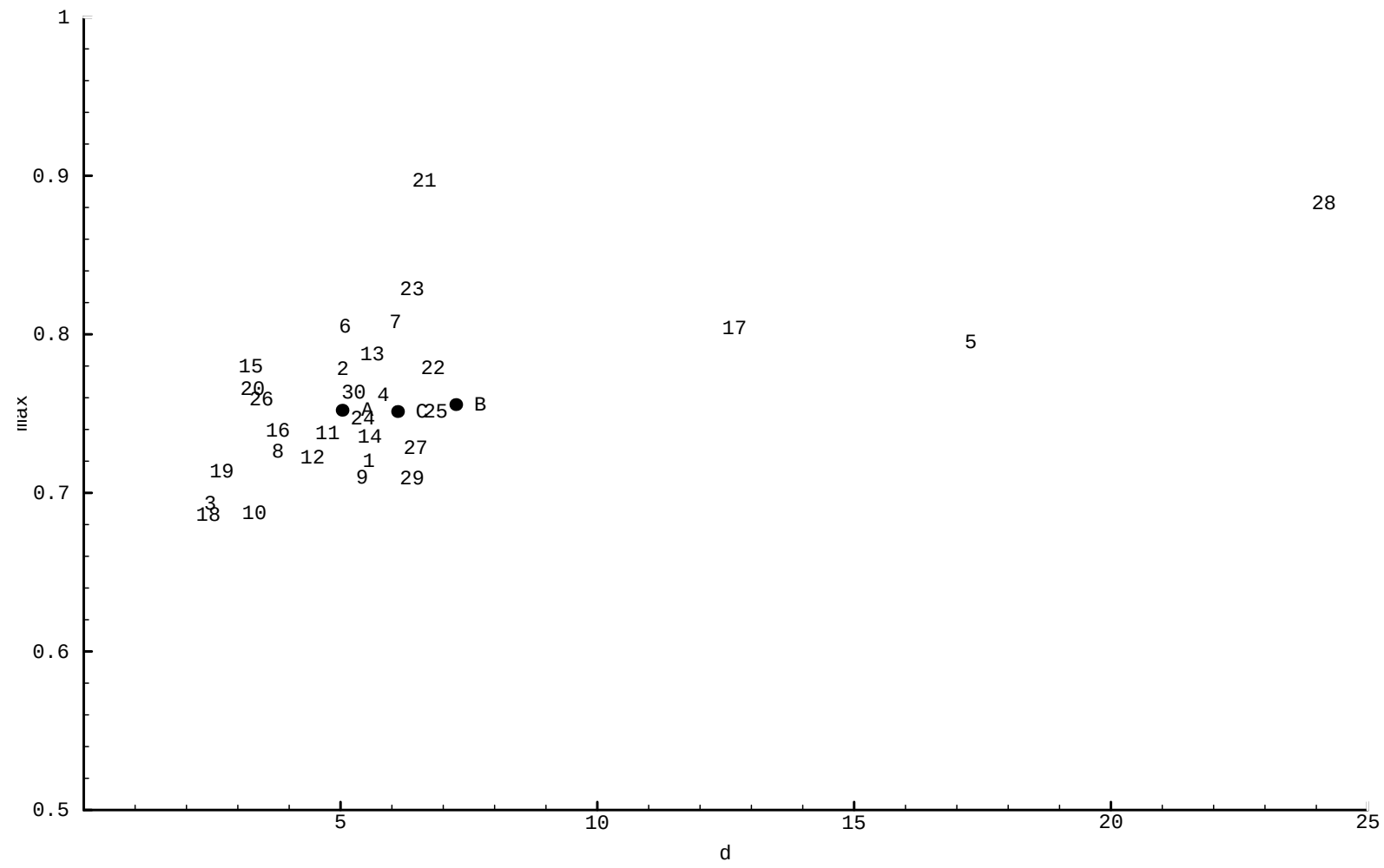

Figure 5.8: Scatter plot for the Heart domain of the *d* and *max* Modelgen parameters. The letters shows the parameter values for the 3 characterization trials. The numbers show the parameter values in for the 30 prediction trials.

Table 5.9 gives the mean and observed standard deviation for all parameters in all three domains. The Soy and Heart parameters estimates seem unbiased, clustering around the "true" parameter values. Audio's *d* estimate is off apparently because the estimates of *max* are too low, and this affects the estimated accuracy.

### 5.2.1.5. Summary of Modelgen experiments

These experiments showed that Modelgen can characterize learning-performance data well. The $R^2_{Seer}$ values for the three domains were 0.974, 0.984, and 0.955. The predictions were generally good, too, with composite $R^2_{Seer}$ values (over all 30 trials) for the three domains of 0.982, 0.963, and 0.925. In the worst of the trials the predictions were not as good. ($R^2_{Seer}$ values 0.822, 0.713, and 0.721.) Even in these cases, however, the Seer using Modelgen seemed to be doing about as well as possible. The problem seemed to be that some samples of classified examples were not representative of the entire set. The Soy and Heart parameter estimates seem unbiased, clustering around the "true" parameter values. Audio's *d* estimate is seemed to be systematically off, perhaps because its estimates of *max* were too low.

| | Characterization (all available classified examples, 3 trials) | | Prediction (100 classifies examples, 30 trials) | |
|---|---|---|---|---|
| | mean | s.d. | mean | s.d. |
| Soy | | | | |
| *d* | 12.680 | 0.1659 | 14.253 | 2.1784 |
| *max* | 0.999 | 0.0018 | 0.988 | 0.0459 |
| *start* | 0.087 | 0.0010 | 0.093 | 0.0089 |
| *skew* | 1.075 | 0.0019 | 1.081 | 0.0169 |
| $p_{300}$ | 0.840 | 0.0209 | 0.840 | 0.0321 |
| Heart | | | | |
| *d* | 6.129 | 1.1067 | 6.207 | 4.4863 |
| *max* | 0.753 | 0.0023 | 0.760 | 0.0516 |
| *start* | 0.503 | 0.0000 | 0.509 | 0.0110 |
| *skew* | 1.015 | 0.0000 | 1.042 | 0.0516 |
| $p_{290}$ | 0.752 | 0.0248 | 0.749 | 0.0452 |
| Audio | | | | |
| *d* | 7.718 | 4.3788 | 4.635 | 1.5372 |
| *max* | 0.884 | 0.1016 | 0.795 | 0.0855 |
| *start* | 0.141 | 0.0000 | 0.149 | 0.0162 |
| *skew* | 1.196 | 0.0000 | 1.191 | 0.0345 |
| $p_{218}$ | 0.788 | 0.0272 | 0.746 | 0.0650 |

Table 5.9: Mean and standard deviation of estimated Modelgen parameters

In the next series of experiments, logistic models were found for exactly the same learning-performance data as was generated for these Modelgen experiments. This will allow head-to-head comparison between Modelgen, a model set that was designed especially for learning prediction, and the Logistic model set, the most popular model set for binary-response data.

### 5.2.2. Logistic Experiments

The Logistic experiments tested the set of learning-performance models described in Table 5.12. The set's deterministic component is the logistic model. Its nondeterministic component is the binomial distribution. This experiment was designed to see how bad (or good) logistic models are for learning data. The model set was applied to the characterization task and prediction task, then it was compared to Modelgen, trial-by-trial.

| Component | Submodel | Value |
| --- | --- | --- |
| Det. | logistic | $p = \dfrac{e^z}{1+e^z}$ |
| Nondet. | binomial | $p^y (1-p)^{(k-y)} \binom{k}{y}$ |

Table 5.10: Learning-performance model set tested in the Logistic experiments

Table 5.11 and Figure 5.9 show the Logistic model set trying to characterize the learning-performance data from the three domains. The composite $R^2_{Seer}$ values reported in the table show that the fit is not good. In two domains, it is less than 0.9. The plots in Figure 5.9 (from trial #1, but the other two trials were similar) show the problem: the curve defined by logistic models are not the right shape for the learning-performance data. The Logistic model set only does best in the Heart domain and from the plot we can see the reason is that this data is the most linear and hence is easiest to model.

| Domain | Loglike. | $R^2_{Seer}$ |
| --- | --- | --- |
| Soy (3) | -9023.51 | 0.826 |
| Heart (3) | -7730.06 | 0.963 |
| Audio (3) | -5971.10 | 0.893 |

Table 5.11: Goodness-of-fit for Logistic on learning-performance data from the three domains

Given the problems the Logistic model set has predicting learning performance, we would expect it to do even worse at predicting from a subset of 100 classified examples. Table 5.12 and Figure 5.10 confirm this expectation. In all 30 trials and in all 3 domains, every prediction is off. Even worse, the predictions are very biased; every prediction is too optimistic. Even though all the $R^2_{Seer}$ values for characterization of the 100 examples is greater than 0.90, the best prediction $R^2_{Seer}$ value is a poor 0.501.

| | Characterize the 100 | | Predict $p_{\text{all most all}}$ | |
| --- | --- | --- | --- | --- |
| Domain | Loglike. | $R^2_{Seer}$ | Loglike. | $R^2_{Seer}$ |
| Soy (30) | -55725. | 0.947 | -7100. | 0.230 |
| Heart (30) | -54648. | 0.970 | -2778. | 0.501 |
| Audio (30) | -58556.99 | 0.937 | -2898. | 0.454 |

Table 5.12: Logistic's goodness-of-fit on learning-performance data from 1) the given 100 classified examples and 2) the full data set -- Values are composites from the 30 trials.

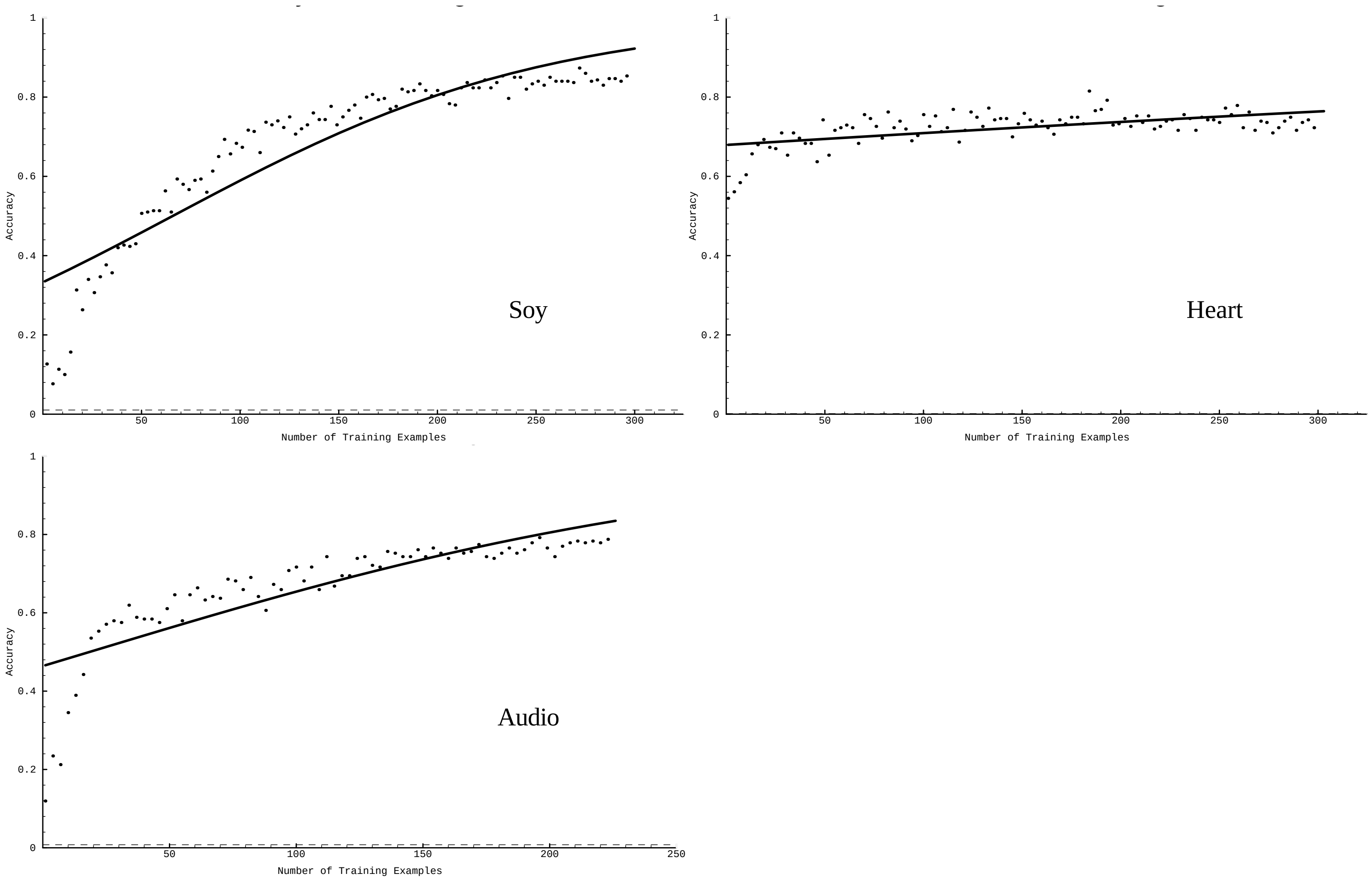

Figure 5.9: Characterizing learning-performance data with logistic models -- The curves defined by the models do not fit the data well.

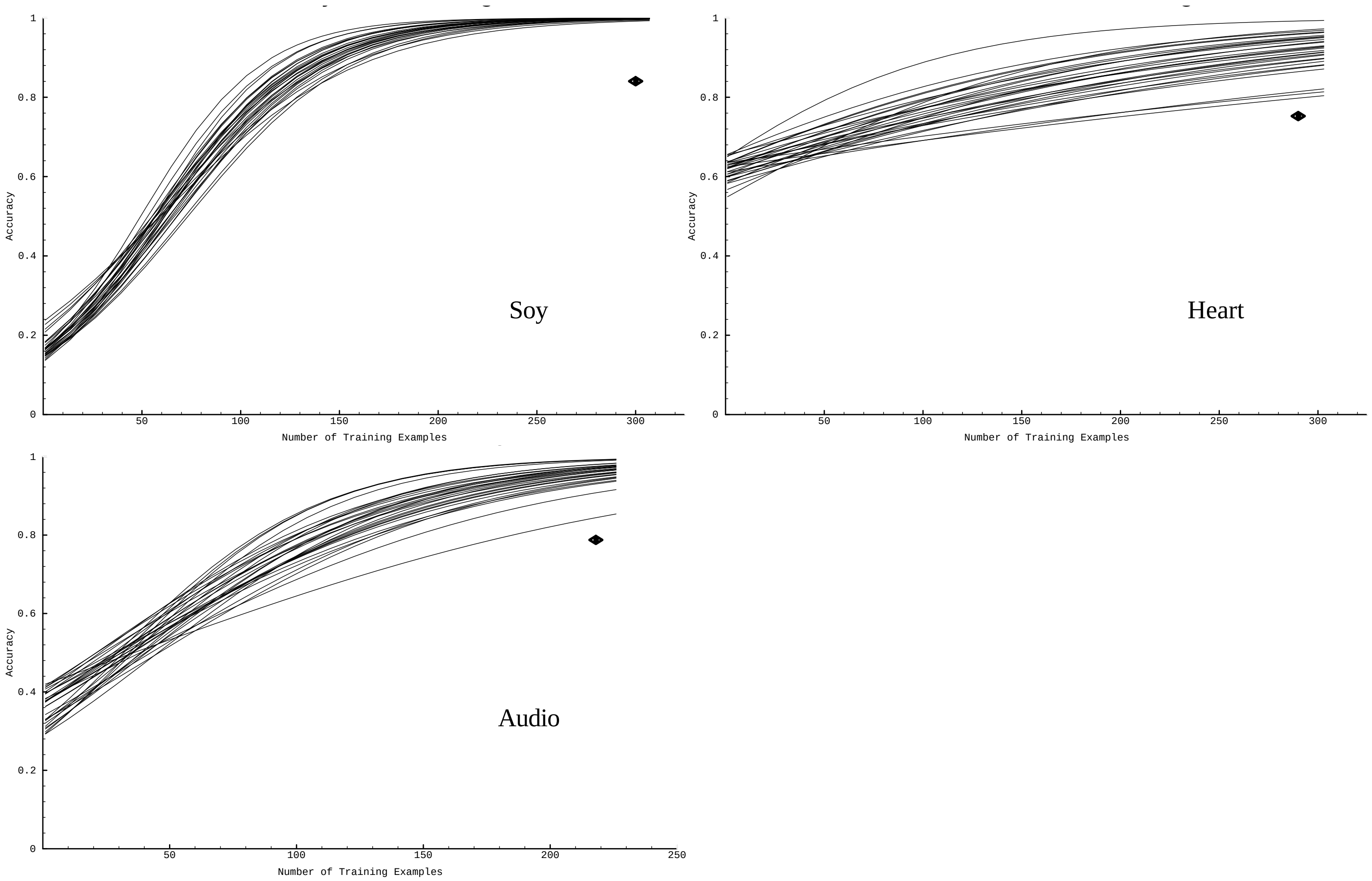

Figure 5.10: Logistic’s predictive models for 30 trials in 3 domains -- All 90 predictions are too high.

Table 5.13 compares, using paired-difference analysis, the performance of the Logistic model set to that of the Modelgen model set in all three domains. Both the performance characterizing the learning performance and performance predicting learning performance are shown. The table tells if one model set (Modelgen or Logistic) produced significantly better learning-prediction models than the other in each scenario. For Soy/Predict, for example, there were 30 learning-prediction problems. For every problem, the two model sets each produced a learning-prediction model. The loglikelihoods of these two learning-prediction models provide a measure of their goodness. A two-tailed *t*-test was used to test if the mean difference between the two loglikelihood values was significantly different from zero. For these tests an $\alpha$ value less than 0.05 was considered significant. For every combination of task and domain, the fit of the Modelgen models (as measured by loglikelihood) was on average better than the fit of the logistic models and this difference was statistically significant.

| Scenario | Task | Domain | # of Trials | Sig.Diff? | Better | $\alpha$ | test stat. |
|---|---|---|---|---|---|---|---|
| 1 | characterize | soy | 3 | yes | modelgen | <0.001 | 24.730 |
| 2 | characterize | heart | 3 | yes | modelgen | 0.001 | 12.946 |
| 3 | characterize | audio | 3 | yes | modelgen | <0.001 | 26.300 |
| 4 | predict | soy | 30 | yes | modelgen | <0.001 | 24.888 |
| 5 | predict | heart | 30 | yes | modelgen | <0.001 | 6.750 |
| 6 | predict | audio | 30 | yes | modelgen | <0.001 | 8.980 |

Table 5.13: Results of paired-difference analysis -- For all six combinations of task and domain, the Modelgen models produce statistically-significantly better loglikelihoods than the Logistic models.

**Analysis:** The logistic models characterized learning performance poorly and predicted learning performance even more poorly. Paired-difference analysis showed that the Modelgen models were better than the logistic models in every combination of task and domain investigated.

### 5.2.3. ED Experiments

The ED experiments tested the set of learning-performance models described in Table 5.14. The set's deterministic component is made up of the $\text{model}_{gen}$ and EDit. Its nondeterministic component is the binomial distribution. The purpose of this experiment was to see if the EDit model, which is based more closely on computational learning results, performs better than the simple $\text{Burr}_1$ model.

| Component | Submodel | Value |
|---|---|---|
| Det. #1 | $\text{model}_{\text{gen}}$ | $p = \text{model}_{\text{gen}}\left[(m-1)/d+1, start, skew, max\right] = start + (max - start)\dfrac{\text{model}_5\left[(m-1)/d+1\right]^{skew(\text{Log}_{1/2}[start/2]-1)} - 0.5^{skew(\text{Log}_{1/2}[start/2]-1)}}{1-0.5^{skew(\text{Log}_{1/2}[start/2]-1)}}$ |
| Det. #2 | EDit | $\text{model}_5[z] = \text{EDit}^{-1}[z]$, where $\text{EDit}[p] = \dfrac{2\log(\frac{6}{1-p})}{(1-\sqrt{1-p})(1-p)}$ |
| Nondet. | binomial | $p^y (1-p)^{(k-y)} \binom{k}{y}$ |

Table 5.14: Learning-performance model set tested in the ED experiments

Figure 5.11 shows a typical characterization trial. Table 5.15 quantifies all the characterization trials in all three domains. The ED model set does well; all $R^2_{\text{Seer}}$ values are greater than 0.95.

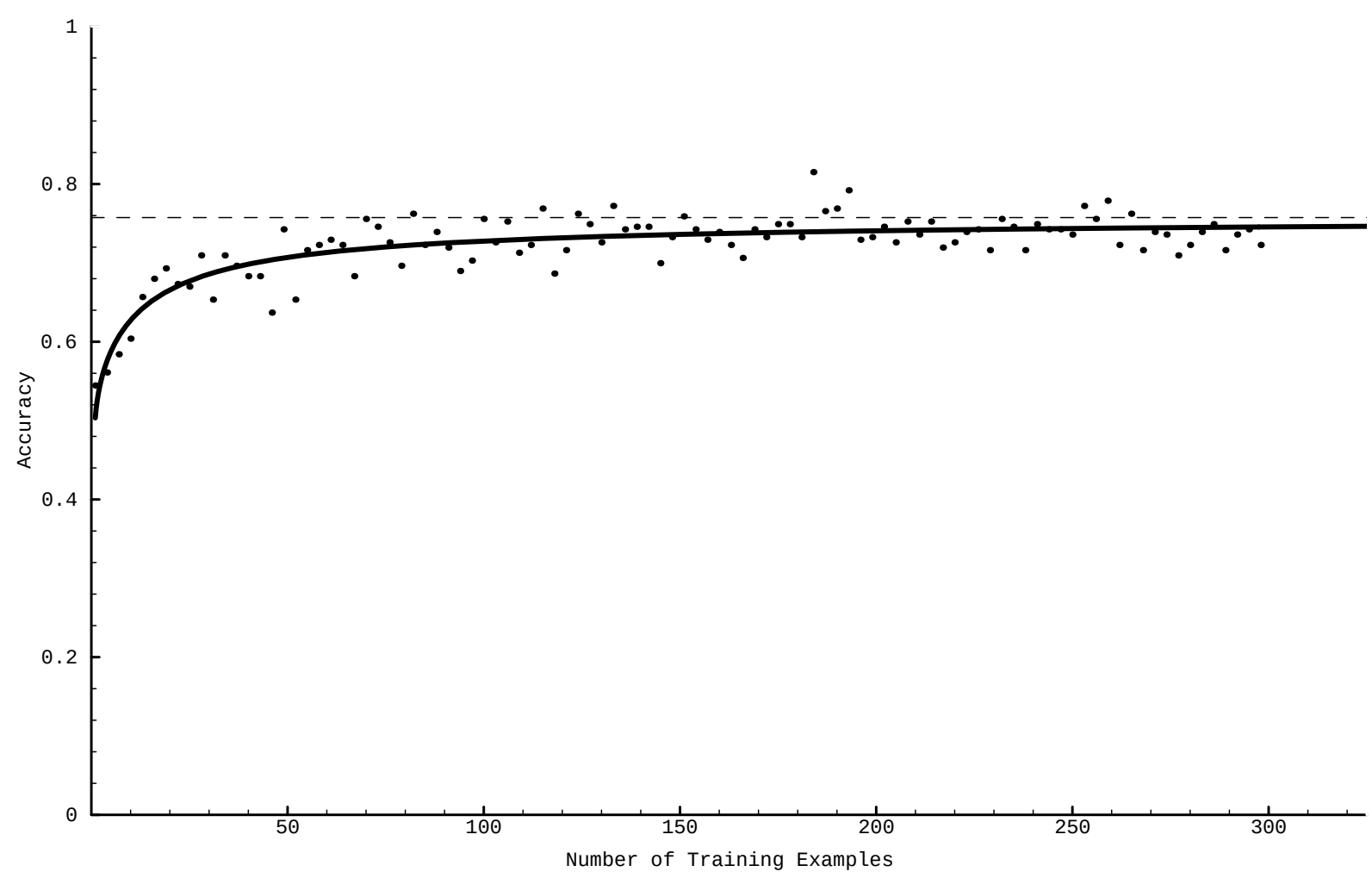

Figure 5.11: Characterizing learning-performance data from the Heart domain (Trial #1) using a ED model

| Domain | Loglike. | $R^2_{\text{Seer}}$ |
|---|---|---|
| Soy (3) | -7828.60 | 0.952 |
| Heart (3) | -7570.18 | 0.984 |
| Audio (3) | -5426.34 | 0.983 |

Table 5.15: Goodness-of-fit for ED predicting learning-performance data in three domains

Figure 5.12 shows all the models found for all 30 prediction trials in the Audio domain. As Table 5.16 shows, the ED model set predicts well based on a subset of examples. The lowest $R^2_{Seer}$ values as 0.939. Also, as the plot illustrates, the predictions are clustered around the true value.

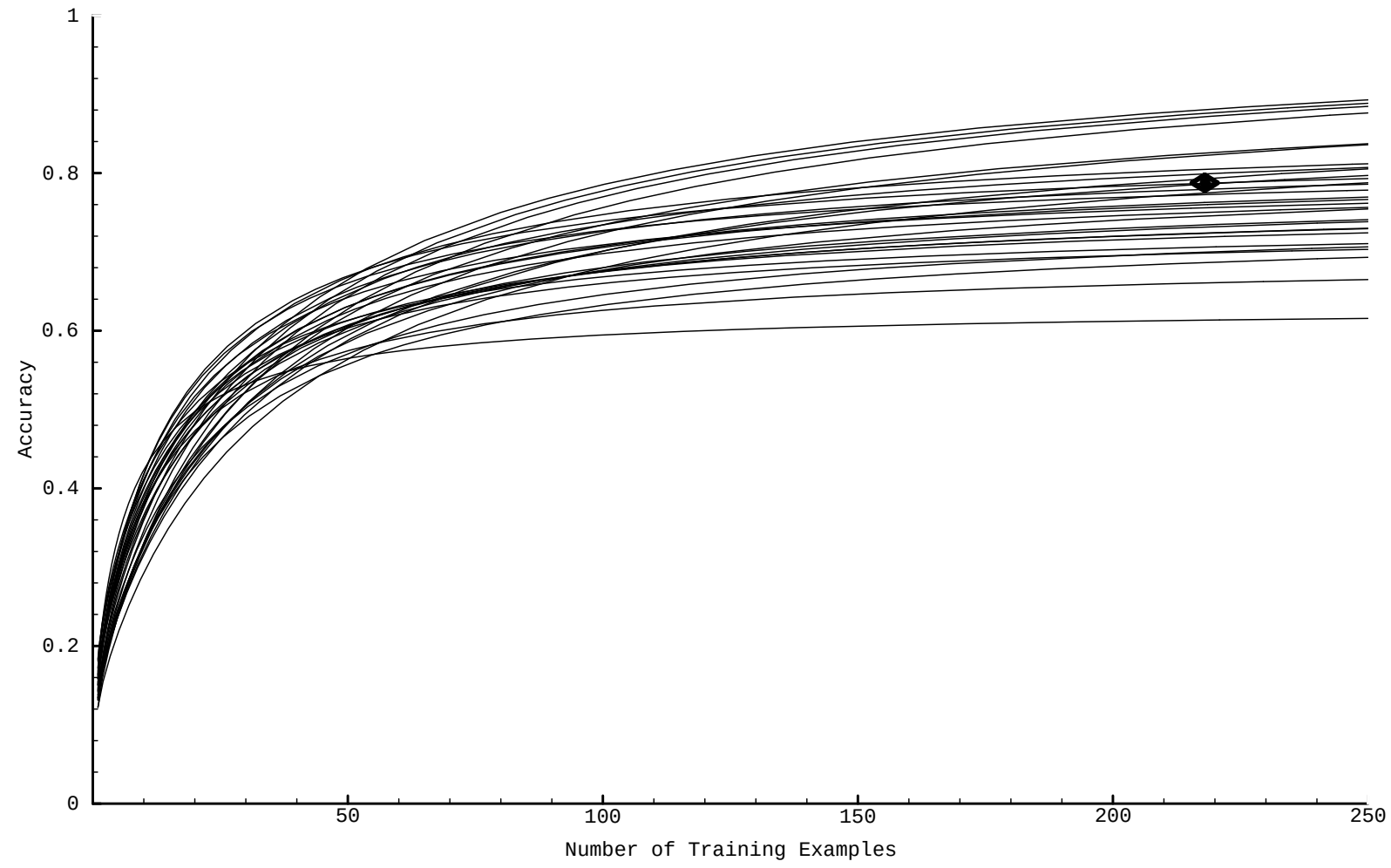

Figure 5.12: ED's predictive models for the 30 trials of the Audio domain -- The diamond shows the "true" accuracy.

| | Characterize | | Predict | |
|---|---|---|---|---|
| Domain | Loglike. | $R^2_{Seer}$ | Loglike. | $R^2_{Seer}$ |
| Soy(30) | -55315. | 0.954 | -1669. | 0.980 |
| Heart(30) | -54184. | 0.978 | -1467. | 0.949 |
| Audio(30) | -56057. | 0.979 | -1400. | 0.939 |

Table 5.16: ED's goodness-of-fit on learning-performance data from 1) the given 100 classified examples and 2) the full data set -- Values are composites from the 30 trials.

Table 5.17 gives the mean and observed standard deviation for all parameters in all three domains. The values of *start* and *skew* are by definition exactly the same as for Modelgen. The ED model set estimated slightly more optimistic *max* values, but they did not seem to be reliably closer or further from the "true" *max* value. Both models have a *d* parameter, but the *d*'s are used differently within the model set and so are not directly comparable. In the Soy domain, the ratio between the two *d*'s is 9.2. In Heart, it is 6.3 and in Audio it is 8.3. On the other hand, ED's *d* parameter *is* comparable to the Vapnik-Chervonenkis (VC) dimension from computational learning theory. For example, the mean *d* value for the Soy-domain characterization task was 47.692. After factoring out the effects of noise, multiple classes, and skew, a ED model will approximate the worst-case learning model of a learning task with a VC dimension of 48.

| | Characterization (all available classified examples, 3 trials) | | Prediction (100 classifies examples, 30 trials) | |
|---|---|---|---|---|
| | mean | s.d. | mean | s.d. |
| Soy | | | | |
| *d* | 1.186 | 0.0086 | 1.555 | 0.2212 |
| *max* | 1.000 | 0.0000 | 0.997 | 0.0171 |
| *start* | 0.087 | 0.0010 | 0.093 | 0.0089 |
| *skew* | 1.075 | 0.0019 | 1.081 | 0.0169 |
| $p_{300}$ | 0.840 | 0.0209 | 0.814 | 0.0220 |
| Heart | | | | |
| *d* | 0.711 | 0.1503 | 0.979 | 1.0641 |
| *max* | 0.760 | 0.0035 | 0.783 | 0.0695 |
| *start* | 0.503 | 0.0000 | 0.509 | 0.0110 |
| *skew* | 1.015 | 0.0000 | 1.042 | 0.0516 |
| $p_{290}$ | 0.752 | 0.0248 | 0.761 | 0.0504 |
| Audio | | | | |
| *d* | 0.562 | 0.0629 | 0.559 | 0.1956 |
| *max* | 0.852 | 0.0148 | 0.840 | 0.0934 |
| *start* | 0.141 | 0.0000 | 0.149 | 0.0162 |
| *skew* | 1.196 | 0.0000 | 1.191 | 0.0345 |
| $p_{218}$ | 0.788 | 0.0272 | 0.764 | 0.0640 |

Table 5.17: Mean and standard deviation of estimated ED parameters

So how does the ED model set do compared to the Modelgen data set? The question is interesting because the two model sets are identical except that Modelgen uses the simple $Burr_1$, while ED uses the computational-learning inspired EDit. Table 5.18 shows which did best in each combination of task and domain. Modelgen characterized better in 2 of the 3 domains. For the more important prediction task, Modelgen did better in one domain, ED did better in another, and there was no significant difference in the third domain.

**Analysis:** There is no reliable, measurable performance difference for prediction between ED and Modelgen, but Modelgen may characterize better. Both models have the same number of parameters (two determined by the class frequency of the classified examples and two determined by fitting learning-performance data). Modelgen's $Burr_1$ component is simpler and computationally much easier to work with than the ED models' EDit component. The ED model set is worth using if we want a *d* parameter we can interpret as the empirical analog of the VC dimension.

| Scenario | Task | Domain | Sig.Diff? | Better | $\alpha$ | test stat. |
|---|---|---|---|---|---|---|
| 1 | characterize | soy | yes | modelgen | 0.002 | 10.868 |
| 2 | characterize | heart | yes | modelgen | 0.016 | 3.702 |
| 3 | characterize | audio | no | - | 0.144 | -0.995 |
| 4 | predict | soy | no | - | 0.180 | 0.362 |
| 5 | predict | heart | yes | modelgen | 0.018 | 1.847 |
| 6 | predict | audio | yes | ED | <0.001 | -3.745 |

Table 5.18: Results of paired-difference analysis -- Neither the Modelgen nor the ED predicts consistently better.

### 5.2.4. NoExp Experiments

The NoExp experiments tested the set of learning-performance models described in Table 5.19. The set's deterministic component is made up of the $\text{model}_{noexp}$ and $\text{Burr}_{1.}$ Its nondeterministic component is the binomial distribution. The purpose of this experiment was to see if an even simpler submodel than $\text{model}_{gen}$ could be as effective. The alternative, $\text{model}_{noexp}$, still scales the curve between *start* and *max*, but it does not adjust for *skew* using exponentiation. The hope is that the *d* in $\text{model}_{noexp}$ can take the role of both *d* and *skew* in $\text{model}_{gen}$ and that exponentiation can be avoided, simplifying the model.

| Component | Submodel | Value |
|---|---|---|
| Det. #1 | $\text{model}_{noexp}$ | $p = \text{model}_{noexp}\left[(m-1)/d+1, start, max\right]$ $= start + (max - start)\dfrac{\text{model}_5\left[(m-1)/d+1\right]-0.5}{1-0.5}$ |
| Det. #2 | $\text{Burr}_1$ | $\text{model}_5[z] = z/(z+1)$ |
| Nondet. | binomial | $p^y(1-p)^{(k-y)}\binom{k}{y}$ |

Table 5.19: Learning-performance model set tested in the NoExp experiments

Figure 5.13 shows a typical characterization trial. Table 5.20 quantifies all the characterization trials in all three domains. The NoExp model does well; all $R^2_{Seer}$ values are greater than 0.96.

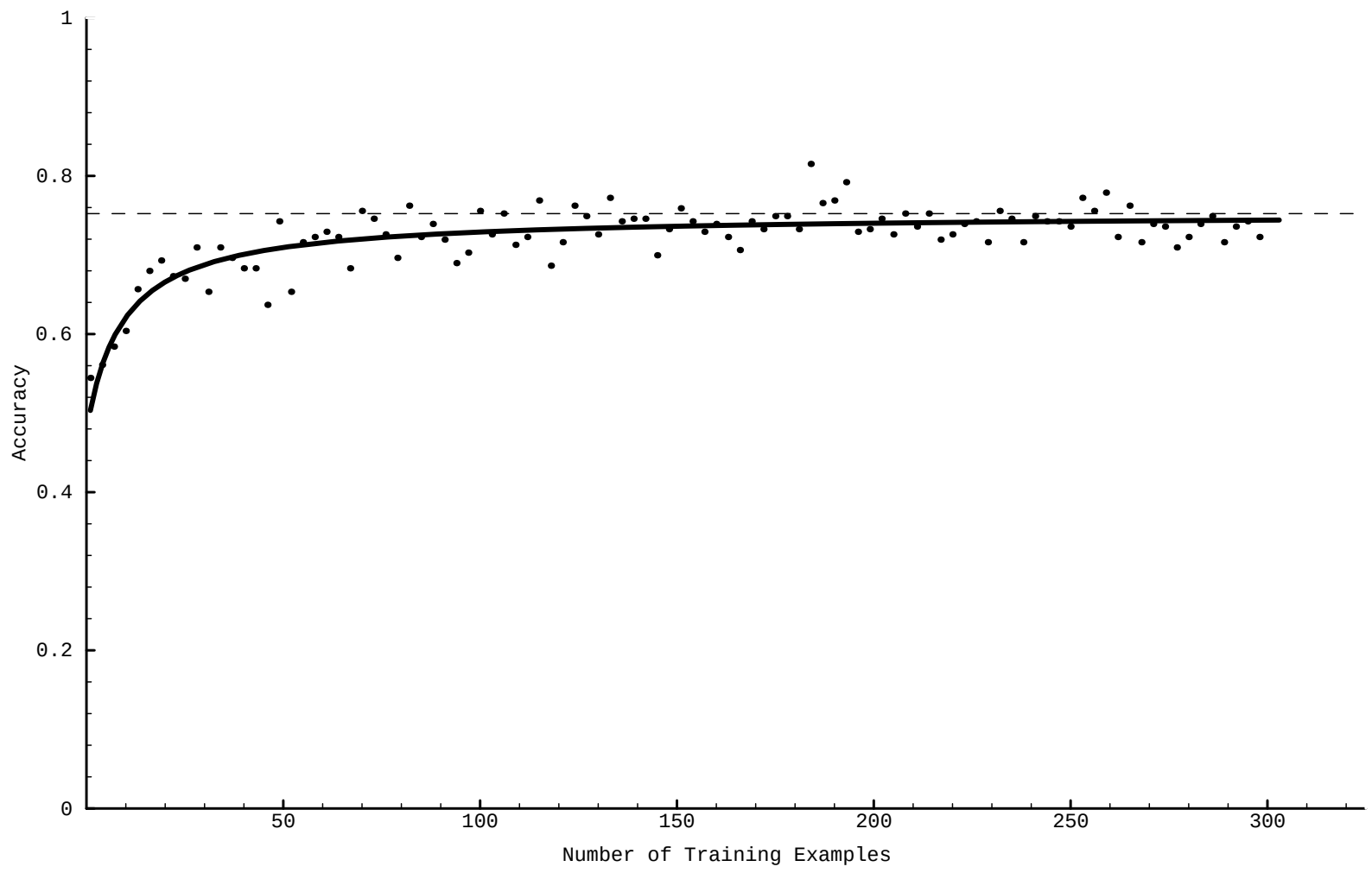

Figure 5.13: Characterizing learning-performance data from the Heart domain (Trial #1) with the NoExp model

| Domain | Loglike. | $R^2_{Seer}$ |
|---|---|---|
| Soy (3) | -7736.67 | 0.964 |
| Heart (3) | -7568.44 | 0.984 |
| Audio (3) | -5424.74 | 0.983 |

Table 5.20: Goodness-of-fit for NoExp characterizing learning-performance data from the three domains

Figure 5.14 shows all the models found for all 30 prediction trials in the Heart domain. As Table 5.21 shows, the NoExp models predicts fairly well based on a subset of examples but the lowest $R^2_{Seer}$ value, 0.866 in Heart domain, is lower than we have seen before. The plot suggests a cause: in a minority of the Heart-domain trials, the estimate of *max*, the asymptotic accuracy, was far off.

Table 5.22 gives the mean and observed standard deviation for all parameters in all three domains. The *d* parameter of the NoExp model set is incomparable with both the *d* parameter of Modelgen and the *d* parameter of ED.

| | Characterize | | Predict | |
|---|---|---|---|---|
| Domain | Loglike. | $R^2_{Seer}$ | Loglike. | $R^2_{Seer}$ |
| Soy(30) | -54903. | 0.962 | -1657. | 0.987 |
| Heart(30) | -54283. | 0.976 | -1607. | 0.866 |
| Audio(30) | -55972. | 0.980 | -1420. | 0.926 |

Table 5.21: NoExp's goodness-of-fit on learning-performance data from 1) the given 100 classified examples and 2) the full data set

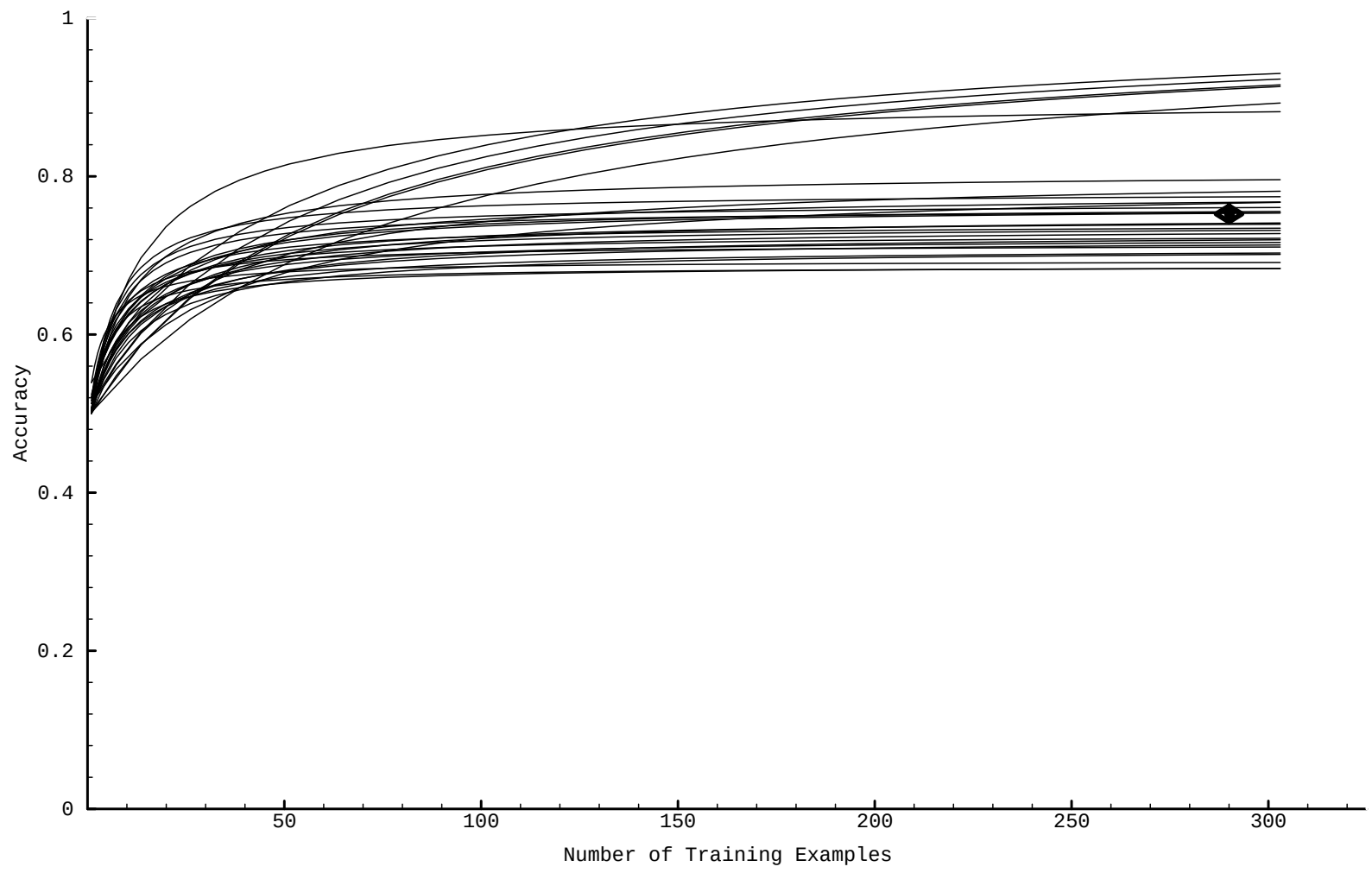

Figure 5.14: NoExp's predictive models for the 30 trials of the Heart domain

| | Characterization (all available classified examples, 3 trials) | | Prediction (100 classifies examples, 30 trials) | |
|---|---|---|---|---|
| | mean | s.d. | mean | s.d. |
| Soy | | | | |
| *d* | 39.561 | 0.3780 | 47.692 | 7.5012 |
| *max* | 1.000 | 0.0000 | 0.995 | 0.0268 |
| *start* | 0.087 | 0.0010 | 0.093 | 0.0089 |
| $p_{300}$ | 0.840 | 0.0209 | 0.828 | 0.0244 |
| Heart | | | | |
| *d* | 8.668 | 1.5806 | 10.866 | 9.9616 |
| *max* | 0.753 | 0.0024 | 0.773 | 0.0792 |
| *start* | 0.503 | 0.0000 | 0.509 | 0.0110 |
| $p_{290}$ | 0.752 | 0.0248 | 0.758 | 0.0621 |
| Audio | | | | |
| *d* | 15.259 | 1.6522 | 15.096 | 6.6226 |
| *max* | 0.835 | 0.0136 | 0.825 | 0.0992 |
| *start* | 0.141 | 0.0000 | 0.149 | 0.0162 |
| $p_{218}$ | 0.788 | 0.0272 | 0.762 | 0.0683 |

Table 5.22: Mean and standard deviation of estimated NoExp parameters

Table 5.23 compares NoExp to Modelgen on each combination of task and domain. For characterization, Modelgen did better in one domain, NoExp did better in the Heart domain, and there was no noticeable difference in the third domain. For prediction, in two domains, there was no significant difference in predictive power. In the Heart domain, Modelgen predicted better.

| Scenario | Task | Domain | Sig.Diff? | Better | α | test stat. |
|---|---|---|---|---|---|---|
| 1 | characterize | soy | yes | modelgen | 0.002 | 8.963 |
| 2 | characterize | heart | yes | NoExp | 0.026 | -2.846 |
| 3 | characterize | audio | no | - | 0.105 | -1.007 |
| 4 | predict | soy | no | - | 0.157 | -0.916 |
| 5 | predict | heart | yes | modelgen | 0.006 | 2.321 |
| 6 | predict | audio | no | - | 0.221 | -0.139 |

Table 5.23: Results of paired-difference analysis -- Modelgen predicted better in one domain.

**Analysis:** In most of the scenarios investigated, the exponent in the Modelgen model set is not important. The *d* parameter can be adjusted to compensate for the lack of an exponent. The NoExp models have the advantage of being simpler and easier to work with. On the other hand, they gets its simplicity by conflating two parameters (*skew* and *d*), one of which can be determined without fitting by just by looking at class frequencies. Modelgen's *d* parameter is thus more useful because it comes closer to quantifying the inherent hardness of learning a classification rule without regard to class frequencies and skew.

### 5.2.5. Pvar Model Set

The Pvar experiments tested the set of learning-performance models described in Table 5.24. The set's deterministic component is made up of the $model_{gen}$ and $Burr_1$, just as in the Modelgen experiments. Its nondeterministic component is the beta-binomial distribution. This experiment was designed to tell if there is any significant difference between using the binomial distribution and using the more sophisticated beta-binomial. The only difficulty in this comparison is that each distribution defines "goodness" differently. If one is much better than the other, however, it ought to be able to beat the other at prediction by either likelihood standard.

Table 5.26 gives the mean and observed standard deviation for all parameters in all three domains. A *pvar* value of 0 reduces the beta-binomial distribution to the binomial distribution. On all three domains, for the characterization task, the mean *pvar* is several standard deviations away from 0, suggesting the parameter is significant. On the prediction task, however, the mean *pvar* value is a best two standard deviations from 0 and, in the Soy domain, only one standard deviation from 0. This suggests that, for these prediction tasks, the beta-binomial distribution was not useful.

| Component | Sub-model | Value |
|---|---|---|
| Det. #1 | $\text{model}_{\text{gen}}$ | $p = \text{model}_{\text{gen}}\left[(m-1)/d+1, start, skew, max\right] = start + (max - start)\dfrac{\text{model}_5\left[(m-1)/d+1\right]^{skew(\text{Log}_{1/2}[start/2]-1)} - 0.5^{skew(\text{Log}_{1/2}[start/2]-1)}}{1-0.5^{skew(\text{Log}_{1/2}[start/2]-1)}}$ |
| Det. #2 | $\text{Burr}_1$ | $\text{model}_5[z] = z/(z+1)$ |
| Nondet. | beta-binomial | $\dfrac{\text{Beta}[y+\frac{p}{pvar},(k-y)+\frac{p}{pvar}]\binom{k}{y}}{\text{Beta}[\frac{p}{pvar},\frac{1-p}{pvar}]}$ |

Table 5.24: Learning-performance model set tested in the Pvar experiments

Table 5.26 compares Pvar to Modelgen on each combination of task and domain. Assuming that the binomial distribution is correct, Modelgen did significantly better in two of the six scenarios. In the other four, neither model set did significantly better. On the other hand, if we assume that beta-binomial is correct, then Pvar does significantly better in 4 of the scenarios, Modelgen does significantly better in one, and neither does significantly better in the remaining scenario.

**Analysis:** Only prediction matters here (since each finds the best characterizing fit by its criteria). Neither model set consistently beats the other by the other’s criteria. Thus, while the beta-binomial distribution has intuitive appeal as a way to measure variance in training examples, for these scenarios the simpler and more popular binomial distribution seems to suffice.

| | Characterization (all available classified examples, 3 trials) | | Prediction (100 classifies examples, 30 trials) | |
|---|---|---|---|---|
| | mean | s.d. | mean | s.d. |
| Soy | | | | |
| *d* | 12.773 | 0.0887 | 14.252 | 2.1762 |
| *max* | 1.000 | 0.0000 | 0.988 | 0.0458 |
| *pvar* | 0.013 | 0.0016 | 0.010 | 0.0060 |
| *start* | 0.087 | 0.0010 | 0.093 | 0.0089 |
| *skew* | 1.075 | 0.0019 | 1.081 | 0.0169 |
| $p_{300}$ | 0.840 | 0.0209 | 0.840 | 0.0321 |
| Heart | | | | |
| *d* | 6.113 | 1.1021 | 6.188 | 4.5237 |
| *max* | 0.753 | 0.0023 | 0.760 | 0.0516 |
| *pvar* | 0.002 | 0.0009 | 0.006 | 0.0063 |
| *start* | 0.503 | 0.0000 | 0.509 | 0.0110 |
| *skew* | 1.015 | 0.0000 | 1.042 | 0.0516 |
| $p_{290}$ | 0.752 | 0.0248 | 0.749 | 0.0451 |
| Audio | | | | |
| *d* | 5.067 | 0.5460 | 4.645 | 1.5405 |
| *max* | 0.823 | 0.0127 | 0.795 | 0.0856 |
| *pvar* | 0.006 | 0.0027 | 0.010 | 0.0052 |
| *start* | 0.141 | 0.0000 | 0.149 | 0.0162 |
| *skew* | 1.196 | 0.0000 | 1.191 | 0.0345 |
| $p_{218}$ | 0.788 | 0.0272 | 0.746 | 0.0650 |

Table 5.25: Mean and standard deviation of estimated Pvar parameters

| Scenario | Task | Domain | Sig.Diff? | Better | $\alpha$ | test stat. |
|---|---|---|---|---|---|---|
| Assuming the Binomial Distribution is Correct | | | | | | |
| 1 | characterize | soy | yes | modelgen | 0.049 | 1.912 |
| 2 | characterize | heart | yes | modelgen | 0.015 | 3.840 |
| 3 | characterize | audio | no | - | 0.144 | -1.000 |
| 4 | predict | soy | no | - | 0.150 | -0.863 |
| 5 | predict | heart | no | - | 0.155 | -0.511 |
| 6 | predict | audio | no | - | 0.140 | -0.794 |
| Assuming the Beta-Binomial Distribution is Correct | | | | | | |
| 1 | characterize | soy | yes | pvar | 0.004 | -7.425 |
| 2 | characterize | heart | yes | pvar | 0.025 | -2.903 |
| 3 | characterize | audio | no | - | 0.099 | -1.078 |
| 4 | predict | soy | yes | modelgen | <0.001 | 7.091 |
| 5 | predict | heart | yes | pvar | 0.015 | -1.947 |
| 6 | predict | audio | yes | pvar | 0.005 | -2.405 |

Table 5.26: Results of paired-difference analysis -- Neither model predicts consistently better.

## 5.3. Summary and Conclusion

The goal of this chapter was to find which (if any) of the model set designs were useful for characterizing and predicting learning performance. The goal was also to help develop criteria for choosing which of the useful models to use in a particular situation.

To meet these goals, Seer generated 99 learning-performance data sets (3 characterization trials and 30 prediction trials $\times$ 3 domains) using generalized cross-validation and thousands of runs of the C4.5 learning algorithm. For each of the 99 learning-performance data sets, Seer found the maximum-likelihood model from 5 model sets. Results were measured with loglikelihood and $R^2_{Seer}$, a measure of goodness-of-fit. The maximum-likelihood models were compared to each other using paired-difference analysis.

The Logistic model set was the only one that did poorly on both characterization and prediction. This model set is not appropriate for learning-performance data. The other models all did well on characterization and seemed to do as well as possible on prediction. The poor prediction seen in some trials seemed mostly attributable to unrepresentative subsets of the classified examples.

One way to select among the good performing model sets would be to always use the simplest. Simplicity is often measured by number of parameters. Most of the models had two parameters that needed to be fitted to the learning-performance data (*max* and *d*) and two parameters (*start* and *skew*) that could be determined from the class frequency of the classified examples. The NoExp model set did not have a *skew* parameter. The Pvar model set had an additional parameter, *pvar*, to fit to the data. Simplicity can also be evaluated by how easy a model set is to work with, computationally. By this criterion, NoExp is again the simplest. The next simplest would be Modelgen, whereas. ED and Pvar would be the most complex.

Another way to select among candidate model sets is to choose the most useful and meaningful parameters. For example, the *max* parameter found in most of the models is very useful because it is an estimate of asymptotic accuracy. Likewise, *start* and *skew* quantify the effect of multiple classes and skewed classes on the learning. Also, the *pvar* parameter tries to quantify variance in the training data. Finally, most of the model sets have a *d* parameter. This is a measure of learning difficulty after other parameterized effects are factored out. Uniquely among the model sets, the ED model set produces a *d* that is comparable to the *d* produced through Vapnik-Chervonenkis-dimension analysis. Thus, although one model set is worst, selection of the best model set depends on a tradeoff between simplicity and usefulness. Table 5.27 summarizes the tradeoff.

| Model Set | Fit | Complexity | Measured effects |
|---|---|---|---|
| Modelgen | good | medium | multiple classes, skew, noise, difficulty |
| Logistic | poor | - | - |
| ED | good | high | multiple classes, skew, noise, VC-comparable difficulty |
| NoExp | good | low | multiple classes, noise, difficulty |
| Pvar | good | high | multiple classes, skew, noise, variance from training examples |

Table 5.27: Tradeoffs in selection of a model set. Except for Logistic, all the model sets characterize well and predict as well as the data allows. Some of the model sets are more simple, but others have useful and interesting parameters.

# 6. Conclusion

This concluding chapter highlights the important points of the preceding chapters, suggests possibilities for future work, and discusses the principle contributions of this research.

## 6.1. Summary

Inductive classification learning is the process of creating a classification rule based on a set of training examples. Two questions are of interest when constructing an expert system from classified examples:

1. How many classified examples will be needed to create a classification rule of the desired accuracy?
2. If an unlimited number of examples were possible, what accuracy would be possible?

Seer answers these questions by first generating learning-performance data from a set of existing classified examples. Then, it fits a learning-performance model to the data.

Seer advances the start of the art by making regression on learning-performance data practicable. Theoretical approaches to modeling learning performance, when applicable, can make predictions over a broad scope of learning tasks. For practical problems, however, with noise and the most useful machine-learning algorithms, theory cannot yet predict average-case performance. Empirical approaches such as Seer are more narrow in scope. They work by gathering observations of actual learning-performance and then generalizing it. (They are in essence empirically learning about empirical learning.) Seer's basic approach, which goes back at least 75 years, is to use regression to fit a curve (a learning-performance model) to the observed learning-performance data. Seer differs from previous learning-curve-regression work:

1. with learning-performance models that embody the best constraints for classification learning and most useful parameters
2. with algorithms that efficiently find maximum-likelihood models, and

3. with a demonstration on real-world data of a practicable application.

Formally, Seer's task starts with learning-performance data in the form of a set of tuples, $\{<m_1,y_1,k_1>, <m_1,y_1,k_1>, ...\}$, where $m_i$ is the number of training examples given to a learning program, $y_i$ is the number of testing examples the classification rule classifies correctly, and $k_i$ is the number of testing examples given to the learner's hypothesis. When the total number of classified examples given to Seer is small, it uses a new procedure called generalized cross-validation to generate learning-performance data (at a cost of some statistical independence). Although we can informally think of the models Seer fits to that data as "curves", more formally a learning-performance model is a function. From the number of training examples, *m*, and the number of testing examples, *k*, a model predicts the probability of correctly classifying exactly *y* examples.

To create Seer, two questions needed answers: first, which sets of candidate learning-performance models should Seer consider, and, second, how should Seer efficiently find the maximum-likelihood model from a model set. The question of creating candidate models was treated as a three-part design problem. The first involved creating deterministic models for noise-free learning of two equally probable classes. Two models inspired by computational learning theory, EDmodel and BurrModel, were found to have a much different shape than the models of logit and probit regression. The second part of the design involved modeling the effect of multiple classes, skewed classes, and noise. The new $model_{gen}$ heuristically models the effects of interest with remarkably few parameters. The third part involved creating a nondeterministic model. The binomial distribution should be sufficient if variation in testing data is the only concern. However, if variation in the training data is also a concern, then the beta-binomial distribution might be better.

The second question of how to find the maximum-likelihood model cannot be solved with popular and simple techniques such as least-squares regression because the assumptions of least-squares regression, such as constant variance, are not reasonable for learning-performance data. The problem of finding the maximum-likelihood model for learning-performance data is an instance of nonlinear programming. Seer, however, overcomes the difficulties of nonlinear programming by exploiting problem constraints and using appropriate transformations to reduce the problem to nonlinear regression, which Seer solves with an efficient iterative algorithm.

Experiments were conducted to determine which, if any, of the model sets designs were useful for characterizing and predicting learning performance. Seer generated 99 learning-performance data sets (3 characterization trials and 30 prediction trials × 3 domains). It used generalized cross-validation and thousands of runs of the C4.5 learning algorithm. For each of the 99 learning-performance data sets, Seer found the maximum-likelihood model from the 5 model sets. Results were measured with loglikelihood and $R^2_{Seer}$, a measure based on loglikelihood but normalized to the range 0.0 to 1.0. The maximum-likelihood models were compared head-to-head using paired-difference analysis. All the model sets (except the logistic model set) did well on characterization and seemed to do as well as possible on prediction. The poor prediction seen in some trials seemed mostly attributable to unrepresentative subsets of the classified examples.

One way to select among the good performing model sets would be to always use the simplest. Of the model sets considered, the one named NoExp had fewer parameters than the other models and was the easiest with which to work computationally. Another way to select model sets, however, would be to choose the one with the most useful and meaningful parameters. For example, the *max* parameter found in most of the models is very useful because it is an estimate of asymptotic accuracy. Likewise, *start* and *skew* quantify and predict the effect of multiple classes and skewed classes on the learning. Also, the *pvar* parameter tries to quantify variance in the training data. Finally, most of the model sets have a *d* parameter. This is a measure of learning difficulty after other parameterized effects are factored out. Uniquely among the model sets, the ED model set produces a *d* that is comparable to the *d* produced through Vapnik-Chervonenkis-dimension analysis. Thus, except for the Logistic model set, all the model sets characterize well and predict as well as the data allows. Some of the model sets are more simple, but others have useful and interesting parameters. Selecting the best one will always involve tradeoffs.

## 6.2. Future Work

This section discusses opportunities for future work. Three directions are considered: 1) expanding the amount of performance-data available, 2) exploring the *pvar* and $d_{\mathbf{ED}}$ model parameters, and 3) generalizing the models to be multivariate.

**Infinite Data Supply --** If our only interest is in estimating how many more classified examples we need to create an acceptable expert system, then Seer's current method of generating learning-performance data suffices. However, what if we are interested in characterizing the behavior of our inductive learning under specific distributions of learning targets and learning examples? In that case, the distributions can be used to generate random targets and examples. This gives Seer an inexhaustible supply of classified examples. By using each of these synthetic examples only once, the learning-performance data Seer creates is statistically independent. Confidence intervals can be obtained for all parameter estimates either from the covariance matrix created during model fitting or via Monte Carlo simulation [Press *et al.* 1992, p. 695]. The down side to this method is that it is no longer based on real-world data, but rather on a synthetic-data generator.

**The *pvar* and $d_{ED}$ Parameters --** This thesis presented the *pvar* parameter from the beta-binomial distribution as a way to characterize variation in training examples. This parameter could be important because it captures a source of uncertainty that is otherwise ignored. For example, it could tell us that although one inductive learner appears to have greater predictive accuracy than another (as measured with ordinary cross validation) that the difference is not statistically significant because it could be explained by training-example variation. In the experiments of this thesis, however, the *pvar* parameter did not contribute, in a compelling way, to prediction accuracy. Further work is needed to determine if and when reliable *pvar* values can be estimated. One way to try to gain an understanding of the problem might be to use infinite supplies of data as suggested above.

The thesis also presented the ED model, a model designed to create a parameter $d$ that would be comparable to the Vapnik-Chervonenkis dimension. Haussler [1988] promotes the VC-dimension, $d_{\mathbf{VC}}$ as a measure of inductive bias. Because $d_{ED}$ can be determined in many cases where $d_{\mathbf{VC}}$ cannot (and visa versa), it may be a useful measure of inductive bias. Future work is needed to explore this application.

**Multivariate Models --** When predicting learning performance, the number of training examples is typically the most important dependent variable, but it is not the only one of possible interest. We might also want to model how other variables affect performance. Examples of such variables include:

- Choice of inductive learning algorithm; for example, C4.5 versus ID3

- Configuration of learning algorithm; for example, number of hidden units in a neural-net algorithm
- Number of irrelevant attributes in the classified examples
- Complexity of the target; for example, number of disjuncts in the target

Some of these variables are nearly always available, for example, the learning algorithm and its configuration are typically known. Other variables, such as number of disjuncts in the target, would only be known for synthetically-generated learning problems.

The result of multivariate regression would be a model that might look something like the plot in Figure 6.15. It would map dependent variables into expected accuracy, capturing with only a few numbers the general effects of learning-task attributes.

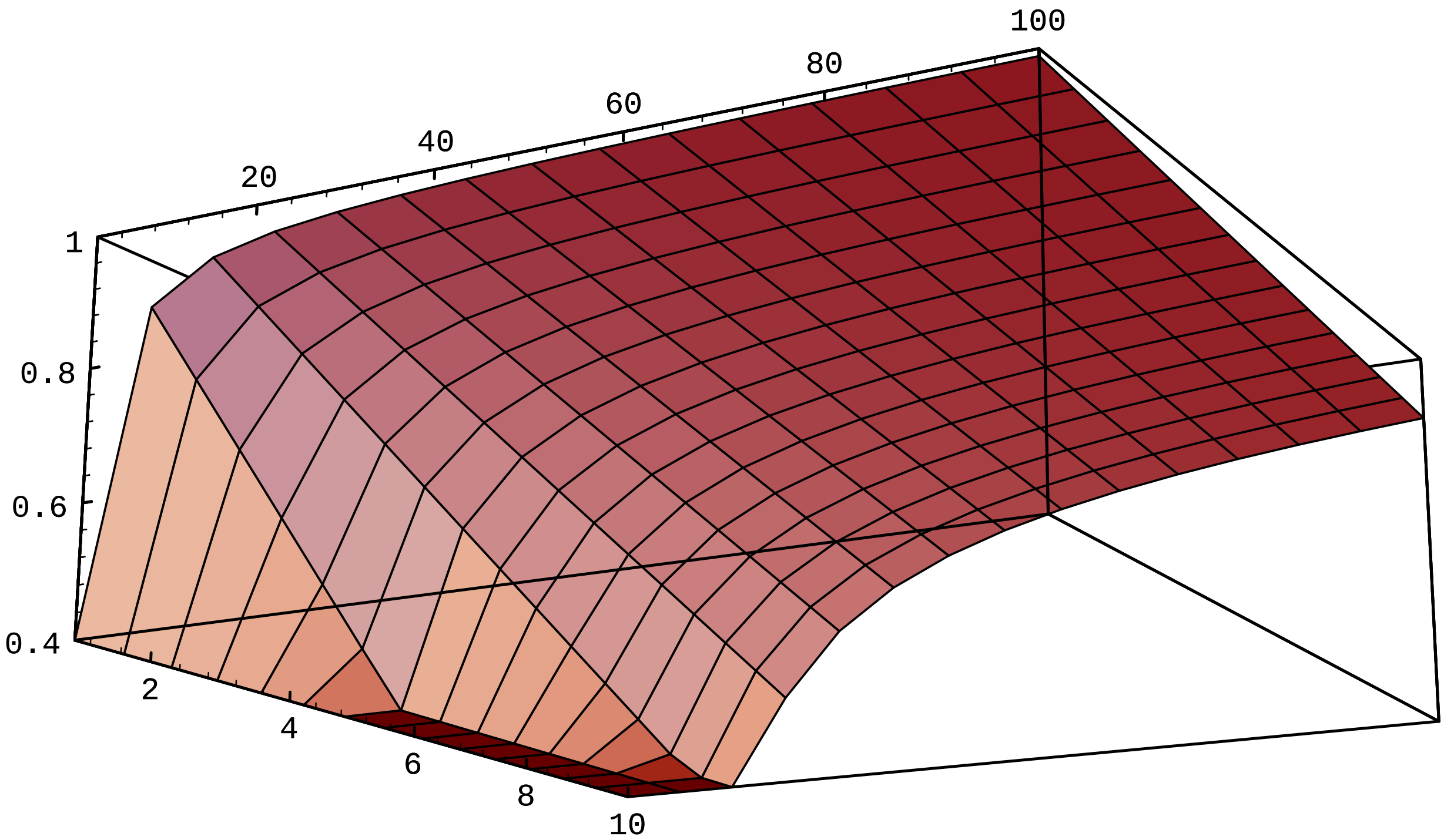

Figure 6.15: Example of what a multivariate learning-performance regression model might look like. In this example, the vertical axis represents accuracy, the horizontal axis represents number of training examples, and the axis that goes into the page represents the number of disjuncts in the learning target.

## 6.3. Contributions

The general contribution of the thesis was to show how regression could be used to practically and efficiently analyze inductive classification learning. More specifically, given a set of classified examples and a learning system, such as C4.5, the thesis presented a way to predict learning average-case learning performance and asymptotic accuracy. The data could be noisy and could originate from multiple and skewed classes. Furthermore, the thesis developed and evaluated several reasonable and effective model sets of learning performance. These models varied in their simplicity and in the usefulness of their parameters. One model set, EDmodel, produced a *d* parameter comparable to the Vapnik-Chervonenkis dimension from Computation Learning Theory. It developed a heuristic that predicts, without needing learning-performance data, the effect of skewed classes and multiple classes on learning. It developed a measure, *pvar*, of the variation caused by different training-example sets. (Most similar statistical methods only measure the variation caused by different testing example sets.) It argued that maximum likelihood is the most appropriate criterion for selecting a model within a family of models that best fits the data and it showed how such maximum-likelihood models could be found efficiently.. It introduced a new method called generalized cross-validation to create learning-performance data from a finite set of classified examples, if dependencies in the data are allowed. It was observed that learning-performance data based on subsets of the classified examples are sometimes unrepresentative. The thesis identified likelihood as the most appropriate way to judge the goodness-of-fit of a model to data and defined $R^2_{Seer}$ (a new measure based on likelihood) as a convenient way to normalize likelihood. Finally, the thesis showed how to use pair-difference analysis to compare competing learning-performance models.

# Vita

Carl Myers Kadie was born on July 4, 1962, in Denver, Colorado. He attended the University of Illinois at Urbana-Champaign, where he received a Bachelor of Science degree in computer science with highest honors in January 1995. While an undergraduate, he worked every other semester as a co-op at IBM, Houston, on Space Shuttle software and related projects. Following graduation he worked for 8 months as a co-op at IBM Research, Yorktown Heights, NY, on the prototype of the Prodigy online service.

In the fall of 1985 he began his graduate studies in computer science at the University of Illinois. He completed his Masters of Science degree in October of 1989 with a thesis on machine learning of robot operators from examples of operator effects. In 1989 he was awarded Fannie and John Hertz Fellowship; it supported his graduate work for the next five years. In August of 1994, he started work in the Advanced Technology Division of Microsoft Corporation as a software development engineer working on Internet features for MSN, the Microsoft Network. His Ph.D., which focused on using maximum-likelihood regression to predict learning-speed curves, was conferred in October 1995.

**Publications and Presentations**